\documentclass{article}

\usepackage[main, final]{neurips_2026}
\makeatletter
\renewcommand{\@noticestring}{}
\makeatother

\usepackage[utf8]{inputenc} 
\usepackage[T1]{fontenc}    
\usepackage{hyperref}       
\usepackage{url}            
\usepackage{booktabs}       
\usepackage{amsfonts}       
\usepackage{nicefrac}       
\usepackage{microtype}      
\usepackage{xcolor}         
\usepackage{soul}
\sethlcolor{gray!20} 
\usepackage{amssymb}
\usepackage{amsmath}
\usepackage{empheq}
\usepackage{enumitem}
\usepackage{algorithm}

\usepackage{algpseudocode}
\usepackage{booktabs}
\usepackage{graphicx}
\usepackage[table]{xcolor}
\usepackage{soul}
\usepackage{xcolor}
\usepackage{subcaption} 
\usepackage{multirow}

\usepackage{wrapfig}
\usepackage{listings}

\usepackage[table]{xcolor}
\usepackage{booktabs}
\definecolor{grpbgfade}{HTML}{F0F0F0}   
\definecolor{grpbg}{HTML}{DEDEDE}       
\definecolor{oursbg}{HTML}{F3F3F3}      
\colorlet{fadetx}{black!65}             

\newcommand{\bl}[1]{\textcolor{fadetx}{#1}}
\newcommand{\grpfade}[1]{\rowcolor{grpbgfade}\multicolumn{8}{c}{\textcolor{fadetx}{\textbf{#1}}} \\}
\newcommand{\grp}[1]{\rowcolor{grpbg}\multicolumn{8}{c}{\textbf{#1}} \\}

\definecolor{codeblue}{rgb}{0.25,0.5,0.5}
\definecolor{codekw}{rgb}{0.85,0.18,0.50}
\definecolor{codesign}{RGB}{0,0,255}
\definecolor{codefunc}{rgb}{0.85,0.18,0.50}

\lstdefinelanguage{PseudoColor}{
    basicstyle=\fontsize{7.2pt}{8.2pt}\ttfamily\selectfont,
    columns=fullflexible,
    breaklines=true,
    showstringspaces=false,
    commentstyle=\color{codeblue},
    keywordstyle=\color{blue}\bfseries,
    morekeywords={for,to,do,return},
    literate=
      {*}{{\color{codesign}*}}{1}
      {-}{{\color{codesign}-}}{1}
      {+}{{\color{codesign}+}}{1}
      {F}{{\color{codefunc}F}}{1}
      {G}{{\color{codefunc}G}}{1}
      {decode}{{\color{codefunc}decode}}{6}
}

\title{Efficient Image Synthesis with Sphere Latent Encoder
}

\author{
Tung Do
\quad
Thuan Hoang Nguyen
\quad
Hao Li \\
Mohamed Bin Zayed University of Artificial Intelligence, UAE \\
\texttt{\{tung.do,thuan.nguyen,hao.li\}@mbzuai.ac.ae} \\[5pt]
\url{https://sphere-latent-encoder.github.io/}
}

\usepackage{graphicx}
\usepackage{array}
\usepackage{threeparttable} 

\definecolor{animalcolor}{RGB}{240,90,90}      
\definecolor{flowercolor}{RGB}{80,140,240}     
\definecolor{imagenetcolor}{RGB}{90,190,120}   

\begin{document}
\def\mA{\mathcal{A}}
\def\mB{\mathcal{B}}
\def\mC{\mathcal{C}}
\def\mD{\mathcal{D}}
\def\mE{\mathcal{E}}
\def\mF{\mathcal{F}}
\def\mG{\mathcal{G}}
\def\mH{\mathcal{H}}
\def\mI{\mathcal{I}}
\def\mJ{\mathcal{J}}
\def\mK{\mathcal{K}}
\def\mL{\mathcal{L}}
\def\mM{\mathcal{M}}
\def\mN{\mathcal{N}}
\def\mO{\mathcal{O}}
\def\mP{\mathcal{P}}
\def\mQ{\mathcal{Q}}
\def\mR{\mathcal{R}}
\def\mS{\mathcal{S}}
\def\mT{\mathcal{T}}
\def\mU{\mathcal{U}}
\def\mV{\mathcal{V}}
\def\mW{\mathcal{W}}
\def\mX{\mathcal{X}}
\def\mY{\mathcal{Y}}
\def\mZ{\mathcal{Z}}
\def\bbN{\mathbb{N}}
\def\bbR{\mathbb{R}}
\def\bbP{\mathbb{P}}
\def\bbQ{\mathbb{Q}}
\def\bbE{\mathbb{E}}
\def\1n{\mathbf{1}_n}
\def\0{\mathbf{0}}
\def\1{\mathbf{1}}
\def\A{{\bf A}}
\def\B{{\bf B}}
\def\C{{\bf C}}
\def\D{{\bf D}}
\def\E{{\bf E}}
\def\F{{\bf F}}
\def\G{{\bf G}}
\def\H{{\bf H}}
\def\I{{\bf I}}
\def\J{{\bf J}}
\def\K{{\bf K}}
\def\L{{\bf L}}
\def\M{{\bf M}}
\def\N{{\bf N}}
\def\O{{\bf O}}
\def\P{{\bf P}}
\def\Q{{\bf Q}}
\def\R{{\bf R}}
\def\S{{\bf S}}
\def\T{{\bf T}}
\def\U{{\bf U}}
\def\V{{\bf V}}
\def\W{{\bf W}}
\def\X{{\bf X}}
\def\Y{{\bf Y}}
\def\Z{{\bf Z}}
\def\a{{\bf a}}
\def\b{{\bf b}}
\def\c{{\bf c}}
\def\d{{\bf d}}
\def\e{{\bf e}}
\def\f{{\bf f}}
\def\g{{\bf g}}
\def\h{{\bf h}}
\def\i{{\bf i}}
\def\j{{\bf j}}
\def\k{{\bf k}}
\def\l{{\bf l}}
\def\m{{\bf m}}
\def\n{{\bf n}}
\def\o{{\bf o}}
\def\p{{\bf p}}
\def\q{{\bf q}}
\def\r{{\bf r}}
\def\s{{\bf s}}
\def\t{{\bf t}}
\def\u{{\bf u}}
\def\v{{\bf v}}
\def\w{{\bf w}}
\def\x{{\bf x}}
\def\y{{\bf y}}
\def\z{{\bf z}}
\def\balpha{\mbox{\boldmath{$\alpha$}}}
\def\bepsilon{\mbox{\boldmath{$\epsilon$}}}
\def\bbeta{\mbox{\boldmath{$\beta$}}}
\def\bdelta{\mbox{\boldmath{$\delta$}}}
\def\bgamma{\mbox{\boldmath{$\gamma$}}}
\def\blambda{\mbox{\boldmath{$\lambda$}}}
\def\bsigma{\mbox{\boldmath{$\sigma$}}}
\def\btheta{\mbox{\boldmath{$\theta$}}}
\def\bomega{\mbox{\boldmath{$\omega$}}}
\def\bxi{\mbox{\boldmath{$\xi$}}}
\def\bnu{\mbox{\boldmath{$\nu$}}}
\def\bphi{\mbox{\boldmath{$\phi$}}}
\def\bmu{\mbox{\boldmath{$\mu$}}}
\def\bDelta{\mbox{\boldmath{$\Delta$}}}
\def\bOmega{\mbox{\boldmath{$\Omega$}}}
\def\bPhi{\mbox{\boldmath{$\Phi$}}}
\def\bLambda{\mbox{\boldmath{$\Lambda$}}}
\def\bSigma{\mbox{\boldmath{$\Sigma$}}}
\def\bGamma{\mbox{\boldmath{$\Gamma$}}}
\newcommand{\myprob}[1]{\mathop{\mathbb{P}}_{#1}}
\newcommand{\myexp}[1]{\mathop{\mathbb{E}}_{#1}}
\newcommand{\mydelta}[1]{1_{#1}}
\newcommand{\myminimum}[1]{\mathop{\textrm{minimum}}_{#1}}
\newcommand{\mymaximum}[1]{\mathop{\textrm{maximum}}_{#1}}
\newcommand{\mymin}[1]{\mathop{\textrm{minimize}}_{#1}}
\newcommand{\mymax}[1]{\mathop{\textrm{maximize}}_{#1}}
\newcommand{\mymins}[1]{\mathop{\textrm{min.}}_{#1}}
\newcommand{\mymaxs}[1]{\mathop{\textrm{max.}}_{#1}}
\newcommand{\myargmin}[1]{\mathop{\textrm{argmin}}_{#1}}
\newcommand{\myargmax}[1]{\mathop{\textrm{argmax}}_{#1}}
\newcommand{\myst}{\textrm{s.t. }}
\newcommand{\denselist}{\itemsep -1pt}
\newcommand{\sparselist}{\itemsep 1pt}

\newcommand{\cyan}[1]{\textcolor{cyan}{#1}}
\newcommand{\blue}[1]{\textcolor{blue}{#1}}
\newcommand{\magenta}[1]{\textcolor{magenta}{#1}}
\newcommand{\pink}[1]{\textcolor{pink}{#1}}
\newcommand{\green}[1]{\textcolor{green}{#1}}
\newcommand{\gray}[1]{\textcolor{gray}{#1}}
\newcommand{\mygreen}[1]{\textcolor{mygreen}{#1}}
\newcommand{\purple}[1]{\textcolor{purple}{#1}}
\newcommand{\greena}[1]{\textcolor{greena}{#1}}
\newcommand{\bluea}[1]{\textcolor{bluea}{#1}}
\newcommand{\reda}[1]{\textcolor{reda}{#1}}
\def\changemargin#1#2{\list{}{\rightmargin#2\leftmargin#1}\item[]}
\let\endchangemargin=\endlist
\newcommand{\cm}[1]{}
\newcommand{\mhoai}[1]{{\color{magenta}\textbf{[MH: #1]}}}
\newcommand{\mtodo}[1]{{\color{red}$\blacksquare$\textbf{[TODO: #1]}}}
\newcommand{\myheading}[1]{\vspace{1ex}\noindent \textbf{#1}}
\newcommand{\htimesw}[2]{\mbox{$#1$$\times$$#2$}}
\newif\ifshowsolution
\showsolutiontrue
\ifshowsolution

\newcommand{\Solution}[2]{\paragraph{\bf $\bigstar $ SOLUTION:} {\sf #2} }
\newcommand{\Mistake}[2]{\paragraph{\bf $\blacksquare$ COMMON MISTAKE #1:} {\sf #2} \bigskip}
\else
\newcommand{\Solution}[2]{\vspace{#1}}
\fi
\newcommand{\truefalse}{
\begin{enumerate}
	\item True
	\item False
\end{enumerate}
}
\newcommand{\yesno}{
\begin{enumerate}
	\item Yes
	\item No
\end{enumerate}
}
\newcommand{\Sref}[1]{Sec.~\ref{#1}}
\newcommand{\Eref}[1]{Eq.~(\ref{#1})}
\newcommand{\Fref}[1]{Fig.~\ref{#1}}
\newcommand{\Tref}[1]{Table~\ref{#1}}
\newcommand{\Aref}[1]{Algo.~\ref{#1}}

\maketitle


\begin{figure}[H]
    \centering
    \noindent
    \begin{minipage}[c]{0.95\linewidth}
        \centering
        \includegraphics[width=\linewidth]{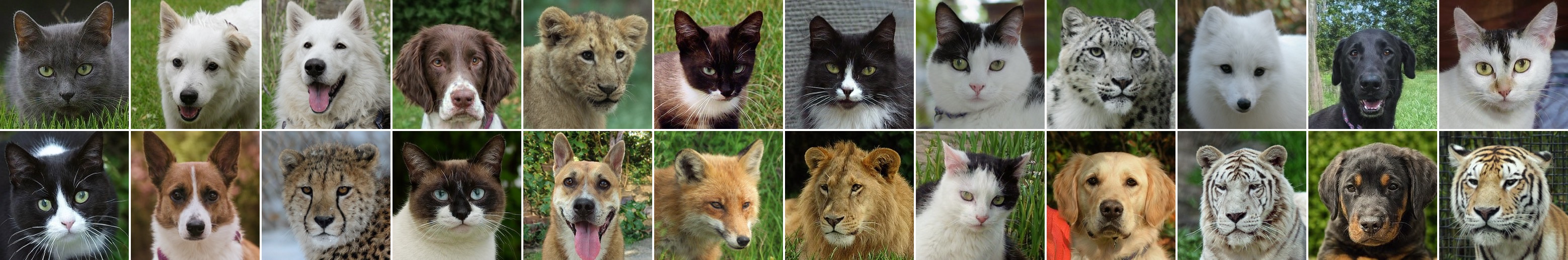}
    \end{minipage}%
    \hspace{-0.2cm}
    \begin{minipage}[c][2.17cm][c]{0.049\linewidth}
        \centering
        \rule{1.5pt}{2.15cm}
        \hspace{-0.1cm}
        \raisebox{1cm}[0cm]{\rotatebox[origin=c]{270}{\footnotesize \textcolor{animalcolor}{Animal-Faces}}}
    \end{minipage}
    
    \vspace{0.17cm}

    \begin{minipage}[c]{0.95\linewidth}
        \centering
        \includegraphics[width=\linewidth]{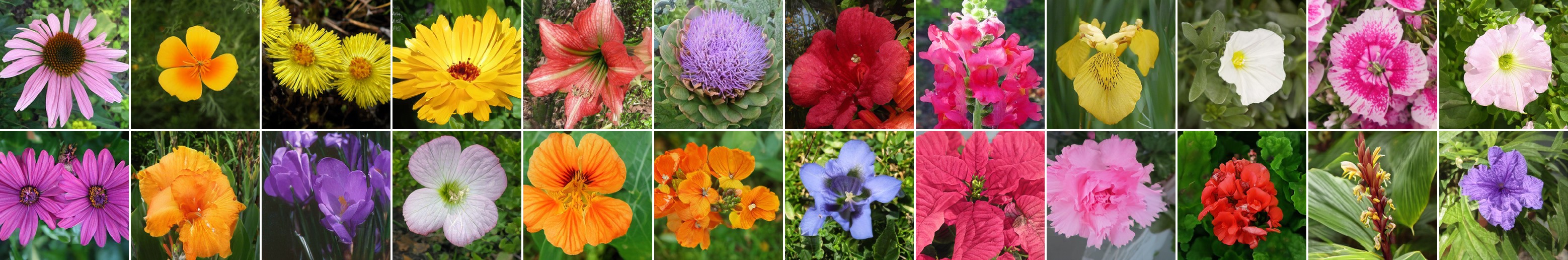}
    \end{minipage}%
    \hspace{-0.2cm}
    \begin{minipage}[c][2.17cm][c]{0.049\linewidth}
        \centering
        \rule{1.5pt}{2.15cm}
        \hspace{-0.12cm}
        \raisebox{1cm}[0cm]{\rotatebox[origin=c]{270}
        {\footnotesize \textcolor{flowercolor}{OxFord-Flowers}}}
    \end{minipage}

    \vspace{0.17cm}
    
    \begin{minipage}[c]{0.95\linewidth}
        \centering
        \includegraphics[width=\linewidth]{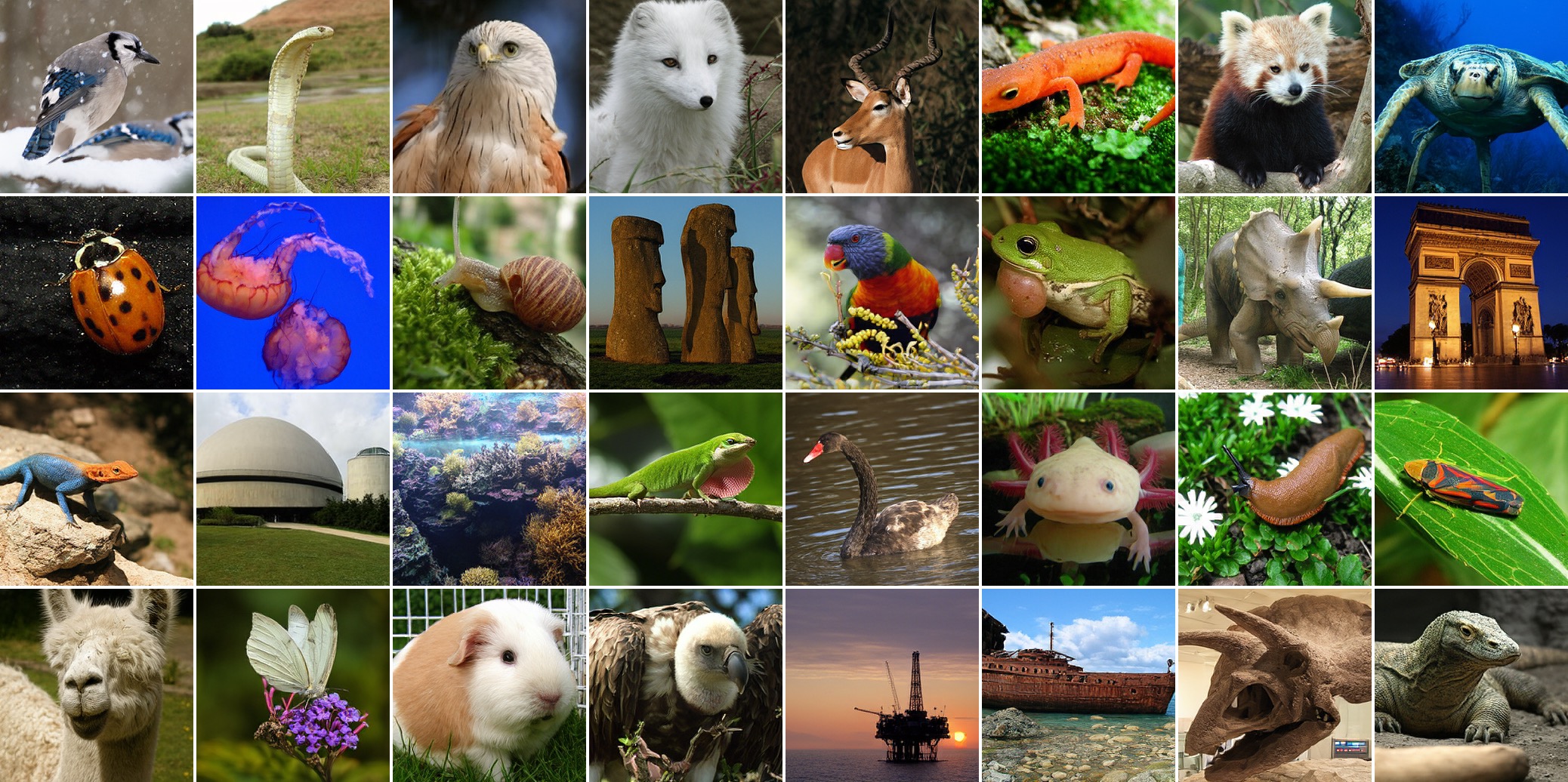}
    \end{minipage}%
    \hspace{-0.2cm}
    \begin{minipage}[c][2.17cm][c]{0.049\linewidth}
        \centering
        \rule{1.5pt}{6.57cm}
        \hspace{-0.15cm}
        \raisebox{3.3cm}[0cm]{\rotatebox[origin=c]{270}{\footnotesize \textcolor{imagenetcolor}{ImageNet-1K}}}
    \end{minipage}

\caption{Generated samples by Sphere Latent Encoder in \textbf{4 steps} on \textcolor{animalcolor}{Animal-Faces} \cite{choi2020stargan}, \textcolor{flowercolor}{Oxford-Flowers} \cite{nilsback2008automated}, and \textcolor{imagenetcolor}{ImageNet-1K} \cite{imagenet} (256 × 256).}

\end{figure}

\begin{abstract}

Few-step image generation has seen rapid progress, with consistency and meanflow-based methods significantly reducing the number of sampling steps. Despite their low inference cost, these approaches often suffer from training instability and limited scalability. Sphere Encoder is a recent alternative that produces high-quality images in only a few steps; however, it requires repeated transitions between the pixel space and latent space during inference while jointly optimizing reconstruction and generation within a single architecture. This design leads to computational inefficiency and objective conflict between reconstruction and generation. To address these limitations, we decouple the framework into a fixed pretrained image encoder and a separate latent denoising model trained entirely in a spherical latent space. Our approach eliminates repeated pixel-space operations during training and inference, improving efficiency and allowing reconstruction and generation to specialize independently. On Animal-Faces, Oxford-Flowers and ImageNet-1K datasets, our method significantly outperforms Sphere Encoder in both generation quality and inference speed, while achieving competitive results against strong few-step and multi-step baselines.
\end{abstract}

\section{Introduction}
Flow Matching \cite{lipman2023flowmatching, albergo2023building, liu2022flowstraightfast} provides a principled framework for transforming a prior distribution into a target data distribution by learning transport paths between them. Due to its scalability and strong performance, it has been widely adopted in large-scale generative systems such as Stable Diffusion 3 \cite{esser2024scaling}, FLUX \cite{flux2024}, and Qwen-Image \cite{wu2025qwenimagetechnicalreport}. However, standard flow matching relies on iterative sampling, which requires multiple function evaluations and results in high inference cost. Recent works \cite{song2023consistency, song2023improved, lu2024simplifying, geng2025meanflow, geng2025improved} aim to reduce the number of sampling steps, moving toward few-step or even single-step generation. Despite these advances in inference efficiency, previous work has shown that these few-step models, including Consistency Models \cite{song2023consistency} and MeanFlow \cite{geng2025meanflow} are often unstable and sensitivity to hyperparameters \cite{hu2025cmt}, and may even exhibit inherent instability when trained from scratch \cite{kim2026stabilizing}. Furthermore, the MeanFlow objective introduces conflicting optimization signals, leading to slow convergence and further difficulties in training \cite{zhang2025alphaflow}. Consequently, these challenges limit the scalability and practical applicability of one-step generative models.


In contrast, Sphere Encoder~\cite{yue2026sphere} represents an alternative to diffusion and flow-matching methods. It adopts an encoder–decoder architecture that jointly models reconstruction and generation, and projects latent representations onto a hypersphere instead of relying on a KL divergence objective~\cite{kingma2013auto}. Combined with tailored training objectives, this design enables generation from pure Gaussian noise in only a few steps. With its simplicity, Sphere Encoder offers a conceptually different perspective on generative modeling and demonstrates promising potential for efficient generation.

\begin{figure}[t]
    \centering
    \includegraphics[width=\linewidth]{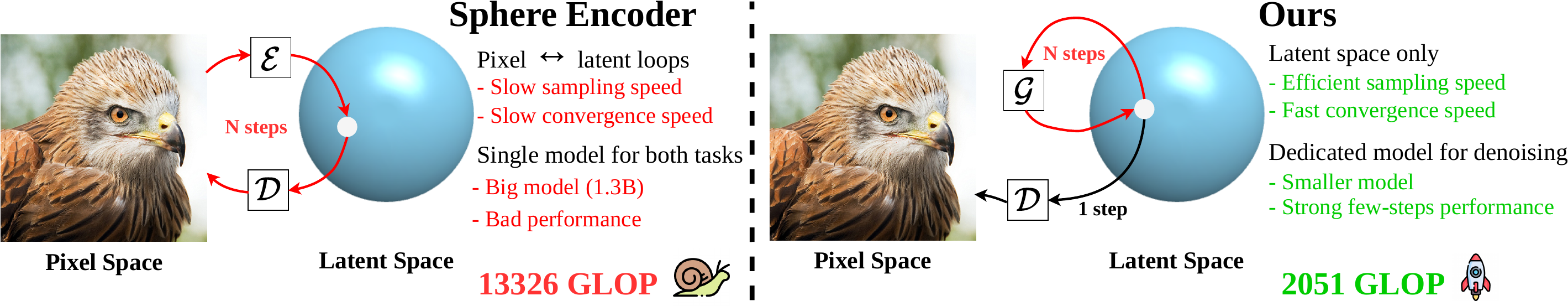}
    \vspace{-0.5cm}
    \caption{
    Comparison between Sphere Encoder and our method. 
{\sethlcolor{blue!15}\hl{\textbf{Left:} Sphere Encoder repeatedly alternates between pixel space and latent space during generation, requiring iterative encoding and decoding steps.}}
{\sethlcolor{red!15}\hl{\textbf{Right:} our method performs denoising entirely in latent space and applies the decoder only once at the end, leading to a simpler and more efficient pipeline.}}
}
    \label{fig:system_comparison}
\end{figure}

However, despite its promise, Sphere Encoder has two important limitations. First, its generation process repeatedly alternates between latent space and pixel space: a latent is decoded into an image, then re-encoded into latent space, and this procedure is repeated for multiple sampling steps. These repeated pixel--latent transitions introduce substantial computational overhead during both training and inference, reducing the practical efficiency of the method. We illustrate the generation process of Sphere Encoder on the left of \Fref{fig:system_comparison}.

Second, Sphere Encoder jointly optimizes reconstruction and generation within a single encoder--decoder architecture. Prior work \cite{yao2025reconstruction, skorokhodov2025improving} suggests that reconstruction and generation favor different representations. Improving reconstruction does not necessarily improve generation, and design choices that benefit one can harm the other. Sphere Encoder itself exhibits this tension. In particular, lower noise levels improve reconstruction quality, but they also reduce hyperspherical coverage and weaken generative performance. As a result, Yue \textit{et al.} \cite{yue2026sphere} must rely on a substantially larger architecture to maintain sample quality.

In this work, we build on the spherical latent perspective of Sphere Encoder while removing its dependence on iterative pixel-space refinement. Our key idea is to perform the entire generative process in latent space. Concretely, we use a pretrained representation autoencoder \cite{zheng2025ditrae} as a fixed image tokenizer and train a separate denoising model directly on spherical latents. During training, supervision is applied purely in latent space. During sampling, the latent is iteratively refined only in latent space, and the decoder is invoked once at the end to map the final latent back to pixels. This decoupled design eliminates repeated encoding and decoding operations, separates reconstruction from generation, and allows each component to specialize in its own role.

Our approach combines the strengths of pretrained latent representations with the efficiency of few-step latent generation. Compared with the original Sphere Encoder, it offers a simpler and more efficient pipeline, while preserving the benefits of spherical latent modeling. Compared with few-step diffusion and flow-based methods, it avoids first-order approximation objectives (e.g., Jacobian-vector products) and instead learns a direct latent denoising process on a structured spherical manifold. As a result, the method is both practically efficient and straightforward to optimize. In \Fref{fig:system_comparison}, we highlight the key difference between Sphere Encoder \cite{yue2026sphere} and our method.

We evaluate our method on Animal-Faces \cite{choi2020stargan}, Oxford-Flowers \cite{nilsback2008automated} and ImageNet-1K \cite{imagenet} datasets following prior work. Experiments show that our latent-only spherical framework substantially improves over Sphere Encoder in both generation quality and inference efficiency, while remaining simple and scalable. In summary, our main contributions are as follows:

\begin{enumerate}[leftmargin=2.5em]
\item We introduce Sphere Latent Encoder, a generative framework operating in spherical latent space that enables efficient training and sampling without repeated transitions between pixel and latent representations. Our method reduces inference cost by approximately 85\% (about 6.5$\times$ fewer FLOPs) compared to Sphere Encoder at the same sampling step.

\item We decouple reconstruction and generation by leveraging a pretrained representation autoencoder \cite{zheng2025ditrae} as a fixed image tokenizer and training a separate denoising model, allowing each component to specialize in its respective task.
\item We conduct extensive experiments on Animal-Faces \cite{choi2020stargan}, Oxford-Flowers \cite{nilsback2008automated} and ImageNet-1K \cite{imagenet}, showing that our method significantly outperforms Sphere Encoder and competes with other baselines in both generation quality and sampling speed.

\end{enumerate}

\section{Related Works}
\vspace{-0.45cm}

\myheading{Efficient Image Generation.} Recent works \cite{geng2025meanflow, geng2025improved, hu2025meanflow, zhang2025alphaflow} explore training few-step flow-matching models from scratch. While these methods achieve strong performance in class-conditional generation, their scalability remains limited due to high training costs and instability in Jacobian–vector product (JVP) computations. These limitations make it challenging to scale this line of work to larger settings. Meanwhile, distillation-based methods \cite{yin2024one, nguyen2024swiftbrush, yin2024improved} remain the most practical solution and are widely adopted in text-to-image generation. Recent work \cite{jiang2025distribution} further extends this paradigm by incorporating reinforcement learning to enable more preference-aware and controllable distillation. Another emerging direction is drifting models \cite{deng2026generative}. Although initial results are promising, their scalability remains limited, as they are prone to mode collapse and often require large batch sizes during training.

\myheading{Sampling in Autoencoder Latent Spaces.} Variational autoencoders \cite{kingma2013auto} match the latent distribution to a Gaussian prior, typically using a unimodal Gaussian posterior. Increasing network depth improves posterior estimation, but does not change the distribution family. Even with full covariance, it remains unimodal and cannot capture multi-modal structure. This mismatch leads to suboptimal inference and weak latent representations. In addition, an overly expressive decoder can lead to posterior collapse, in which the model ignores the latent variables \cite{lai2025principles}. As a result, standard VAEs struggle to produce high-quality samples directly from the prior. Prior work addresses these issues using hierarchical VAEs or diffusion models, which improve generative performance through multi-step refinement \cite{ho2020denoising}. Sphere Encoder \cite{yue2026sphere} addresses this limitation by projecting latent representations onto a hypersphere using RMSNorm \cite{zhang2019root}, encouraging a more uniform latent structure and enabling efficient few-step generation from Gaussian noise.

\myheading{Trade-off between reconstruction and generation.} \cite{yao2025reconstruction} shows that increasing the capacity of the visual tokenizer improves reconstruction quality; however, this makes generation harder, requiring larger denoisers and more training to reach similar sample quality. \cite{yao2025reconstruction, xu2026making} pointed out that this is a trade-off between reconstruction and generation, where better reconstruction does not necessarily lead to better generation. RAE \cite{zheng2025ditrae} further suggests that strong semantic latent representations are crucial for generative performance. At the same time, training primarily with a reconstruction objective can produce weak semantical features, resulting in poor generation quality. Sphere Encoder \cite{yue2026sphere} also observes a trade-off between generation and reconstruction: lower noise improves reconstruction but degrades generation, as the resulting latents do not sufficiently cover the hypersphere. 
\section{Background}
We briefly review the Sphere Encoder~\cite{yue2026sphere} by first introducing the intuition behind its objective design, and then outlining its limitations.

\myheading{Spherification function and additive noise.} Given an input image $\mathbf{x} \in \mathbb{R}^{H \times W \times 3}$, the encoder $\mathcal{E}$ maps the image to a latent representation $\mathbf{z} \in \mathbb{R}^{h \times w \times d}$, while the decoder $\mathcal{D}$ reconstructs the image from the latent where $h = H/P$, $w = W/P$, $d$ is the channel dimension, and $P$ denotes the patch size. A spherification function $\mathcal{F}$ first flattens $\mathbf{z}$ and projects it onto a hypersphere via RMSNorm \cite{zhang2019root}, yielding $\mathbf{v} = \mathcal{F}(\mathbf{z})$. To facilitate training, the latent representation is perturbed with two levels of Gaussian noise, producing $\mathbf{v}_{\text{NOISY}}$ and $\mathbf{v}_{\text{noisy}}$:
\begin{equation}
\label{eq:noise}
\mathbf{v}_{\text{NOISY}} = \mathcal{F}(\mathbf{v} + \sigma \epsilon), \quad
\mathbf{v}_{\text{noisy}} = \mathcal{F}(\mathbf{v} + \sigma_{\text{sub}} \epsilon)
\end{equation}
where $\sigma \sim \mathcal{U}(0, \sigma_{\max})$ and $\sigma_{\text{sub}} \sim \mathcal{U}(0, \sigma \times 0.5)$, where $\sigma_{\max}$ is a predefined hyperparameter. This noise injection encourages the latents to densely populate the hyperspherical space, enabling the decoder to learn a smooth and continuous mapping over the latent manifold rather than relying on a discrete set of embeddings. To supervise the encoder and decoder, three loss functions are introduced, each serving a distinct purpose.

\myheading{Pixel reconstruction loss} ensures faithful reconstruction between pixel and latent spaces. It combines $\ell_1$ and LPIPS losses applied between the reconstructed image from $\mathbf{v}_{\text{noisy}}$ and the input:
\[
\mathcal{L}_{\text{pixel\_recon}} = \mathcal{L}_{\text{L1+LPIPS}} \big( \mathcal{D}(\mathbf{v}_{\text{noisy}}), \mathbf{x} \big).
\]

\myheading{Pixel consistency loss} enforces consistency between reconstructions from different noise levels, here $\mathrm{sg}(\boldsymbol{\cdot})$ denotes the stop-gradient operation:
\[
\mathcal{L}_{\text{pixel\_con}} = \mathcal{L}_{\text{L1+LPIPS}} \big( \mathcal{D}(\mathbf{v}_{\text{NOISY}}), \mathrm{sg}(\mathcal{D}(\mathbf{v}_{\text{noisy}})) \big).
\]

\myheading{Latent consistency loss} encourages the encoder to map distorted or off-manifold reconstructions back to clean latent representations, improving stability during iterative generation. Specifically, the reconstruction from $\mathbf{v}_{\text{NOISY}}$ is re-encoded, and a cosine similarity loss is minimized against the clean latent:
\[
\mathcal{L}_{\text{lat\_con}} = \mathcal{L}_{\text{cos}} \big( \mathbf{v}, \mathcal{E}(\mathcal{D}(\mathbf{v}_{\text{NOISY}})) \big).
\]

\myheading{Total training loss} is the weighted combination of three above-mentioned losses where $\lambda_1, \lambda_2 \text{ and } \lambda_3$ are the weights for each loss.
\[
\mathcal{L}_{\text{total}} = \lambda_{1} \mathcal{L}_{\text{pixel\_recon}} + \lambda_{2} \mathcal{L}_{\text{pixel\_con}} + \lambda_{3} \mathcal{L}_{\text{lat\_con}}.
\]

\myheading{Limitation.} During generation process, Sphere Encoder relies on iteratively applying encoder and decoder. Specifically, a latent code is sampled from a Gaussian distribution, decoded into an image, and then re-encoded back into latent space. This procedure is repeated several times until satisfactory image quality is achieved. While only a few iterations (e.g., 4--8 steps) are typically required, the repeated transitions between latent and pixel spaces introduce significant computational overhead, resulting in inefficient generation. In addition, both prior works~\cite{zheng2025ditrae, yao2025reconstruction} and Sphere Encoder~\cite{yue2026sphere} reveal an inherent trade-off between reconstruction and generation. This suggests that jointly optimizing reconstruction and generation within a single model is suboptimal. Instead, decoupling these objectives into separate stages allows each to be optimized more effectively.

\section{Methodology}
\begin{figure}[H]
\vspace{-0.35cm}
    \centering
    \includegraphics[width=\linewidth]{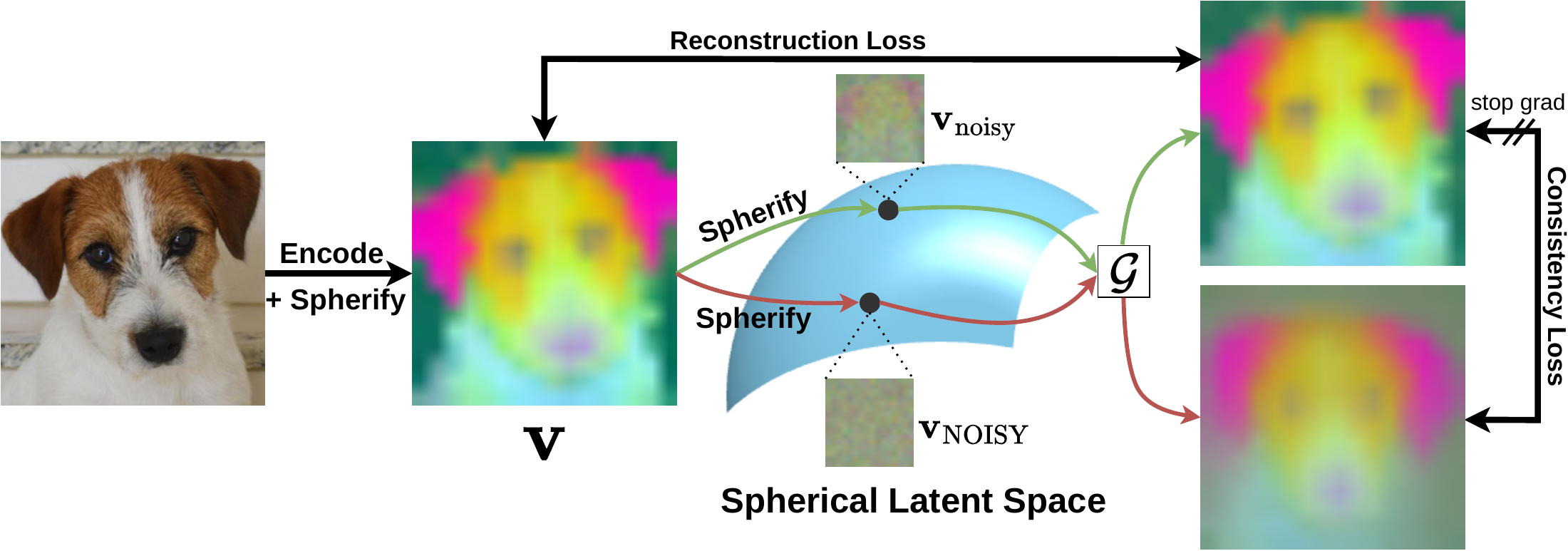}
    \vspace{-0.5cm}
\caption{Overview of our Sphere Latent Encoder framework and training objectives.}
    \label{fig:system}
    \vspace{-0.3cm}
\end{figure}

To address these limitations, we perform generation entirely in latent space using a pretrained encoder–decoder. This eliminates repeated pixel–latent transitions and improves test-time efficiency. Our approach consists of three main components. First, we introduce the overall architecture, which enables efficient few-step generation directly in latent space (\Sref{method:arch}). Next, we present the training objective, designed to learn denoising and reconstruction within the latent domain (\Sref{method:objective}, \Fref{fig:system}). Finally, we describe a simple yet effective sampling procedure that avoids pixel–latent transitions, enabling fast generation (\Sref{method:sampling}, \Aref{alg:sampling}).

\subsection{Denoising model on Spherical Latent Space}
\label{method:arch}
To facilitate training in latent space, we adopt a pretrained representation autoencoder (RAE) from~\cite{zheng2025ditrae}, where the encoder $\mathcal{E}$ is based on DINOv2 and the decoder $\mathcal{D}$ is a ViT-based model. This autoencoder maps an input image $\mathbf{x} \in \mathbb{R}^{256 \times 256 \times 3}$ to a latent representation $\mathbf{z} \in \mathbb{R}^{16 \times 16 \times 768}$. We adopt this design because DINOv2 provides semantically rich representations, while the decoder ensures high-fidelity reconstruction, resulting in a structured latent space that captures both semantic and discriminative information. We then introduce a transformer-based denoising network $\mathcal{G}$ operating in the latent space. Following prior work~\cite{lipman2023flowmatching, geng2025meanflow, yu2025repa}, we adopt the SiT architecture~\cite{ma2024sit}. We denote the Euclidean latent as $\mathbf{z}$ and its spherical projection as $\mathbf{v}$. Specifically, we first corrupt the latent with Gaussian noise, $\mathbf{z} + \sigma \epsilon$, and project it onto a hypersphere via a normalization function $\mathcal{F}$, yielding $\mathbf{v} = \mathcal{F}(\mathbf{z} + \sigma \epsilon)$. The network $\mathcal{G}$ takes $\mathbf{v}$ as input and predicts the clean latent $\mathbf{z}$, which is then decoded by $\mathcal{D}$ to reconstruct the image.

Our formulation is closely related to diffusion models in that it learns to denoise corrupted latent inputs. However, unlike standard diffusion approaches, our method does not condition on the noise level or timestep during either training or inference.

\subsection{Training Objective}
\label{method:objective}
While Sphere Encoder~\cite{yue2026sphere} mainly computes losses in pixel space, which incurs high GPU memory consumption, we instead shift all supervision to latent space for improved efficiency. Similarly, we consider two levels of noisy spherical latents, denoted as $\mathbf{v}_{\text{NOISY}}$ and $\mathbf{v}_{\text{noisy}}$ in \Eref{eq:noise}.

\myheading{Reconstruction loss} in \Eref{eq:recons} is defined as a combination of an $\ell_1$ loss and a cosine similarity loss between the predicted latent $\mathcal{G}(\mathbf{v}_{\text{noisy}})$ and the clean latent $\mathbf{z}$. This loss encourages the denoised spherical latent to align with the clean latent representation, thereby facilitating accurate reconstruction from both clean and noisy spherical latents.
\begin{empheq}[box=\fbox]{equation}
\label{eq:recons}
\mathcal{L}_{\text{recon}} = \| \mathcal{G}(\mathbf{v}_{\text{noisy}}) - \mathbf{z} \|_1 
+ \mathcal{L}_{\text{cosine similarity}}\big( \mathcal{G}(\mathbf{v}_{\text{noisy}}), \mathbf{z} \big)
\end{empheq}

\myheading{Consistency loss} in \Eref{eq:consistent} combines an $\ell_1$ loss and a cosine similarity loss between predictions obtained from two noise levels, $\mathbf{v}_{\text{NOISY}}$ and $\mathbf{v}_{\text{noisy}}$. To enforce consistency across noise scales, we treat the prediction from the lower-noise latent as a fixed target by applying a stop-gradient operator. This encourages the model to align predictions from higher-noise latents with those from lower-noise ones, promoting local smoothness on the hypersphere and ensuring that nearby regions correspond to similar image content.
\begin{empheq}[box=\fbox]{equation}
\label{eq:consistent}
\mathcal{L}_{\text{cons}} = \| \mathcal{G}(\mathbf{v}_{\text{NOISY}}) - \mathrm{sg}\big(\mathcal{G}(\mathbf{v}_{\text{noisy}})\big) \|_1 
+ \mathcal{L}_{\text{cosine}}\big( \mathcal{G}(\mathbf{v}_{\text{NOISY}}), \mathrm{sg}\big(\mathcal{G}(\mathbf{v}_{\text{noisy}})\big) \big)
\end{empheq}
The two losses above are motivated by the pixel reconstruction and pixel consistency objectives in Sphere Encoder~\cite{yue2026sphere}. In our method, however, both objectives are shifted from pixel space to latent space for greater efficiency. With these two losses, our denoising model can generate new samples in just a few steps. 

\subsection{Revisiting Noise Sampling and Training Objectives}
\myheading{Noise Distribution.}
Instead of adopting the sampling strategy used in Sphere Encoder~\cite{yue2026sphere}, we propose an alternative approach for sampling $\sigma$ and $\sigma_{sub}$. While we retain the constraint $\sigma \geq \sigma_{sub}$, we modify the underlying sampling procedure.
Concretely, for each data point, we draw two independent samples from a logit-normal distribution \cite{esser2024scaling, geng2025meanflow}. This distribution is obtained by first sampling $z \sim \mathcal{N}(\mu, \sigma^2)$ and then applying the logistic function to map $z$ into the interval $(0,1)$. We assign the larger sample to $\sigma$ and the smaller one to $\sigma_{sub}$, inducing a joint distribution over $(\sigma, \sigma_{sub})$ that differs from the original Sphere Encoder formulation. Sphere Encoder~\cite{yue2026sphere} jointly optimizes reconstruction and generation within a single framework, so its noise schedule must balance hyperspherical coverage for generation against reconstruction fidelity. This coupling leads to a relatively conservative noise sampling strategy. In contrast, our framework decouples reconstruction from generation using a fixed pretrained autoencoder and a separate latent denoising model, allowing us to adopt more aggressive noise schedules that are better suited to few-step sampling.



\myheading{Removing Latent Consistency Loss.} Sphere Encoder \cite{yue2026sphere} introduces a latent consistency loss to encourage consistency between decoder and encoder, but this objective adds notable computational overhead during training. We explore a simplified variant of this loss in latent space (see Appendix Section~\ref{appendix:additional}) and observe that removing it improves both performance and training efficiency. Our approach instead uses a dedicated denoising model, avoiding the need to impose such a consistency constraint and eliminating the latent consistency loss. Our denoising model is trained over a broader and smoother noisy latent space on the hypersphere, leading to stronger denoising performance without relying on this objective.

\begin{wrapfigure}{r}{0.5\linewidth}
\vspace{-0.7cm}
\centering
\begin{minipage}{0.98\linewidth}

\begin{algorithm}[H]
\caption{Iterative Latent Sampling with Classifier-Free Guidance}
\label{alg:sampling}
\small
\begin{algorithmic}[1]

\Require steps $T$, $\sigma_{max}$, guidance $\omega$, label $y$

\State $\mathbf{z} \sim \mathcal{N}(0, I)$ \label{alg:init}
\State $\boldsymbol{\epsilon} \sim \mathcal{N}(0, I)$

\For{$t = 0$ to $T-1$} 
\State $\mathbf{v} \gets \mathcal{F}(\mathbf{z})$ \label{alg:proj} 

\State $\mathbf{z}_{\text{guided}} \gets \mathcal{G}(\mathbf{v}, \varnothing) + \omega \left( \mathcal{G}(\mathbf{v}, y) - \mathcal{G}(\mathbf{v}, \varnothing) \right)$ \label{alg:cfg} 

\If{$t = T - 1$}
\State break
\EndIf

\State $r \gets \left(1 - \frac{t+1}{T} \right)^{\gamma}$ \label{alg:decay} 

\State \textit{// Spherify and re-noise} \State $\mathbf{v}' \gets \mathcal{F}(\mathbf{z}_{\text{guided}})$ \label{alg:reproj} 

\State $\mathbf{z} \gets \mathbf{v}' + \boldsymbol{\epsilon} \cdot \boldsymbol{\sigma_{max}} \cdot r$ \label{alg:renoise} \EndFor

\State $\mathbf{x} \gets \texttt{decode}(\mathbf{z}_{\text{guided}})$
\State \Return $\mathbf{x}$

\end{algorithmic}
\end{algorithm}

\end{minipage}
\vspace{-1.5em}
\end{wrapfigure}

\subsection{Sampling in Spherical Latent Space}\label{method:sampling}
With the proposed loss functions, our model enables multi-step sampling entirely in latent space (see \Aref{alg:sampling}). We begin by sampling a latent from a Gaussian distribution (Alg.~\ref{alg:sampling}, line~\ref{alg:init}). At each iteration, the latent is projected onto the hypersphere (Alg.~\ref{alg:sampling}, line~\ref{alg:proj}) and passed through the denoising model $\mathcal{G}$ to obtain a refined latent $\mathbf{z}$ via classifier-free guidance (Alg.~\ref{alg:sampling}, line~\ref{alg:cfg}).

This guided latent is then re-projected onto the hypersphere (Alg.~\ref{alg:sampling}, line~\ref{alg:reproj}), perturbed with noise (Alg.~\ref{alg:sampling}, line~\ref{alg:renoise}), and used as input for the next iteration. We progressively decrease the noise magnitude over iterations (Alg.~\ref{alg:sampling}, line~\ref{alg:decay}), following Sphere Encoder \cite{yue2026sphere}, so that smaller perturbations are introduced as sampling proceeds. This iterative procedure alternates between projection, denoising, and re-noising, allowing the model to gradually refine the latent and converge to high-quality samples. For detailed hyperparameter settings, please refer to Appendix Section \ref{appendix:noise_perturbation}.
\section{Experiments}
\subsection{Few-step Image Generation}
\myheading{Evaluation protocol.} We evaluate generation quality using the Fréchet Inception Distance (FID)~\cite{heusel2017gans}. The metric is computed on 50{,}000 randomly sampled images; for detailed calculation, please refer to Appendix Section \ref{appendix:metric_calc}. For class-conditional generation, we adopt a balanced sampling strategy by drawing an equal number of samples from each class, following RAE \cite{zheng2025ditrae}. We conduct experiments on Animal-Faces ~\cite{choi2020stargan}, Oxford-Flowers~\cite{nilsback2008automated}, with all images resized to $256 \times 256$, and ImageNet-1K~\cite{imagenet}, where images are first center-cropped to a square and then resized to $256 \times 256$. We apply minimal data augmentation, consisting only of horizontal flipping with a probability of $0.5$. On ImageNet-1K, we additionally provide the result on Normalized Fréchet Distance Ratios (FDr$^6$) \cite{yang2604representation}. For more implementation details, please refer to Appendix Section~\ref{appendix:implementation}.

\begin{table}[H]
\vspace{-0.4cm}
\centering
\caption{Quantitative comparison (gFID $\downarrow$) and Computational cost (GFLOP $\downarrow$) on Animal-Faces \cite{choi2020stargan} and Oxford Flowers \cite{nilsback2008automated} with different numbers of sampling steps (\textcolor{blue}{2}, \textcolor{green!60!black}{4}, \textcolor{red}{6}).}

\vspace{0.1cm}

{\fontsize{8}{9}\selectfont
\setlength{\tabcolsep}{8pt}
\renewcommand{\arraystretch}{0.95}
\begin{tabular}{l c >{\columncolor{blue!8}}c >{\columncolor{green!8}}c >{\columncolor{red!8}}c c c >{\columncolor{blue!8}}c >{\columncolor{green!8}}c >{\columncolor{red!8}}c}
\toprule
& \multicolumn{4}{c}{\textbf{Animal-Faces \cite{choi2020stargan}}} 
& 
& \multicolumn{4}{c}{\textbf{Oxford-Flowers \cite{nilsback2008automated}}} \\
\cmidrule(lr){2-5} \cmidrule(lr){7-10}
\textbf{Metric / Method}
& Params & 2 & 4 & 6
& 
& Params & 2 & 4 & 6 \\
\midrule
\rowcolor{gray!12}
\multicolumn{10}{l}{\textit{FID (gFID $\downarrow$)}} \\
Sphere Encoder \cite{yue2026sphere} 
& 642M & 19.29 & 18.23 & 17.97 
& 
& 948M & 16.60 & 12.96 & 12.26 \\
Ours 
& 130M & 10.63 & 6.89 & \textbf{6.18} 
& 
& 130M & 12.22  & 8.61 &  \textbf{7.85} \\

\midrule

\rowcolor{gray!12}
\multicolumn{10}{l}{GFLOPs $\downarrow$} \\
Sphere Encoder \cite{yue2026sphere} 
& 642M & 1965 & 4554 & 7144
& 
& 948M & 3932 & 9118 & 14300 \\
Ours 
& 130M & 302 & 390 & 478
& 
& 130M & 390 & 567 & 743 \\

\bottomrule
\end{tabular}
}

\label{tab:animal_faces_flowers}
\vspace{-0.5cm}
\end{table}

\myheading{Quantitative results on small datasets.} We compare our method against Sphere Encoder~\cite{yue2026sphere}. To ensure a fair comparison, we report two variants of our approach: one with classifier-free guidance for Oxford-Flowers, and one without classifier-free guidance for Animal-Faces. Quantitative results are presented in Table~\ref{tab:animal_faces_flowers}. Our method significantly outperforms prior work by a large margin, while using a smaller architecture and operating entirely in the latent space during sampling, resulting in substantially lower computational cost in terms of GFLOPs across all sampling steps compared to Sphere Encoder (see Appendix Section~\ref{appendix:flop_calc} for details).

\begin{figure}[H]
    \centering
    \begin{subfigure}[t]{0.4925\linewidth}
        \centering
        \includegraphics[width=\linewidth]{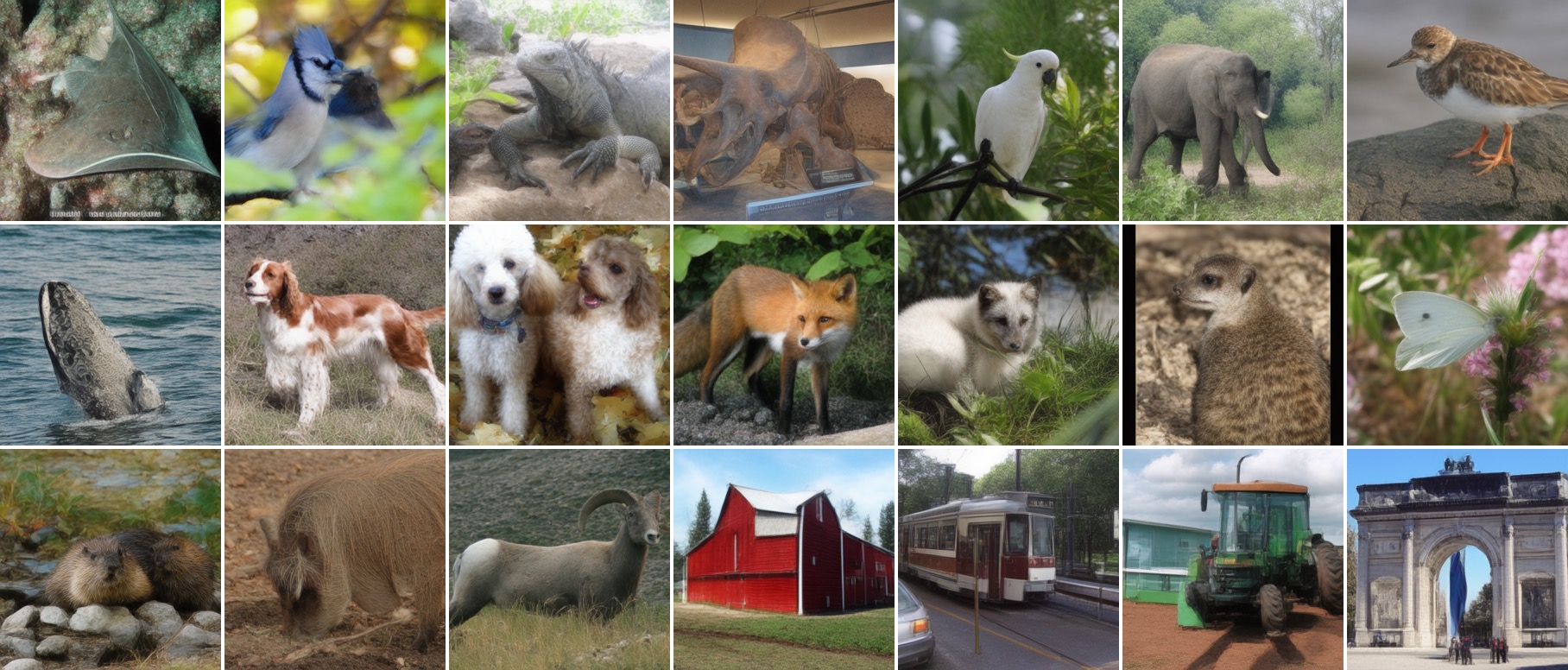}
        \caption{Sphere Encoder \cite{yue2026sphere}}
        \label{fig:method1}
    \end{subfigure}
    \hfill
    \begin{subfigure}[t]{0.4925\linewidth}
        \centering
        \includegraphics[width=\linewidth]{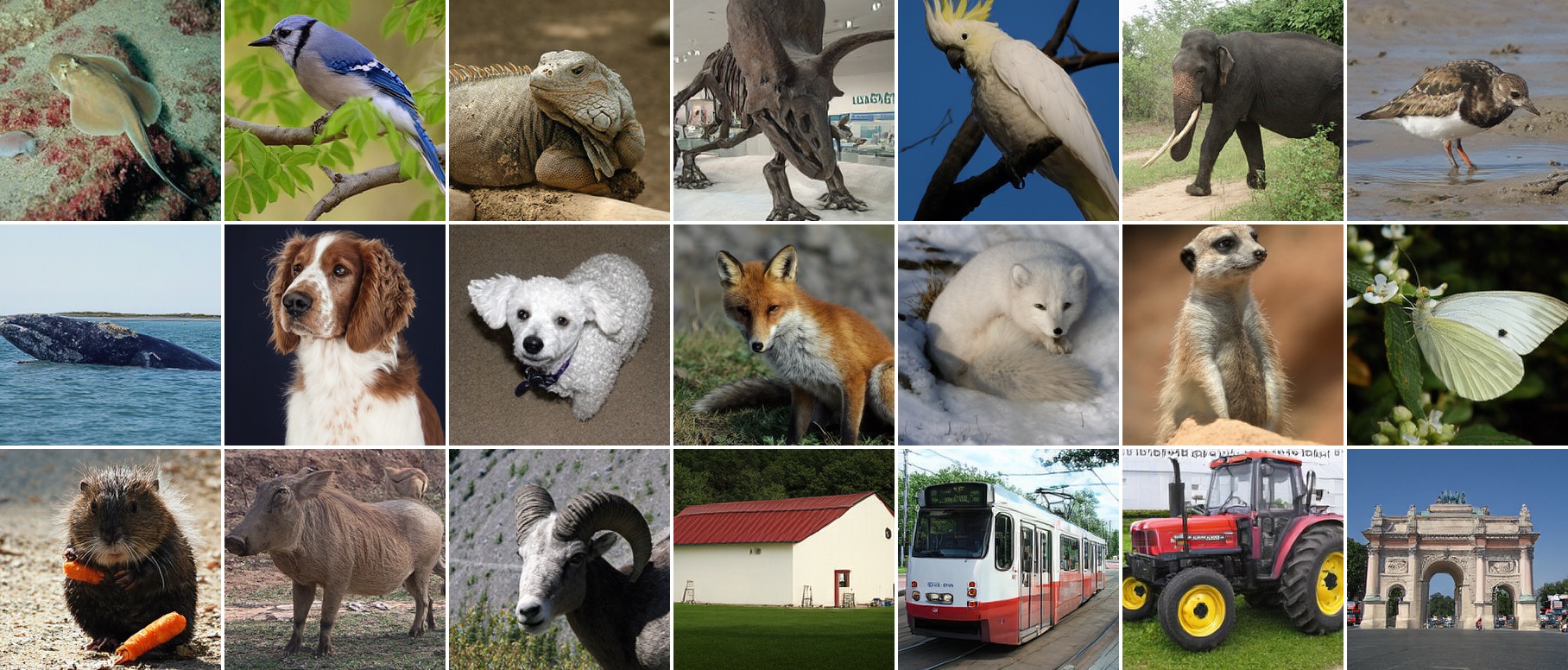}
        \caption{Ours}
        \label{fig:method2}
    \end{subfigure}
\caption{Qualitative comparison between Sphere Encoder~\cite{yue2026sphere} and our method on 4-steps image generation on ImageNet-1K~\cite{imagenet}.}
    \label{fig:comparison}
    \vspace{-0.6cm}
\end{figure}

\myheading{Quantitative results on ImageNet-1K.} We compare our method with the sphere-based generators Sphere Encoder~\cite{yue2026sphere} and Sphere Encoder 2~\cite{yue2026sphere2}, representative multi-step diffusion and flow matching models~\cite{ma2024sit, yu2025repa, wu2025reg, yao2025reconstruction, leng2025repae, liu2026geometric, zheng2025ditrae}, and recent few-step generation approaches~\cite{geng2025meanflow, zhang2025alphaflow, deng2026generative, geng2025improved, yang2604representation}. Besides FID, we report FDr$^6$~\cite{yang2604representation}, which averages normalized Fr\'echet distances over six representation spaces and is therefore less sensitive to the biases of the Inception feature space alone. Table~\ref{tab:system_comparison} summarizes the results. Within the sphere-model family, our method delivers the strongest performance on both metrics. Under the same sampling budget of $4\times2$ NFE, it reduces FID from $4.02$ to $\mathbf{2.25}$ and FDr$^6$ from $10.44$ to $4.36$ relative to Sphere Encoder; increasing the budget to $6\times2$ NFE further lowers them to $\mathbf{2.11}$ and $\mathbf{4.08}$, respectively and even outperforming the concurrent Sphere Encoder 2 as well. These quantitative gains are consistent with the qualitative comparison in \Fref{fig:comparison}, where our samples exhibit sharper structures and finer details. Beyond the sphere family, the strongest multi-step diffusion and flow-based methods still attain lower FID, but only at the cost of orders of magnitude more function evaluations ($250\times2$). Even so, our method already surpasses several of these multi-step models in FDr$^6$, including SiT-XL/2 ($8.44$), REPA ($5.45$), REG ($4.64$), and LightningDiT ($4.57$). Our method also remains competitive with recent 1-NFE diffusion/flow-based approaches. Finally, this efficiency extends to actual compute: as reported in \Tref{tab:appendix_flop}, our method requires substantially fewer FLOPs than Sphere Encoder, even when using more sampling steps. More details about FLOP calculation is shown in Appendix Section~\ref{appendix:additional}.

\begin{table*}[t]
\centering
\caption{\textbf{Class-conditional image generation on ImageNet-1K $256\times256$}~\cite{imagenet}.
Lower FID and FDr$^6$ are better ($\downarrow$); higher IS, Precision, and Recall are better ($\uparrow$).
NFE $\times2$ denotes classifier-free guidance. FDr$^6$ of prior methods is taken from~\cite{yang2604representation}.
\textbf{Bold} marks the best result among sphere models.}
\label{tab:system_comparison}
\footnotesize
\setlength{\tabcolsep}{8pt}
\setlength{\aboverulesep}{0pt}
\setlength{\belowrulesep}{0pt}
\renewcommand{\arraystretch}{1.2}

\resizebox{\linewidth}{!}{%
\begin{tabular}{l r c c c c c c}
\toprule
\textbf{Method} & \textbf{\#Params} & \textbf{NFE} & \textbf{FID}\,$\downarrow$ & \textbf{IS}\,$\uparrow$ & \textbf{Prec.}\,$\uparrow$ & \textbf{Rec.}\,$\uparrow$ & \textbf{FDr$^6$}\,$\downarrow$ \\
\midrule

\grpfade{Multi-step diffusion / flow}
\bl{SiT-XL/2~\cite{ma2024sit}}                      & \bl{675M} & \bl{$250\times2$} & \bl{2.06} & \bl{277.5} & \bl{0.83} & \bl{0.59} & \bl{8.44} \\
\bl{SiT-XL/2 + REPA~\cite{yu2025repa}}              & \bl{675M} & \bl{$250\times2$} & \bl{1.42} & \bl{305.7} & \bl{0.80} & \bl{0.65} & \bl{5.45} \\
\bl{SiT-XL/2 + REG~\cite{wu2025reg}}                & \bl{675M} & \bl{$250\times2$} & \bl{1.36} & \bl{299.4} & \bl{0.77} & \bl{0.66} & \bl{4.64} \\
\bl{LightningDiT-XL/2~\cite{yao2025reconstruction}} & \bl{675M} & \bl{$250\times2$} & \bl{1.35} & \bl{295.3} & \bl{0.79} & \bl{0.65} & \bl{4.57} \\
\bl{REPA-E~\cite{leng2025repae}}                    & \bl{675M} & \bl{$250\times2$} & \bl{1.15} & \bl{304.0} & \bl{0.79} & \bl{0.66} & \bl{3.04} \\
\bl{GAE~\cite{liu2026geometric}}                    & \bl{675M} & \bl{$250\times2$} & \bl{1.13} & \bl{294.9} & \bl{0.79} & \bl{0.67} & \bl{1.97}   \\
\bl{RAE + DiT-XL~\cite{zheng2025ditrae}}            & \bl{839M} & \bl{$50\times2$}  & \bl{1.13} & \bl{262.6} & \bl{0.78} & \bl{0.67} & \bl{3.26} \\
\midrule

\grpfade{1-NFE diffusion / flow from scratch}
\bl{MeanFlow-XL/2~\cite{geng2025meanflow}}            & \bl{676M} & \bl{1} & \bl{3.43} & \bl{216.4} & \bl{0.75} & \bl{0.59} & \bl{12.30}    \\
\bl{$\alpha$-Flow-XL/2+~\cite{zhang2025alphaflow}}    & \bl{676M} & \bl{1} & \bl{2.58} & \bl{242.4} & \bl{0.76} & \bl{0.61} & \bl{14.14}    \\
\bl{Drift-L~\cite{deng2026generative}}                & \bl{463M} & \bl{1} & \bl{1.43} & \bl{305.8} & \bl{0.81} & \bl{0.60} & \bl{10.51} \\
\bl{iMF-XL/2~\cite{geng2025improved}}                 & \bl{610M} & \bl{1} & \bl{1.72} & \bl{282.0} & \bl{0.78} & \bl{0.62} & \bl{8.39}  \\
\bl{iMF-XL/2 + FD-loss~\cite{yang2604representation}} & \bl{610M} & \bl{1} & \bl{0.76} & \bl{301.3} & \bl{0.77} & \bl{0.67} & \bl{2.45}  \\
\midrule

\grp{Sphere models}
Sphere Encoder~\cite{yue2026sphere}              & 1.3B & $4\times2$ & 4.02 & 265.9 & 0.78 & 0.58 & 10.44 \\
Sphere Encoder 2~\cite{yue2026sphere2}           & 487M & $4\times2$ & 2.86 & --    & --   & --   & 4.92  \\
\rowcolor{oursbg} \textbf{Ours-XL/1}             & 675M & $4\times2$ & 2.25 & \textbf{308.5} & 0.78 & 0.61 & 4.36 \\
\rowcolor{oursbg} \textbf{Ours-XL/1}             & 675M & $6\times2$ & \textbf{2.11} & 293.8 & \textbf{0.78} & \textbf{0.62} & \textbf{4.08} \\
\bottomrule
\end{tabular}%
}
\end{table*}

\begin{table}[H]
\vspace{-0.3cm}
\centering
\caption{Computational cost analysis on ImageNet-1K. (a) Total sampling FLOPs compared with Sphere Encoder~\cite{yue2026sphere}. (b) Per-forward-pass parameters and FLOPs of each component used during sampling.}
\label{tab:flops}
\footnotesize
\setlength{\tabcolsep}{6pt}
\renewcommand{\arraystretch}{1.15}

\subcaptionbox{Total sampling cost.\label{tab:appendix_flop}}{%
\begin{tabular}{lcr}
\toprule
\textbf{Method} & \textbf{NFE} & \textbf{GFLOPs}\,$\downarrow$ \\
\midrule
Sphere Encoder & $4\times2$ & 13{,}326 \\
\midrule
\multirow{2}{*}{Ours} & $4\times2$ & \cellcolor{gray!20}\textbf{2{,}051} \\
                      & $6\times2$ & 2{,}969 \\
\bottomrule
\end{tabular}}%
\hspace{2em}%
\subcaptionbox{Per-component cost (one forward pass).\label{tab:imagenet_component_flops}}{%
\begin{tabular}{llrr}
\toprule
\textbf{Method} & \textbf{Component} & \textbf{Params} & \textbf{GFLOPs} \\
\midrule
\multirow{2}{*}{Sphere Encoder} & Encoder         & 682M & 918 \\
                                & Decoder         & 702M & 977 \\
\midrule
\multirow{2}{*}{Ours}           & Denoising model & 674M & 230 \\
                                & Decoder         & 415M & 213 \\
\bottomrule
\end{tabular}}
\vspace{-0.3cm}
\end{table}

\subsection{Ablation Studies}
We conduct a series of ablation studies to systematically analyze the key design choices of our method using the SiT-B/1 architecture~\cite{ma2024sit}. For quantitative evaluation, we construct an ImageNet-100 subset by random sampling 100 classes from ImageNet-1K~\cite{imagenet}, which provides a well-representative yet computationally efficient benchmark. To evaluate the impact of the latent autoencoder, we use only the Animal-Faces dataset~\cite{choi2020stargan}, as it provides a simpler and more controlled setting and for inference step and effect of pretrained model, we evaluate on ImageNet-1K~\cite{imagenet} for completeness. We examine five main factors: \textcolor{red!70!black}{Noise Distribution}, \textcolor{green!50!black}{Training Losses}, \textcolor{purple!70!black}{Latent Projection}, \textcolor{brown!70!black}{Inference Steps}, \textcolor{teal!70!black}{Autoencoder} and \textcolor{cyan!70!black}{Pretrained model}.

\myheading{\textcolor{red!70!black}{Noise Distribution.}}
We analyze the effect of different noise distributions during training by comparing our variants with the baseline method~\cite{yue2026sphere}. Specifically, we evaluate Uniform, LogNorm$(-0.4, 1.0)$, and LogNorm$(+0.4, 1.0)$, and report results on ImageNet-100 in \Tref{tab:ablation}. All proposed distributions outperform the baseline, yielding at least a 10.0\% improvement in FID. We attribute these gains to the use of a dedicated denoising model, which can better handle higher noise levels during training and thus improves generation quality. Among the variants, LogNorm$(-0.4, 1.0)$ achieves a 13.5\% improvement, while LogNorm$(+0.4, 1.0)$ performs best with a 17.4\% gain. We adopt the latter in all subsequent experiments.

\myheading{\textcolor{green!50!black}{Training Objectives.}}
We analyze the impact of the proposed training objectives. Introducing the consistency loss yields a substantial improvement of 40.8\% over the reconstruction-only baseline, as it enforces proximity between noisy latent pairs and encourages the denoising network to produce outputs that are both clean and semantically aligned with the underlying latent representation. We further incorporate a latent consistency loss following the formulation of Sphere Encoder~\cite{yue2026sphere} (see Appendix~\ref{appendix:additional} for details); however, this additional constraint degrades performance, resulting in only a 46.3\% improvement. We hypothesize that, since the model is already trained under strong noise conditions, the extra regularization imposes unnecessary constraints and increases optimization difficulty. Overall, combining reconstruction and consistency losses achieves the best performance, with a 47.8\% improvement.

\myheading{\textcolor{purple!70!black}{Latent Projection.}} 
We investigate the importance of the spherify function by replacing RMSNorm \cite{zhang2019root} with an identity mapping. This modification leads to a dramatic performance drop, with outputs becoming incoherent, increasing FID from 4.68 to 89.68 in \Tref{tab:ablation}. The results suggest that the latent denoiser and representation autoencoder cannot function effectively out-of-the-box, and instead require careful design within a spherical latent space training framework.

\begin{figure}[H]
\centering
\begin{subfigure}[t]{0.59\linewidth}
\centering

\raisebox{0.54cm}{\rotatebox[origin=c]{90}{\small 2-\emph{nfe}}}
\includegraphics[width=0.9\linewidth]{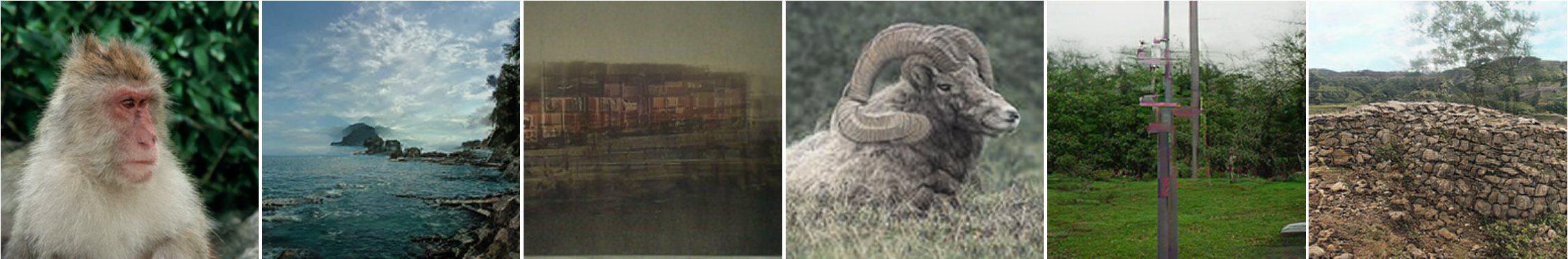}%

\raisebox{0.54cm}{\rotatebox[origin=c]{90}{\small 4-\emph{nfe}}}
\includegraphics[width=0.9\linewidth]{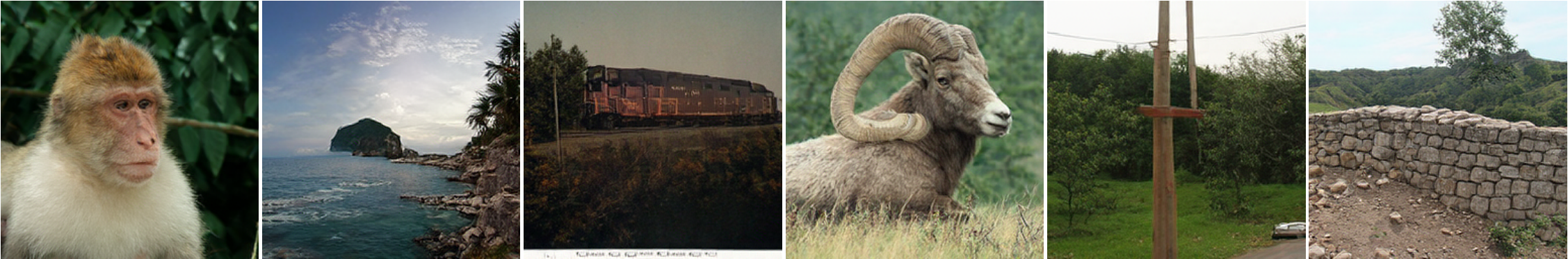}%

\raisebox{0.54cm}{\rotatebox[origin=c]{90}{\small 8-\emph{nfe}}}
\includegraphics[width=0.9\linewidth]{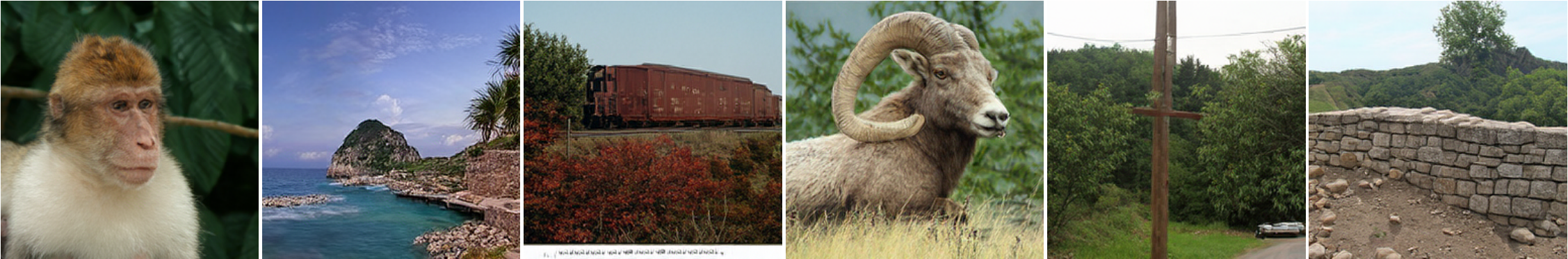}%

\caption{Effect of inference steps on ImageNet-100.}
\label{fig:inference_steps}
\end{subfigure}
\hspace{-0.15cm}
\begin{subfigure}[t]{0.39398\linewidth}
\centering
\includegraphics[width=0.9\linewidth]{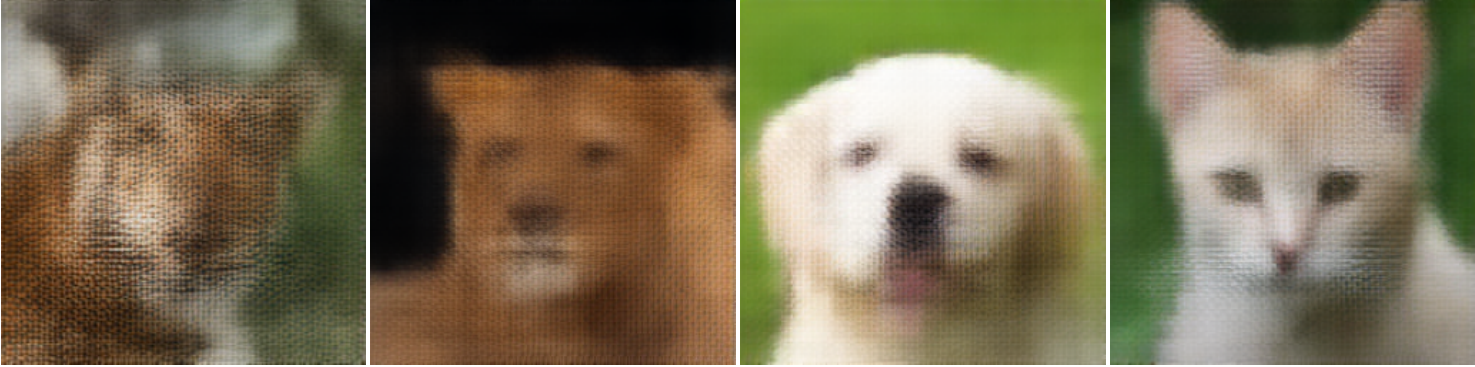}%
\hspace{0.01cm}
\raisebox{0.535cm}{\rotatebox[origin=c]{270}{\fontsize{7}{6}\selectfont FLUX \cite{flux2024}}}

\includegraphics[width=0.9\linewidth]{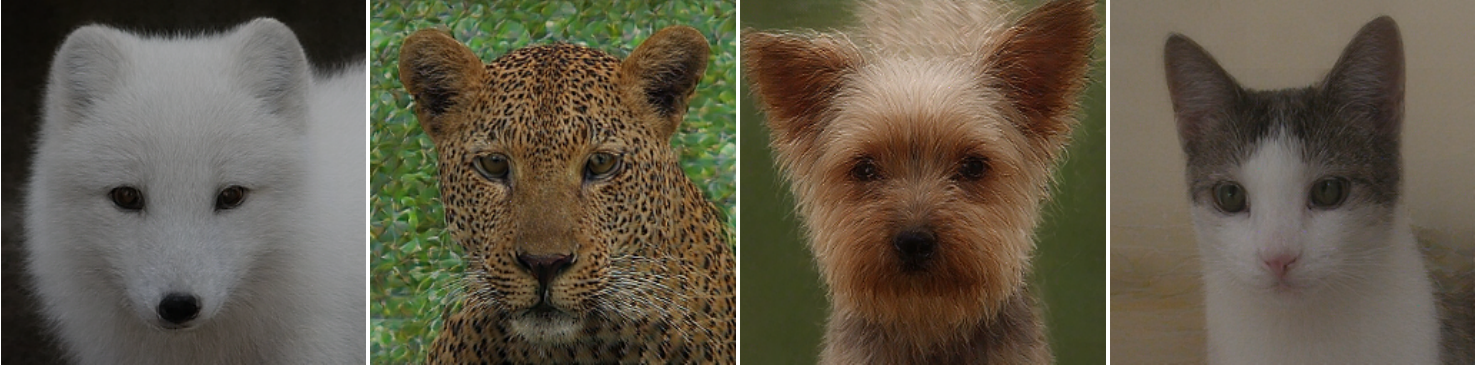}%
\hspace{0.01cm}
\raisebox{0.535cm}{\rotatebox[origin=c]{270}{\fontsize{7}{6}\selectfont GAE\cite{liu2026geometric}}}

\includegraphics[width=0.9\linewidth]{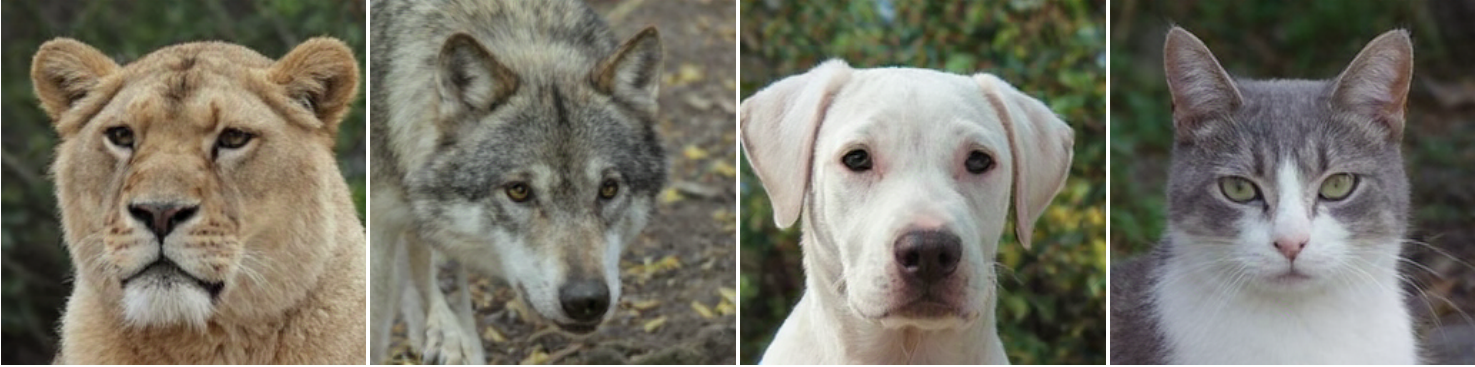}%
\hspace{0.01cm}
\raisebox{0.535cm}{\rotatebox[origin=c]{270}{\fontsize{7}{6}\selectfont RAE\cite{zheng2025ditrae}}}

\caption{Effect of autoencoders on Animal-Faces.}
\label{fig:autoencoder_ablation}
\end{subfigure}

\caption{Increasing inference steps improves fidelity (a), while stronger semantic representations and larger latent dimensionality yield sharper samples (b).}
\label{fig:combined_qualitative}
\vspace{-0.4cm}
\end{figure}

\begin{figure}[H]
    \centering
    \begin{subfigure}[t]{0.49\linewidth}
        \centering
        \includegraphics[width=\linewidth]{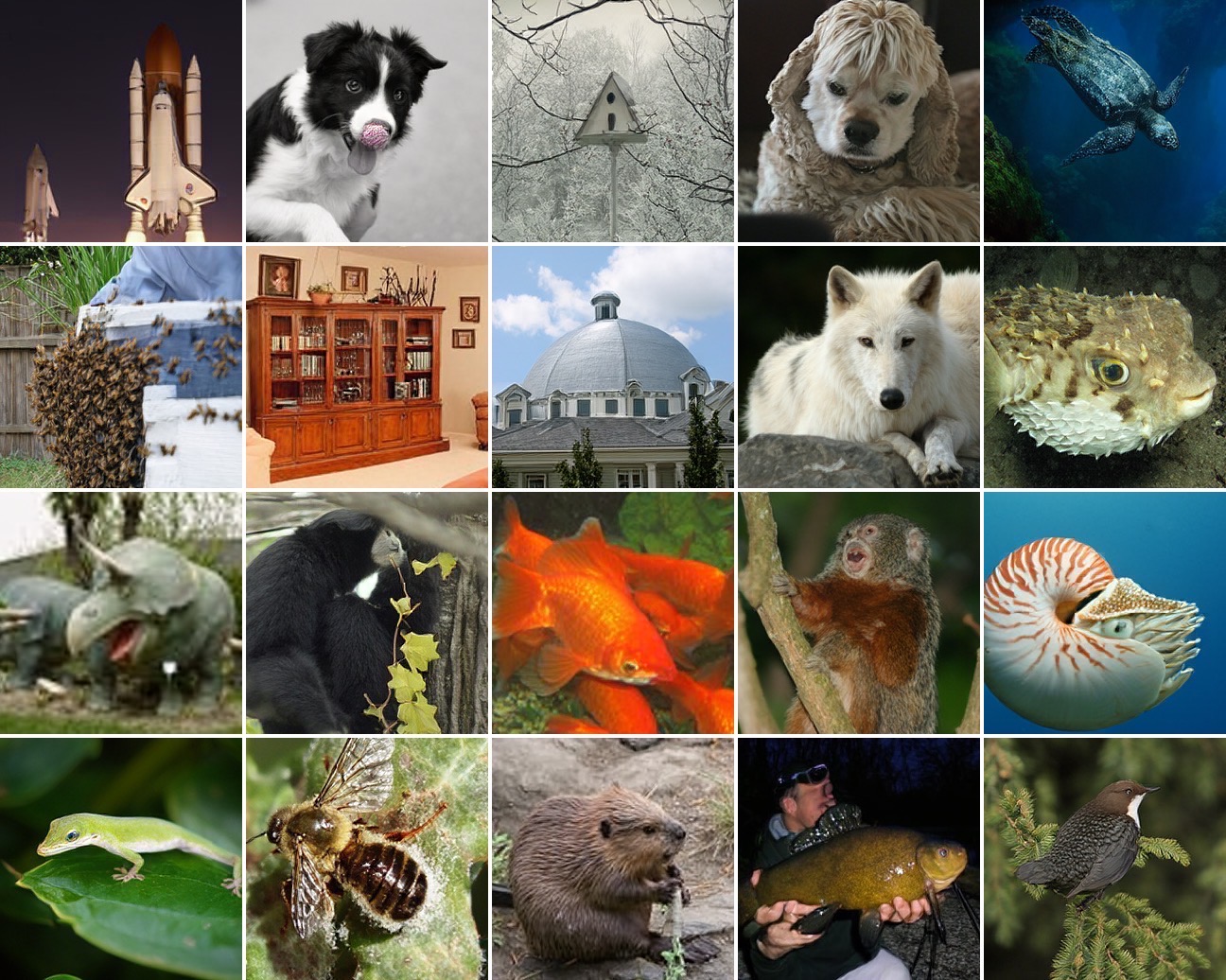}
        \caption{4-\emph{nfe}}
        \label{fig:imagenet_1}
    \end{subfigure}
    \hfill
    \begin{subfigure}[t]{0.49\linewidth}
        \centering
        \includegraphics[width=\linewidth]{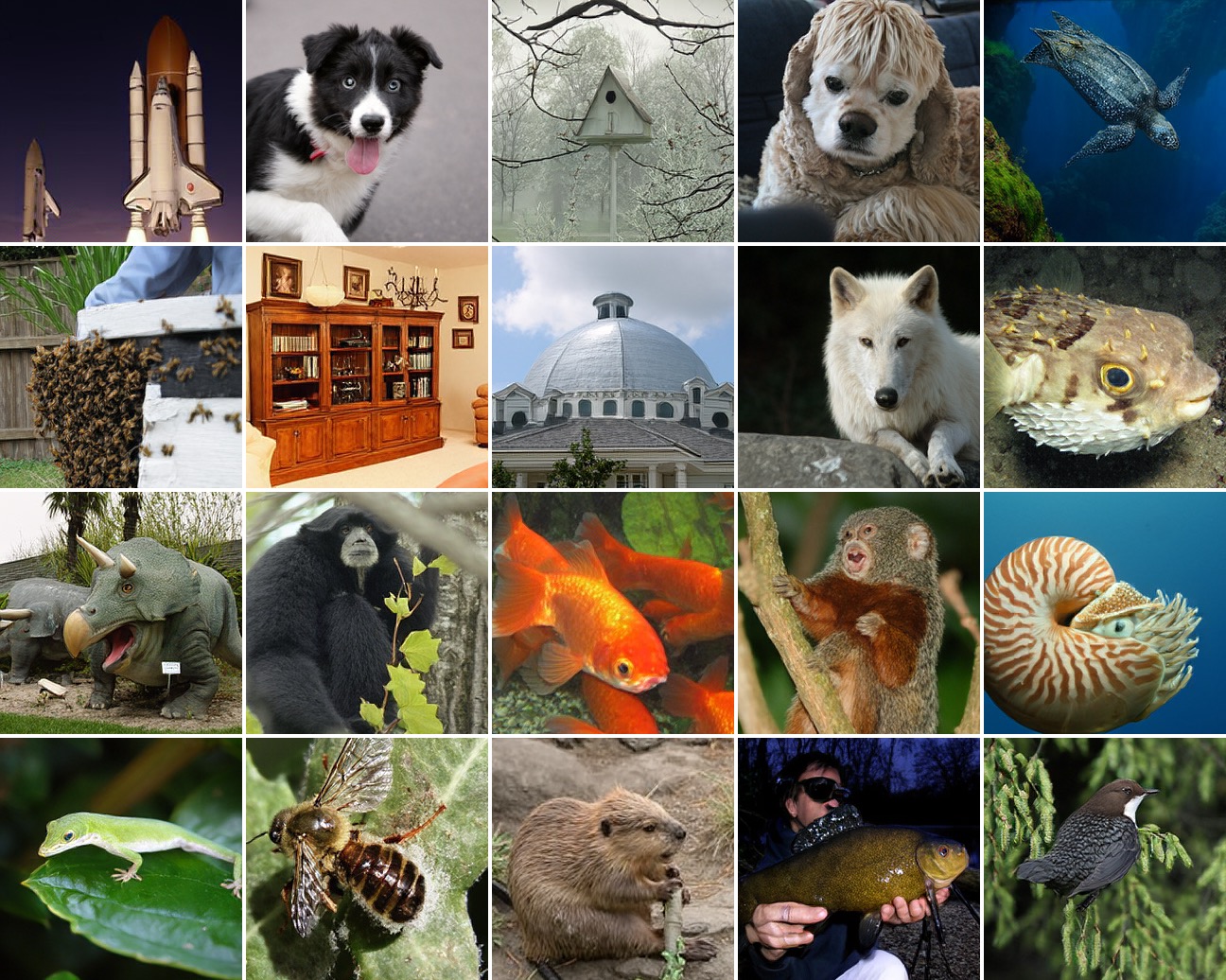}
        \caption{6-\emph{nfe}}
        \label{fig:imagenet_2}
    \end{subfigure}
\caption{Qualitative comparison between different number of sampling steps on ImageNet-1K~\cite{imagenet}.}
    \label{fig:step_comparison}
\end{figure}

\myheading{\textcolor{brown!70!black}{Inference Steps.}}
We analyze the effect of the number of inference steps at test time. As shown in Table~\ref{tab:ablation}, \Fref{fig:inference_steps} and \Fref{fig:step_comparison}, generation quality improves substantially when increasing the number of steps from 2 to 4, with diminishing gains beyond 4 steps and near saturation at 8 steps. These results suggest that only a small number of inference steps is needed to obtain high-quality samples.

\myheading{\textcolor{cyan!70!black}{Pretrained Model.}}
We investigate the effectiveness of using a pretrained representation encoder (DINOv2~\cite{oquab2023dinov2}) for both Sphere Encoder and our method on ImageNet-1K~\cite{imagenet}. Both models are initialized with DINOv2 and trained for 150 epochs using four sampling steps. In Table~\ref{tab:ablation}, our method significantly outperforms Sphere Encoder, reducing FID from 18.50 to 2.60, corresponding to an 85.9\% improvement. Moreover, our model achieves performance close to that of our best model trained for 600 epochs, demonstrating its fast convergence and training efficiency.

\myheading{\textcolor{teal!70!black}{Choice of Latent Autoencoder.}}
We study the impact of the latent autoencoder using the Animal-Faces dataset~\cite{choi2020stargan}. As shown in \Tref{tab:ablation} and \Fref{fig:autoencoder_ablation}, performance strongly depends on the quality and dimensionality of the latent representation. FLUX-VAE~\cite{flux2024} performs worst, with a FID of 172.25, due to its low latent dimensionality $(d=16)$, which limits capacity and leads to poorly separated embeddings. GAE~\cite{liu2026geometric} improves the representation by incorporating vision foundation models during training, achieving 23.26, but remains constrained by its compact latent space $(d=32)$. In contrast, RAE~\cite{zheng2025ditrae} achieves the best results, with 10.63, by combining semantically rich features (e.g., DINOv2~\cite{oquab2023dinov2}) with a high-dimensional latent space $(d=768)$, providing both strong representation quality and sufficient capacity for effective denoising.

\begin{table}[h]
\centering

\caption{Ablation studies for few-step image generation on ImageNet-100, ImageNet-1K, and Animal-Faces (FID $\downarrow$). Improvement is reported relative to the first setting within each factor. 
\textcolor{blue}{R}: reconstruction loss; 
\textcolor{orange!80!black}{C}: consistency loss; 
\textcolor{purple}{L}: latent consistency loss.}
\vspace{0.2cm}

{\fontsize{8}{10}\selectfont
\setlength{\tabcolsep}{6pt}
\renewcommand{\arraystretch}{0.85}

\begin{tabular}{lllcc}
\toprule
\fontsize{9}{10}\selectfont
\textbf{Dataset} & \fontsize{9}{10}\selectfont\textbf{Factor} & \fontsize{9}{10}\selectfont\textbf{Setting} & \fontsize{9}{10}\selectfont\textbf{FID} & \fontsize{9}{10}\selectfont\textbf{Impr. (\%)} \\

\midrule

\multirow{10}{*}{\fontsize{9}{10}\selectfont ImageNet-100}

& \multirow{4}{*}{\textcolor{red!70!black}{Noise Distribution}}
& Baseline & 6.43 & -- \\
& & Uniform & 5.79 & {\color{green!60!black}10.0} \\
& & LogNorm $(-0.4, 1.0)$ & 5.56 & {\color{green!60!black}13.5} \\
& & LogNorm $(+0.4, 1.0)$ & \textbf{5.31} & {\color{green!60!black}\textbf{17.4}} \\

\cmidrule(lr){2-5}

& \multirow{4}{*}{\textcolor{green!50!black}{Training Losses}}
& \textcolor{blue}{R} & 8.97 & -- \\
& & \textcolor{blue}{R}+\textcolor{orange!80!black}{C} & 5.31 & {\color{green!60!black}40.8} \\
& & \textcolor{blue}{R}+\textcolor{orange!80!black}{C}+\textcolor{purple}{L}$^{*}$ & 4.82 & {\color{green!60!black}46.3} \\
& & \textcolor{blue}{R}+\textcolor{orange!80!black}{C}$^{*}$ & \textbf{4.68} & {\color{green!60!black}\textbf{47.8}} \\

\cmidrule(lr){2-5}

& \multirow{2}{*}{\textcolor{purple!70!black}{Latent Projection}}
& \emph{w/o} spherify & 89.68 & -- \\
& & \emph{w/} spherify & \textbf{4.68} & {\color{green!60!black}\textbf{94.8}} \\

\midrule

\multirow{6}{*}{\fontsize{9}{10}\selectfont ImageNet-1K}
& \multirow{3}{*}{\textcolor{brown!70!black}{Sampling Steps}}
& 2 & 6.85 & -- \\
& & 4 & 2.25 & {\color{green!60!black}67.2} \\
& & 6 & \textbf{2.11} & {\color{green!60!black}\textbf{69.2}} \\
\cmidrule(lr){2-5}

& \multirow{2}{*}{\textcolor{cyan!70!black}{Pretrained model}}
& Sphere Encoder~\cite{yue2026sphere} & 18.50 & -- \\
& & Ours & 2.60 & {\color{green!60!black} 85.9} \\

\midrule

\multirow{3}{*}{\fontsize{9}{10}\selectfont Animal-Faces}
& \multirow{3}{*}{\textcolor{teal!70!black}{Autoencoder}}
& Flux VAE \cite{flux2024} & 172.25 & -- \\
& & GAE \cite{liu2026geometric} & 23.26 & {\color{green!60!black}86.5} \\
& & RAE \cite{zheng2025ditrae} & \textbf{10.63} & {\color{green!60!black}\textbf{93.8}} \\

\bottomrule
\end{tabular}}

\vspace{0.1cm}
{\footnotesize\raggedright $^{*}$ indicates continued fine-tuning for 300 epochs.\par}

\label{tab:ablation}
\vspace{-0.2cm}
\end{table}

\section{Limitations and Conclusion}

Our method still has some limitations. First, we focus on class-conditional image generation and do not evaluate harder settings such as text-to-image generation. Second, our approach relies on a strong pretrained representation autoencoder to provide semantically meaningful latent representations. This dependency is central to the method and may limit its generality, as performance can vary with the choice and quality of the pretrained encoder, and may introduce additional computational or availability constraints. Third, due to this reliance on a fixed encoder, it remains challenging to achieve high-quality one-step sampling; future work on jointly fine-tuning the encoder with representation alignment from vision foundation models may help address this limitation.

In conclusion, we propose an efficient image generation framework that operates entirely in a spherical latent space. By decoupling reconstruction from generation, our approach avoids repeated pixel–latent transitions, resulting in a simpler pipeline with improved efficiency and better performance. We hope this work encourages further exploration of spherical latent-space modeling, as well as extensions to more general conditioning settings and challenging generation tasks.


\newpage
\bibliographystyle{plain}
\bibliography{references}

\appendix

\begin{center}
{\Large \bf Appendix for}\\[0.4em]
{\Large \bf Efficient Image Synthesis with Sphere Latent Encoder}
\end{center}

\section{Implementation}

\begin{table}[H]
\vspace{-0.5cm}
\centering
\caption{Configurations on different datasets.}
\label{tab:dataset_configs}

\footnotesize
\setlength{\tabcolsep}{4pt} 
\renewcommand{\arraystretch}{1} 

\begin{tabular}{l|ccc}
\toprule
\textbf{dataset}
& \textbf{Animal-Faces}~\cite{choi2020stargan}
& \textbf{Oxford-Flowers}~\cite{nilsback2008automated}
& \textbf{ImageNet-1K}~\cite{imagenet} \\
\midrule
\rowcolor{gray!20}
\multicolumn{4}{l}{\textbf{model configuration}} \\
params (M)     & 130 & 130 & 676 \\
depth          & 12 & 12  & 28 \\
hidden dim     & 768 & 768 & 1152 \\
heads          & 12 & 12  & 16 \\
patch size     & \multicolumn{3}{c}{$1\times1$} \\

\midrule
\rowcolor{gray!20}
\multicolumn{4}{l}{\textbf{training}} \\
epochs         & 500  & 2000 & 600 \\
batch size     & 256  & 256  & 512 \\
dropout        & \multicolumn{3}{c}{0.0} \\
optimizer      & \multicolumn{3}{c}{AdamW} \\
lr schedule    & \multicolumn{3}{c}{constant} \\
lr             & \multicolumn{3}{c}{0.0001} \\
Adam $(\beta_1,\beta_2)$ & \multicolumn{3}{c}{$(0.9, 0.95)$} \\
weight decay   & \multicolumn{3}{c}{0.0} \\
ema decay      & \multicolumn{3}{c}{0.9995} \\

\midrule
\rowcolor{gray!20}
\multicolumn{4}{l}{\textbf{Loss configuration}} \\
$\sigma, \sigma_{sub}$ sampler         & \multicolumn{3}{c}{lognorm$(0.4, 1.0)$} \\
L1 Reconstruction weight  & \multicolumn{3}{c}{50.0} \\
L1 Consistency weight  & \multicolumn{3}{c}{25.0} \\
Cosine Reconstruction weight  & \multicolumn{3}{c}{1.0} \\
Cosine Consistency weight  & \multicolumn{3}{c}{1.0} \\

$\sigma$ range & \multicolumn{3}{c}{$[0, 85]$} \\
$\sigma$ mix range & \multicolumn{3}{c}{$[85, 89]$} \\
Mix probability  & \multicolumn{3}{c}{0.2} \\
Cls-cond drop  & \multicolumn{3}{c}{0.1} \\

\bottomrule
\end{tabular}
\end{table}

\myheading{Denoising model architecture.} \label{appendix:implementation}
We adopt a standard transformer-based architecture as the denoising model. The input latent has spatial size $16 \times 16$ with 768 channels. The backbone follows the SiT architecture~\cite{ma2024sit}. Detailed configuration for each dataset is provided in \Tref{tab:dataset_configs}. For experiments on Animal-Faces \cite{choi2020stargan} and Oxford-Flowers \cite{nilsback2008automated}, we adopt the SiT-B/1 architecture~\cite{ma2024sit} as our denoising model, corresponding to the \textit{Base} configuration with a patch size of 1. For the large-scale ImageNet-1K dataset~\cite{imagenet}, we adopt the SiT-XL/1 architecture~\cite{ma2024sit} as our denoising model with a patch size of 1, resulting in a comparable number of processing tokens to prior works~\cite{geng2025meanflow, yu2025repa, geng2025improved}.

\myheading{Representation autoencoder.}
We use a pretrained representation autoencoder from~\cite{zheng2025ditrae}, where the encoder is DINOv2-B~\cite{oquab2023dinov2} and the decoder is a ViT-XL. The model contains 502M parameters and requires 128 GFLOPs for a forward pass (encoder + decoder), which is substantially lower than SD-VAE (445 GFLOPs). The autoencoder maps input images of size $256 \times 256 \times 3$ to latents of size $16 \times 16 \times 768$. The reconstruction FID (rFID) reported in~\cite{zheng2025ditrae} is $0.49$.

\myheading{Datasets and training details.}
We conduct experiments on Animal-Faces~\cite{choi2020stargan}, Oxford Flowers~\cite{nilsback2008automated}, and ImageNet-1K~\cite{imagenet}. All experiments are performed at a resolution of $256 \times 256$, with horizontal flipping applied as the only data augmentation with a probability of $0.5$. For ImageNet-1K, non-square images are first center-cropped to a square before resizing.  To improve training efficiency, all images are pre-encoded into latent representations using DINOv2-B~\cite{oquab2023dinov2}, significantly reducing computational overhead. We train our models on 8 AMD Instinct MI210 GPUs (64 GB memory each) for Animal-Faces and Oxford-Flowers, and scale up to 64 such GPUs for ImageNet-1K.

\myheading{Sampling Configuration.}
We conduct experiments on Animal-Faces \cite{choi2020stargan}, Oxford Flowers~\cite{nilsback2008automated}, and ImageNet-1K~\cite{imagenet}. All experiments use a resolution of $256 \times 256$ with horizontal flipping applied with probability $0.5$. For ImageNet-1K, non-square images are center-cropped before resizing. All images are pre-encoded into latent representations using DINOv2-B~\cite{oquab2023dinov2}, which significantly reduces training overhead.

\section{Additional Technical Details}
\label{appendix:additional}
\myheading{Latent Consistency Loss.} We follow the latent consistency loss in Sphere Encoder \cite{yue2026sphere} and adapt it to latent space. Starting from a noisy latent, we obtain a refined prediction $\mathbf{z}_{\text{refined}} = \mathcal{G}(\mathbf{v}_{\text{NOISY}})$ and project it onto the hypersphere to obtain $\mathbf{v}_{\text{refined}} = \mathcal{F}(\mathbf{z}_{\text{refined}})$. We then apply the same noise level $\sigma$ to produce $\mathbf{v}_{\text{re-noisy}} = \mathcal{F}(\mathbf{v}_{\text{refined}} + \sigma \epsilon)$, which is passed through the model again to obtain $\mathbf{z}_{\text{re-refined}} = \mathcal{G}(\mathbf{v}_{\text{re-noisy}})$. Finally, we enforce consistency between the two predictions using a cosine similarity loss:

\begin{empheq}[box=\fbox]{equation}
\label{eq:latent_consistency}
\mathcal{L}_{\text{lat\_cons}} =
\mathcal{L}_{\text{cosine}}\big(
\mathbf{z}_{\text{re-refined}},
\mathbf{z}_{\text{refined}}
\big)
\end{empheq}

\myheading{Experiment with latent consistency loss.}
In early experiments, we observe that including this loss from the beginning of training substantially slows convergence and increases iteration time (from 201\,ms to 336\,ms per iteration). We hypothesize that, during early training, the denoising model is not yet capable of producing high-quality cleaned latents; consequently, the content of $\mathbf{z}_{\text{refined}}$ is unreliable, making the latent consistency objective $\mathcal{L}_{\text{lat\_cons}}$ difficult to optimize. To mitigate this issue, we perform an ablation study where we first train the model for 500 epochs using both the reconstruction loss and the latent consistency loss, and then fine-tune for an additional 300 epochs under two settings: with and without the latent consistency loss. Results are reported in \Tref{tab:ablation}, showing that removing the latent consistency loss yields better performance.

\myheading{Metric Calculation.} \label{appendix:metric_calc} We follow RAE~\cite{zheng2025ditrae} and use class-balanced sampling to generate 50,000 images. For ImageNet-1K \cite{imagenet}, we adopt the reference statistics from Guided Diffusion~\cite{dhariwal2021diffusion}. For CMMD~\cite{jayasumana2024rethinking}, we evaluate using the same generated and reference sets. For Animal-Faces~\cite{choi2020stargan} and Oxford-Flowers~\cite{nilsback2008automated} datasets, we use the training set as the reference distribution. Following standard practice, we sample 50,000 images for evaluation, even when the training sets contain fewer samples.

\myheading{FLOP Calculation.} \label{appendix:flop_calc}
We compute the computational cost of all models using the THOP library\footnote{\url{https://github.com/ultralytics/thop}}, which estimates the number of floating-point operations (FLOPs). All measurements are reported per image with a batch size of 1. For our method, the total FLOPs are determined by the number of sampling steps. Each step consists of one forward pass of the denoising model, followed by a final decoder pass to reconstruct the image. When classifier-free guidance (CFG) is applied, the FLOPs of the denoising model are doubled at each step. For Sphere Encoder, the generation process alternates between decoder and encoder passes. Consequently, the total FLOPs for $T$ sampling steps are computed as $T$ decoder passes and $(T-1)$ encoder passes. When CFG is enabled, both encoder and decoder computations are doubled accordingly. All reported FLOPs in Table~\ref{tab:animal_faces_flowers} follow these conventions. We further provide a detailed breakdown of FLOPs and parameter counts for each component of our method and Sphere Encoder~\cite{yue2026sphere} in \Tref{tab:imagenet_component_flops}.

\myheading{Noise perturbation during sampling.} \label{appendix:noise_perturbation} Before the beginning of each new sampling cycle, we deliberately perturb the predicted latent following the Sphere Encoder approach \cite{yue2026sphere}. Specifically, we adopt their simple strategy of progressively decaying the injected noise. Across most experiments, we observe that applying a decreasing noise schedule at each timestep improves model performance. For sampling on ImageNet-1K \cite{imagenet}, we use a decay factor of $\gamma = 0.5$ to gradually reduce the noise as sampling progresses. For other datasets, including Animal-Faces \cite{choi2020stargan} and Oxford-Flowers \cite{nilsback2008automated}, we set $\gamma = 0.75$. All experiments use the same maximum noise level $\sigma_{\max} = 24$. We use $\omega = 1.0$ (i.e., no classifier-free guidance) for Animal-Faces, $\omega = 4.8$ for Oxford-Flowers, and $\omega = 3.2$ for ImageNet-1K.
 
\section{Additional Quantitative Results}

\paragraph{Comparison under Matched Inference Budgets.}
Comparisons against 250-step multi-step baselines do not fully characterize the
practical efficiency gap, since such baselines can also be run with fewer steps.
We therefore evaluate RAE~\cite{zheng2025ditrae} at matched low NFE
using both the standard Euler sampler and a fast high-order solver, Flow-DPM-Solver,
adapted from DPM-Solver~\cite{lu2022dpm} for flow matching\footnote{Following
\url{https://github.com/NVlabs/Sana/issues/110}.}.
The results in Table~\ref{tab:matched_nfe} show that our method retains a clear quality–compute advantage at matched 4–8 NFE, even against a strong fast solver.

\begin{table}[h]
\centering
\caption{Quality--compute comparison on ImageNet-1K $256\times256$ under matched
low inference budgets. GFLOPs are measured over the full sampling process,
including decoding.}
\label{tab:matched_nfe}
\small
\begin{tabular}{llccc}
\toprule
Method & Sampler & NFE & GFLOPs & FID$\downarrow$ \\
\midrule
  & Euler           & $4\times2$ & 2717 & 15.64 \\
  & Euler           & $6\times2$ & 3969 & 5.84 \\
  & Euler           & $8\times2$ & 5221 & 3.30 \\
RAE & Flow-DPM-Solver & $4\times2$ & 2717 & 15.09 \\
  & Flow-DPM-Solver & $6\times2$ & 3969 & 5.33 \\
  & Flow-DPM-Solver & $8\times2$ & 5221 & 3.11 \\
\midrule
Ours-XL/1 & Ours & $4\times2$ & \textbf{2051} & 2.25 \\
Ours-XL/1 & Ours & $6\times2$ & 2969 & \textbf{2.11} \\
\bottomrule
\end{tabular}
\end{table}

\section{Broader Societal Impacts}
Our method improves the efficiency of image generation, enabling applications in content creation and machine learning while reducing computational cost. However, easier generation of realistic images may increase risks of misuse, such as deceptive or misleading content. We encourage responsible use and further research on safeguards like detection and usage guidelines.

\section{Additional Qualitative Results}
We present $256 \times 256$ samples generated by our best-performing model, SiT-XL/1 with DINOv2-B, using classifier-free guidance of 3.2, $\sigma_{\max}=24$, and $\gamma=0.5$.

\begin{figure*}[t]
\centering

\begin{subfigure}[t]{0.47\linewidth}
    \centering
    \includegraphics[width=\linewidth]{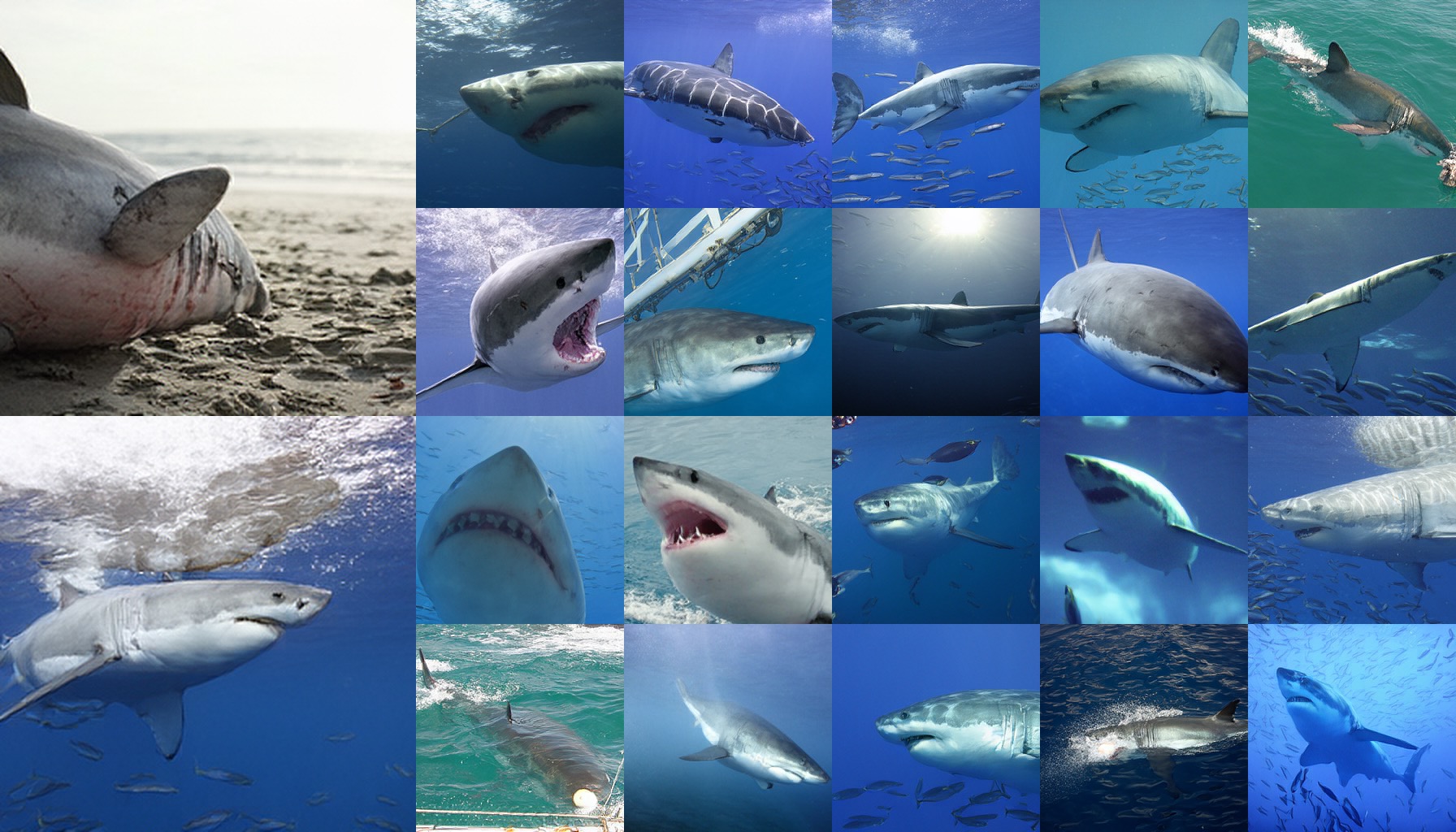}
    \caption{class 0002: great white shark, white shark}
\end{subfigure}
\hspace{0.1cm}
\begin{subfigure}[t]{0.47\linewidth}
    \centering
    \includegraphics[width=\linewidth]{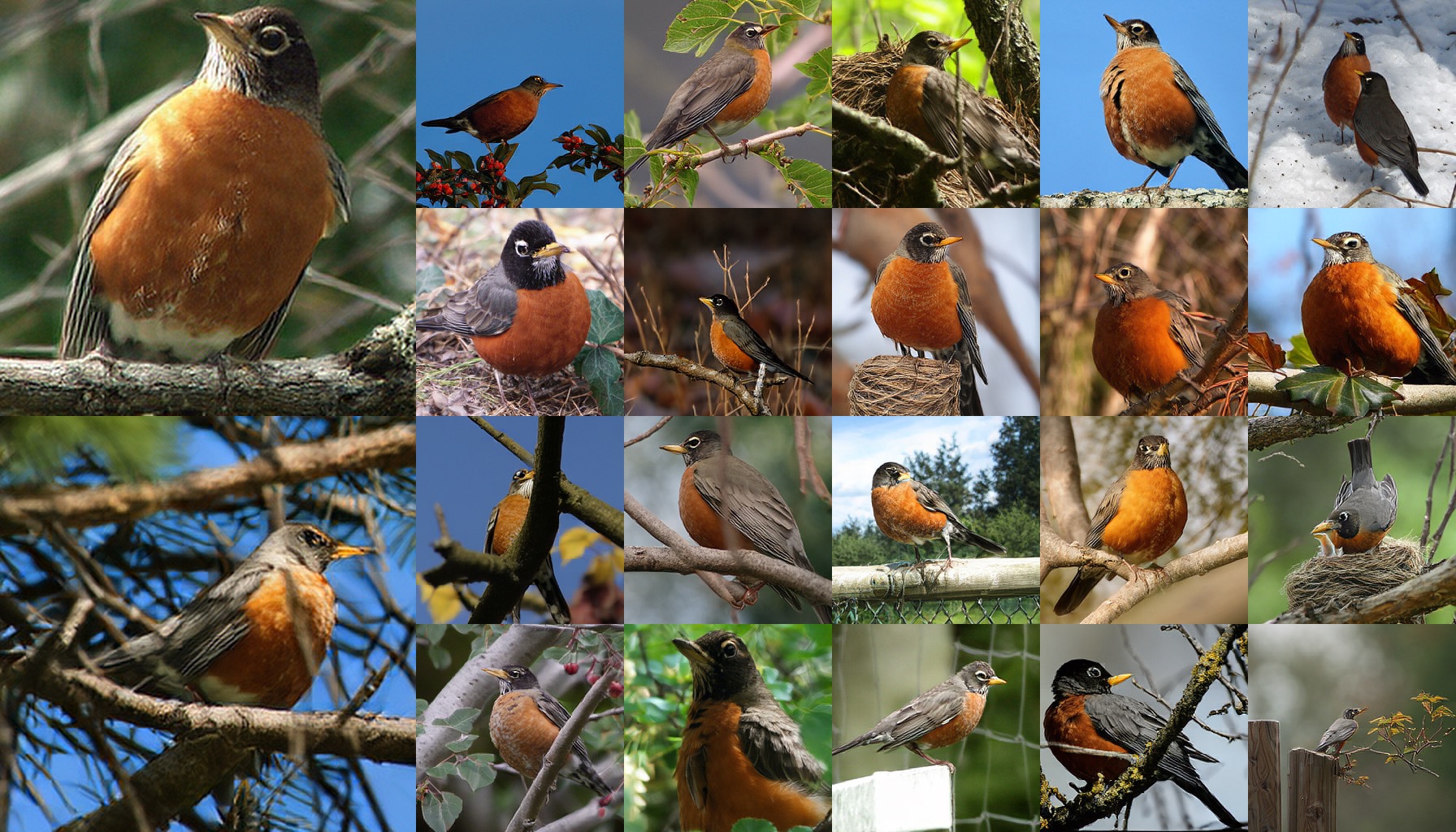}
    \caption{class 0015: robin, American robin}
\end{subfigure}

\begin{subfigure}[t]{0.47\linewidth}
    \centering
    \includegraphics[width=\linewidth]{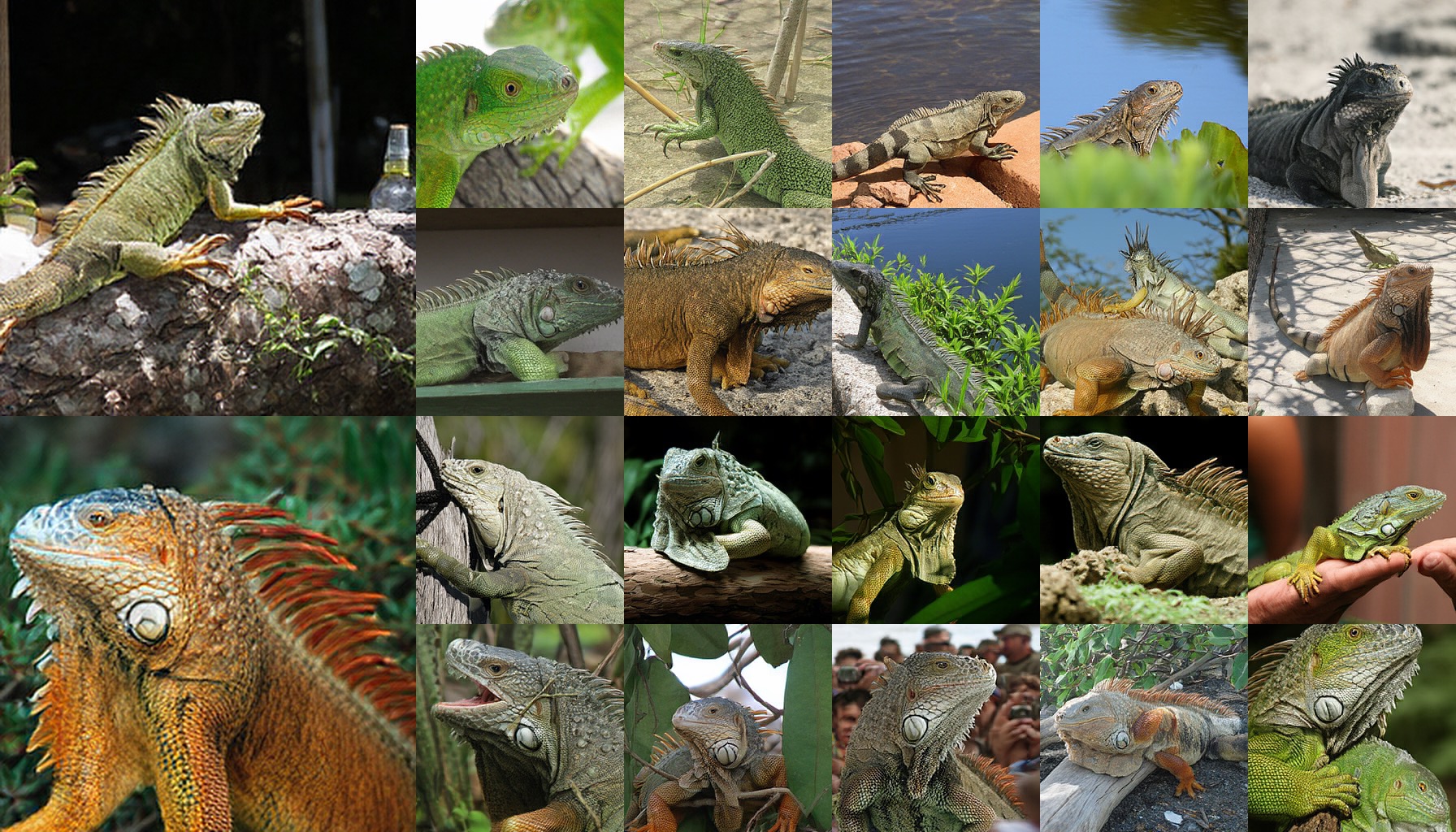}
    \caption{class 0039: common iguana, iguana}
\end{subfigure}
\hspace{0.1cm}
\begin{subfigure}[t]{0.47\linewidth}
    \centering
    \includegraphics[width=\linewidth]{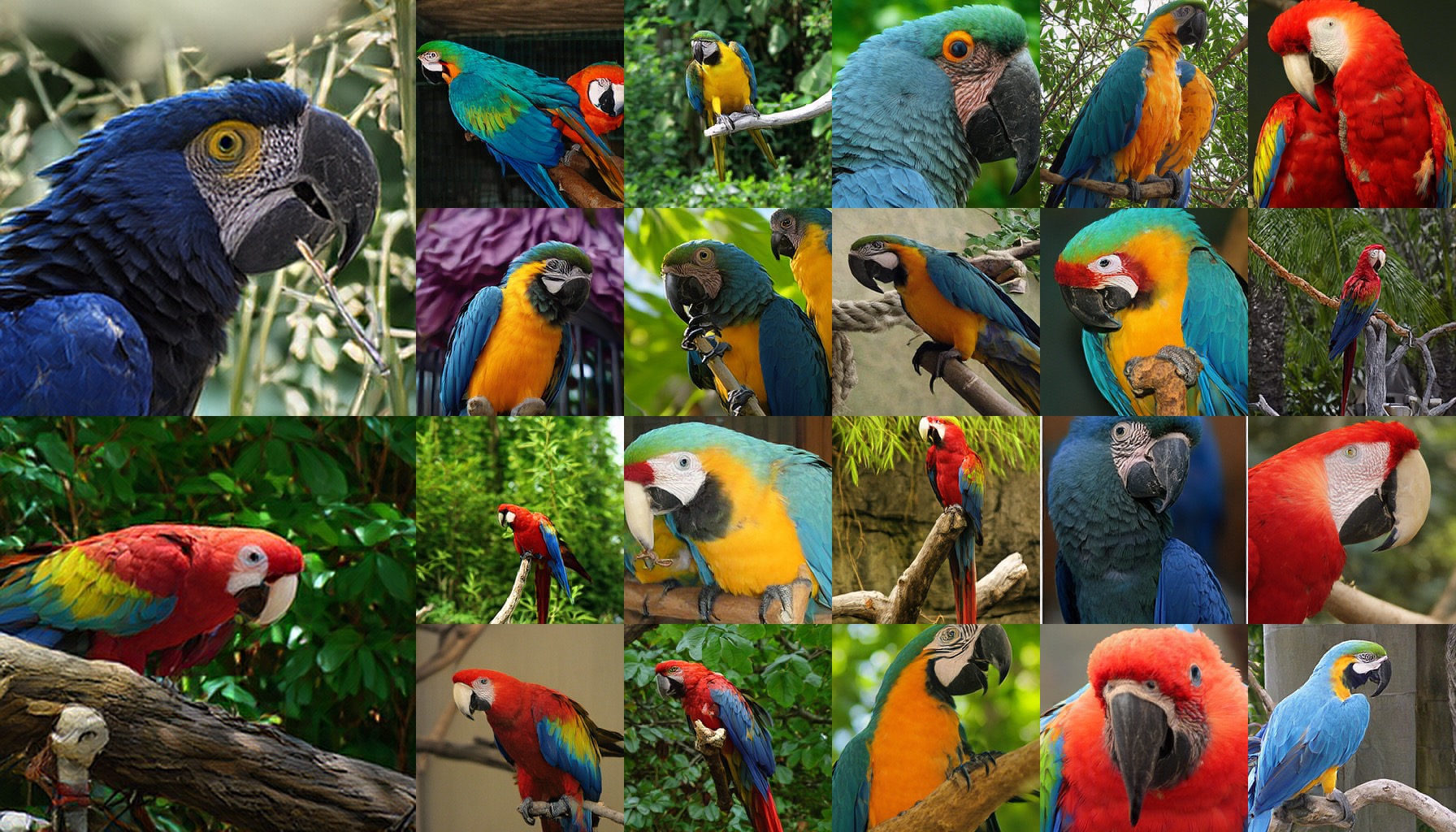}
    \caption{class 0088: macaw}
\end{subfigure}

\begin{subfigure}[t]{0.47\linewidth}
    \centering
    \includegraphics[width=\linewidth]{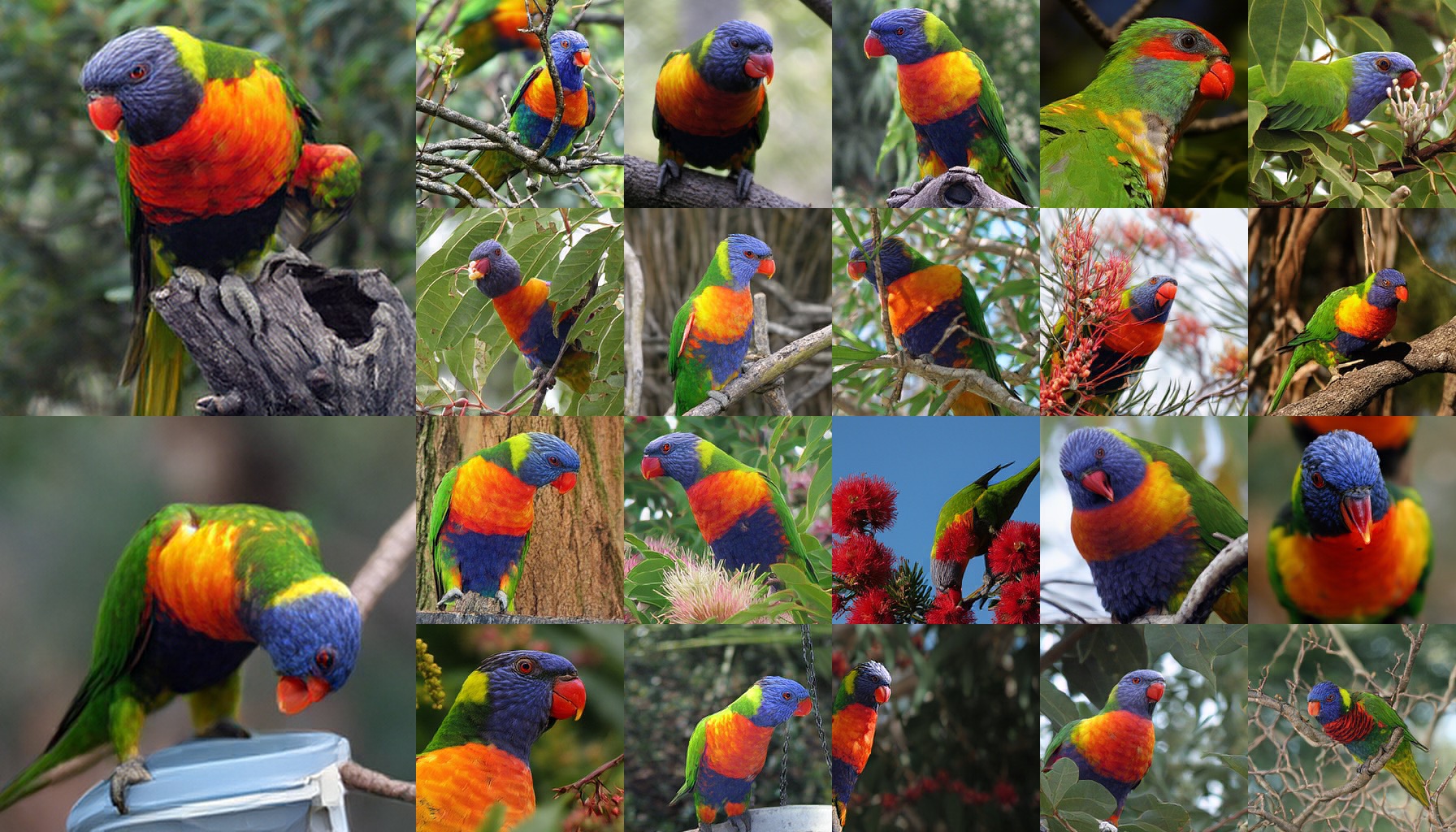}
    \caption{class 0090: lorikeet}
\end{subfigure}
\hspace{0.1cm}
\begin{subfigure}[t]{0.47\linewidth}
    \centering
    \includegraphics[width=\linewidth]{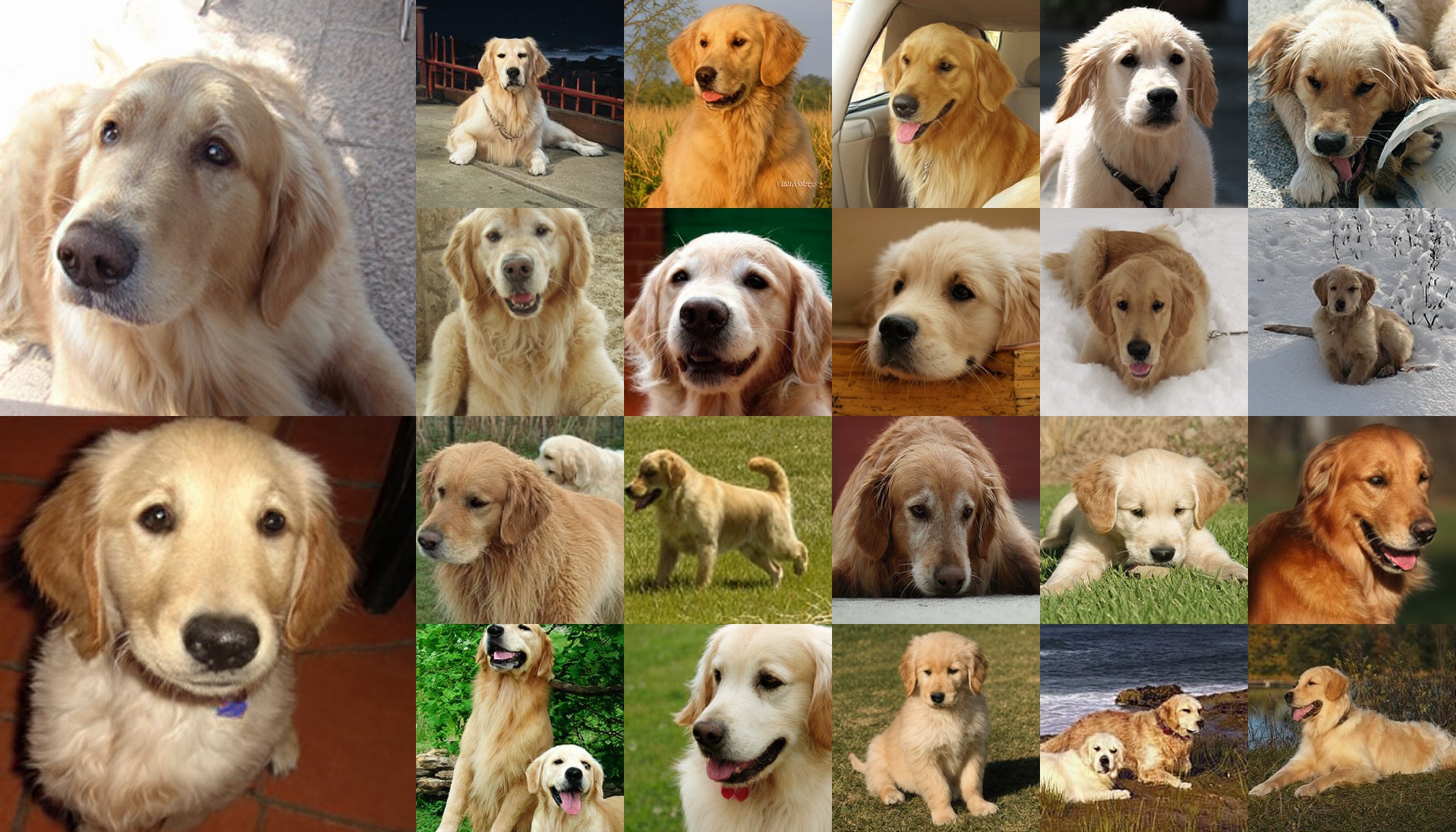}
    \caption{class 0207: golden retriever}
\end{subfigure}

\begin{subfigure}[t]{0.47\linewidth}
    \centering
    \includegraphics[width=\linewidth]{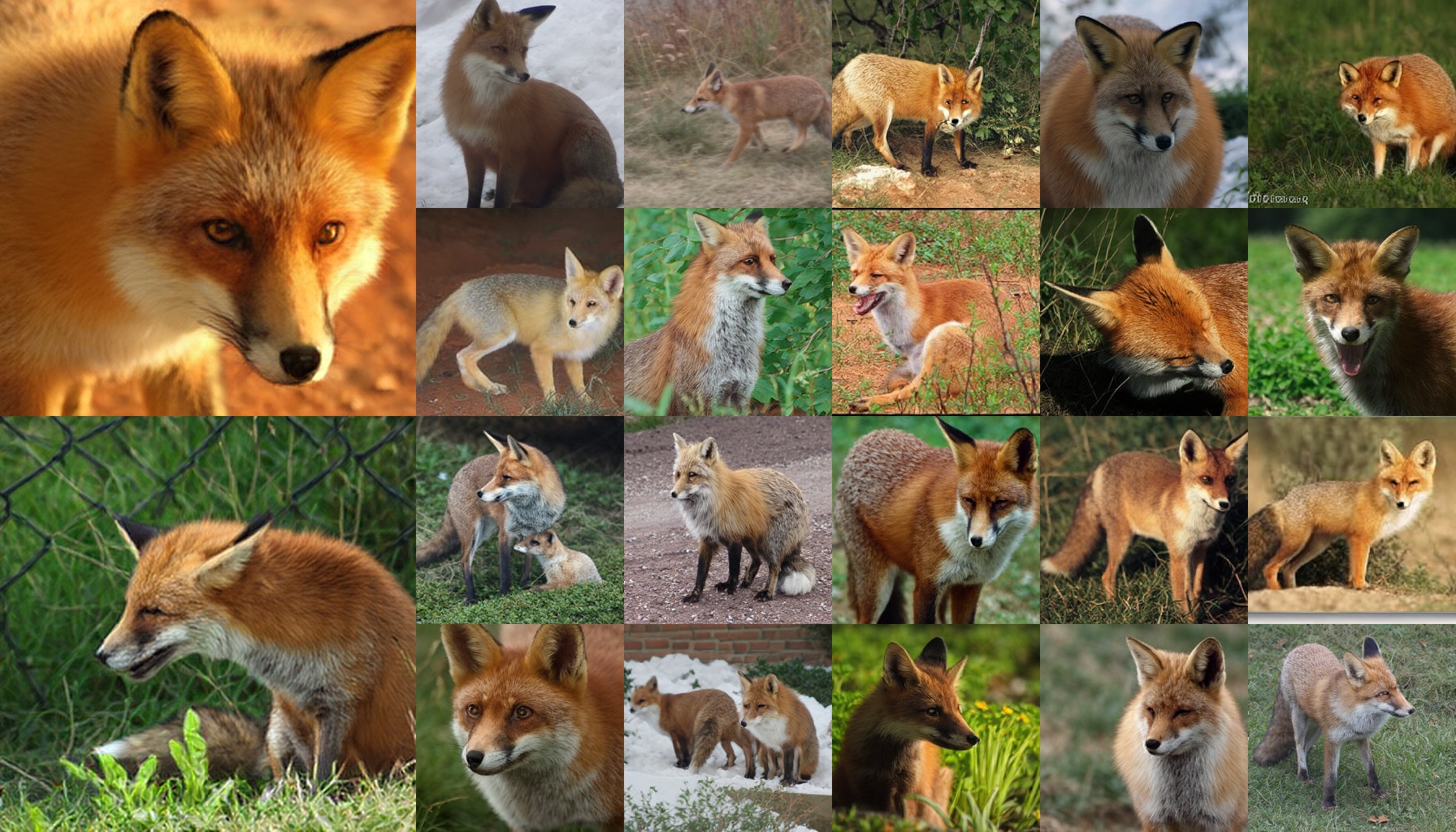}
    \caption{class 0277: red fox, Vulpes vulpes}
\end{subfigure}
\hspace{0.1cm}
\begin{subfigure}[t]{0.47\linewidth}
    \centering
    \includegraphics[width=\linewidth]{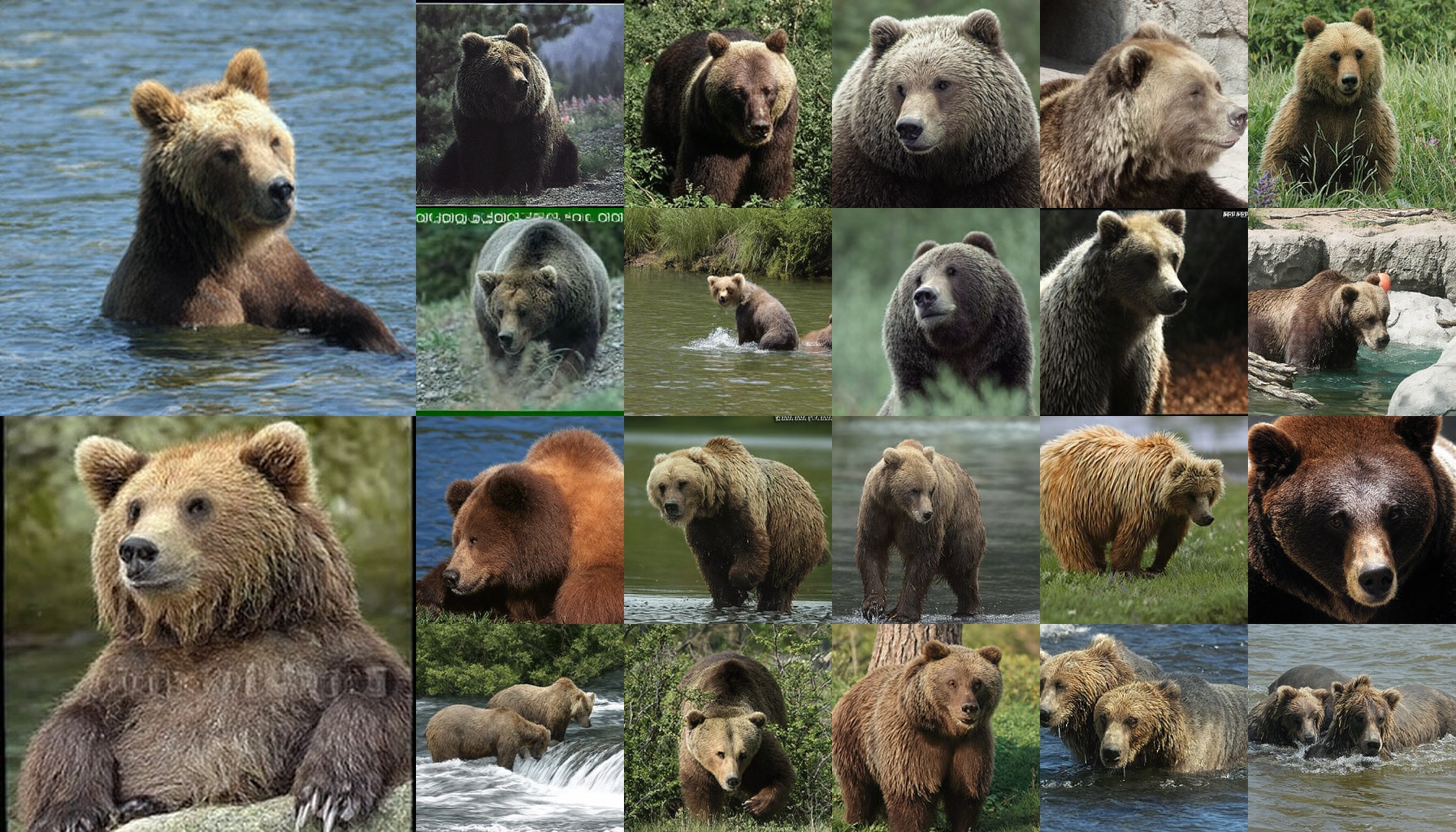}
    \caption{class 0294: brown bear, bruin, Ursus arctos}
\end{subfigure}

\begin{subfigure}[t]{0.47\linewidth}
    \centering
    \includegraphics[width=\linewidth]{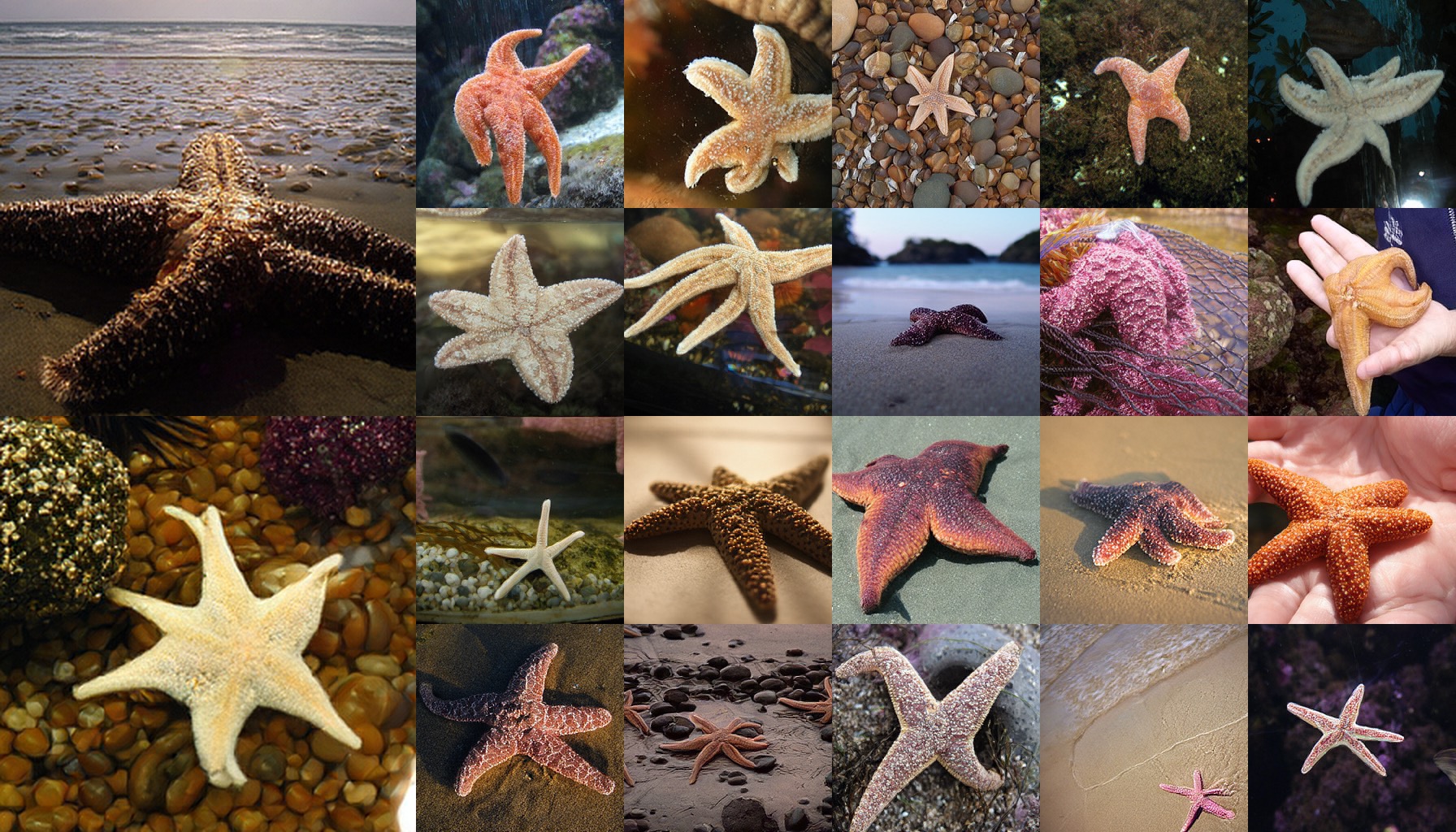}
    \caption{class 0327: starfish, sea star}
\end{subfigure}
\hspace{0.1cm}
\begin{subfigure}[t]{0.47\linewidth}
    \centering
    \includegraphics[width=\linewidth]{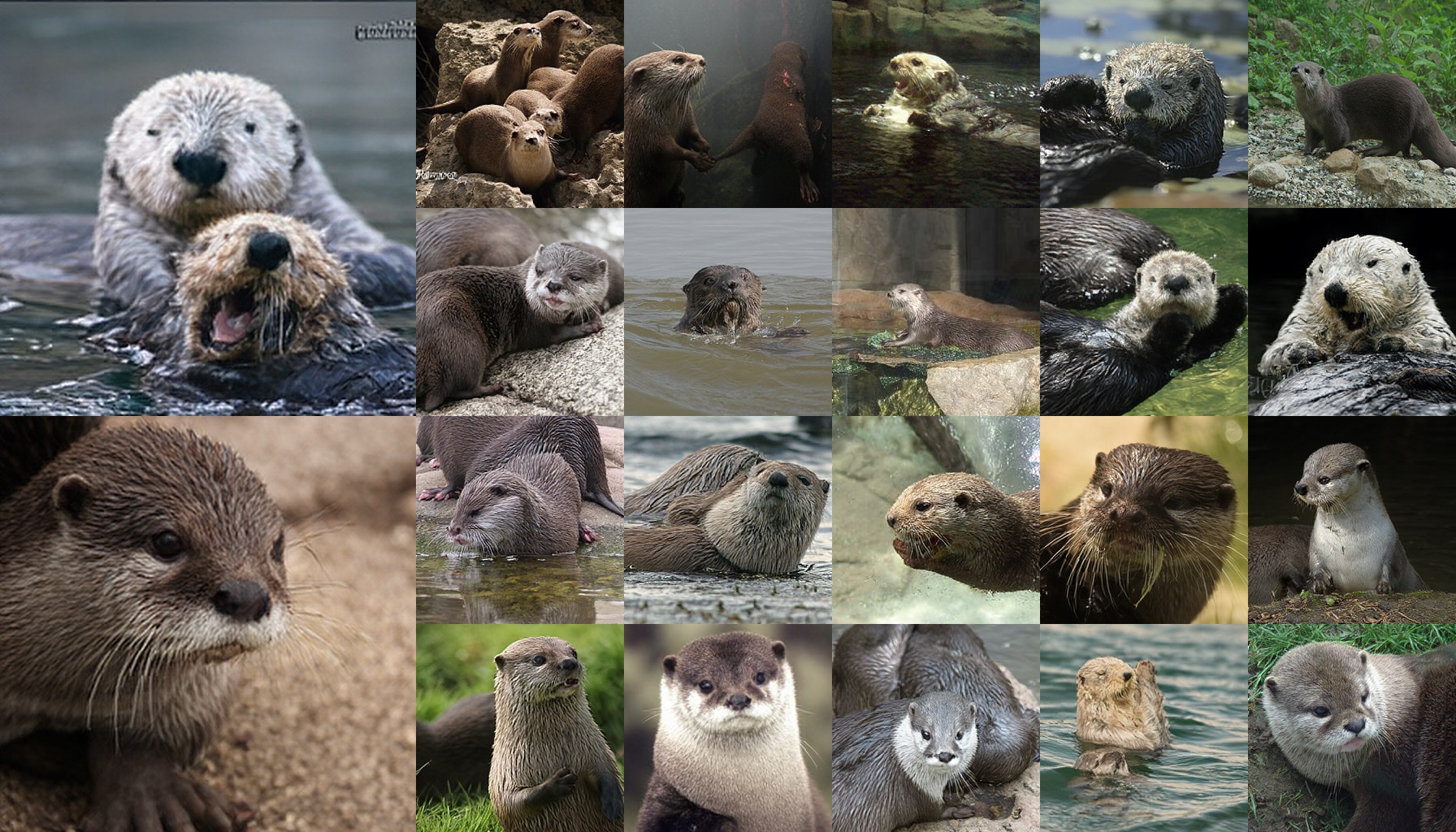}
    \caption{class 0360: otter}
\end{subfigure}

\caption{\textbf{4-NFE Generation Results.} Examples of class-conditional generation on ImageNet $256\times256$ using our 4-NFE model.}
\end{figure*}

\begin{figure*}[t]
\centering

\begin{subfigure}[t]{0.47\linewidth}
    \centering
    \includegraphics[width=\linewidth]{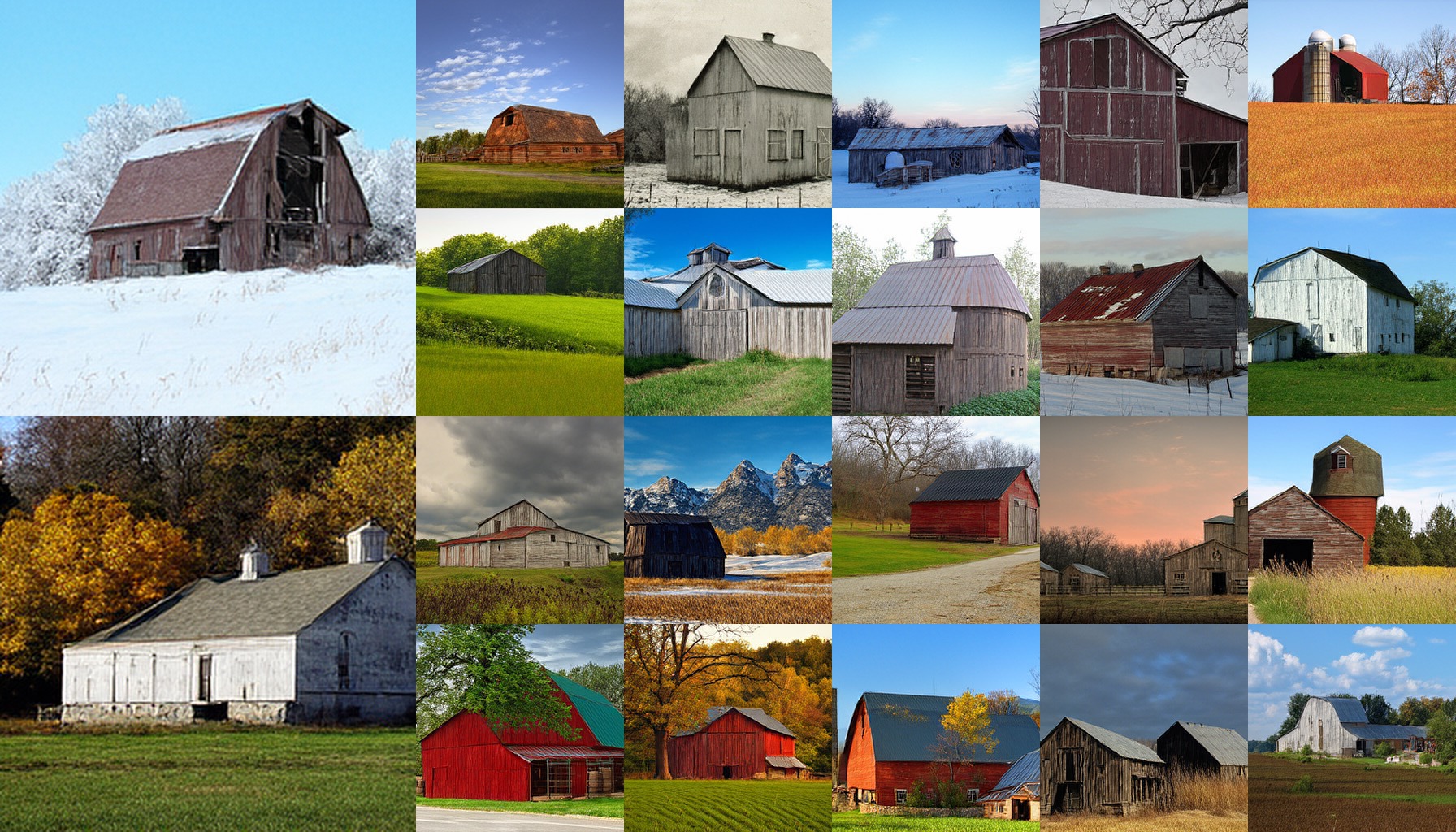}
    \caption{class 0425: barn}
\end{subfigure}
\hspace{0.1cm}
\begin{subfigure}[t]{0.47\linewidth}
    \centering
    \includegraphics[width=\linewidth]{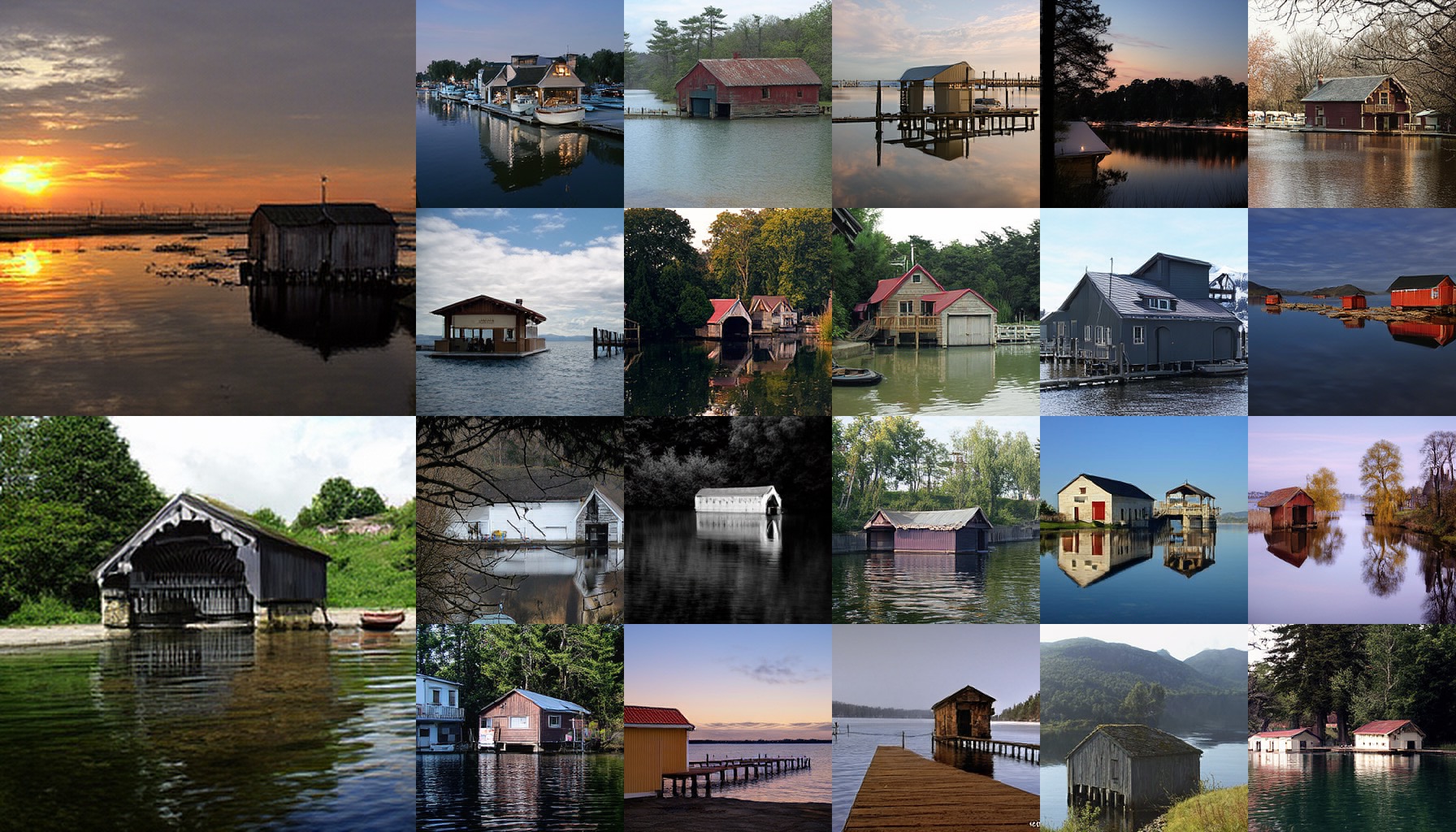}
    \caption{class 0449: boathouse}
\end{subfigure}

\begin{subfigure}[t]{0.47\linewidth}
    \centering
    \includegraphics[width=\linewidth]{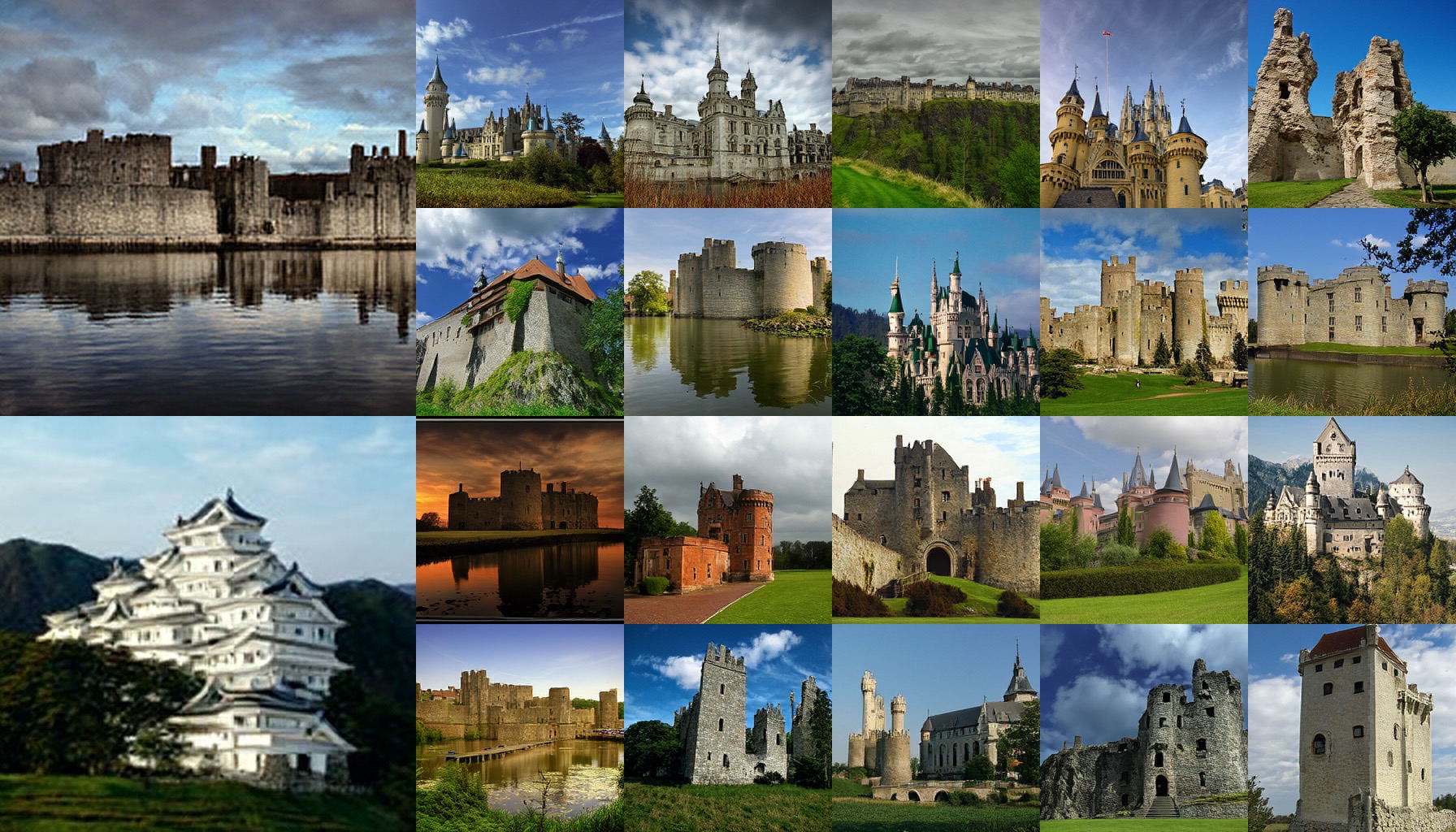}
    \caption{class 0483: castle}
\end{subfigure}
\hspace{0.1cm}
\begin{subfigure}[t]{0.47\linewidth}
    \centering
    \includegraphics[width=\linewidth]{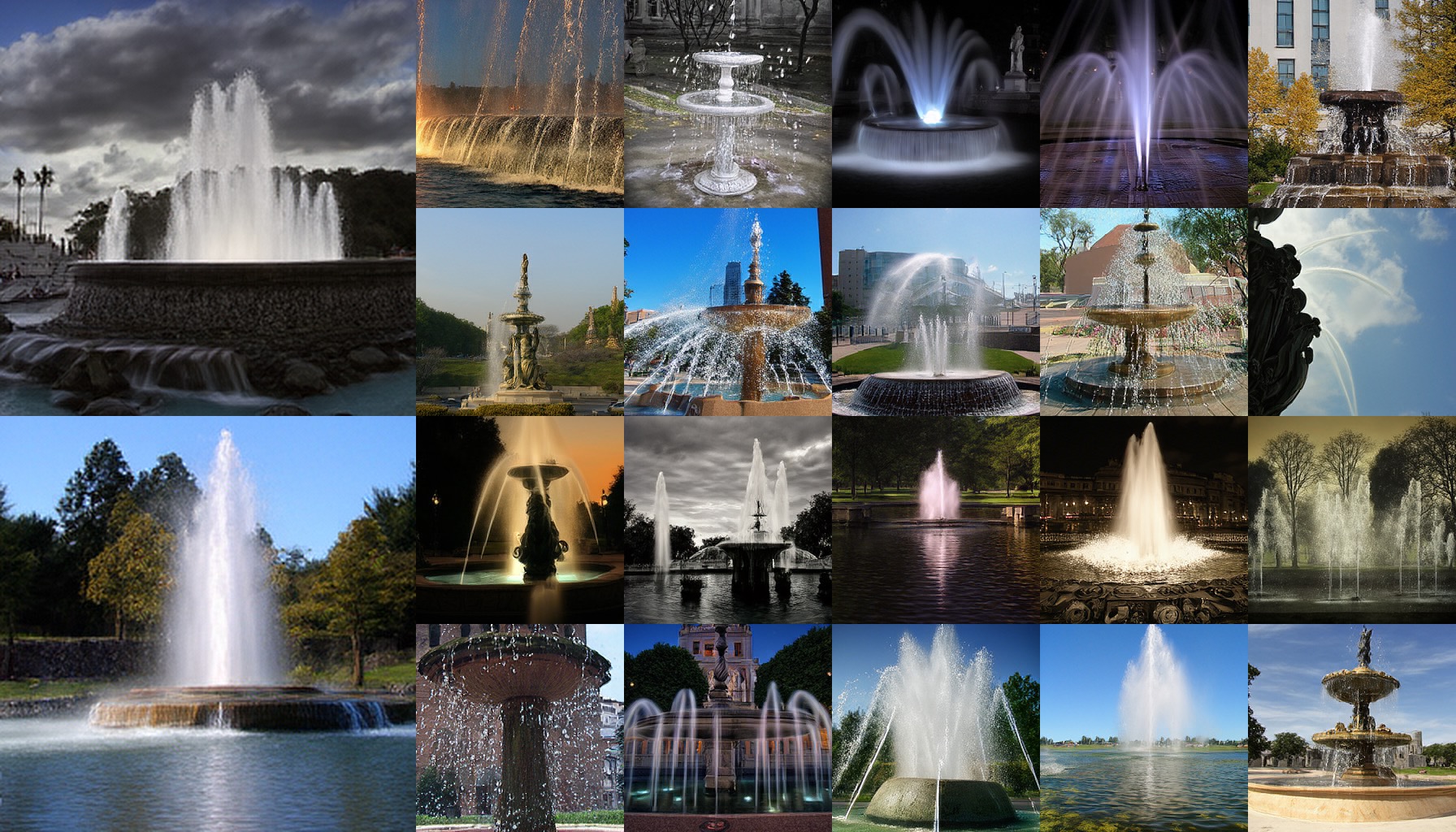}
    \caption{class 0562: fountain}
\end{subfigure}

\begin{subfigure}[t]{0.47\linewidth}
    \centering
    \includegraphics[width=\linewidth]{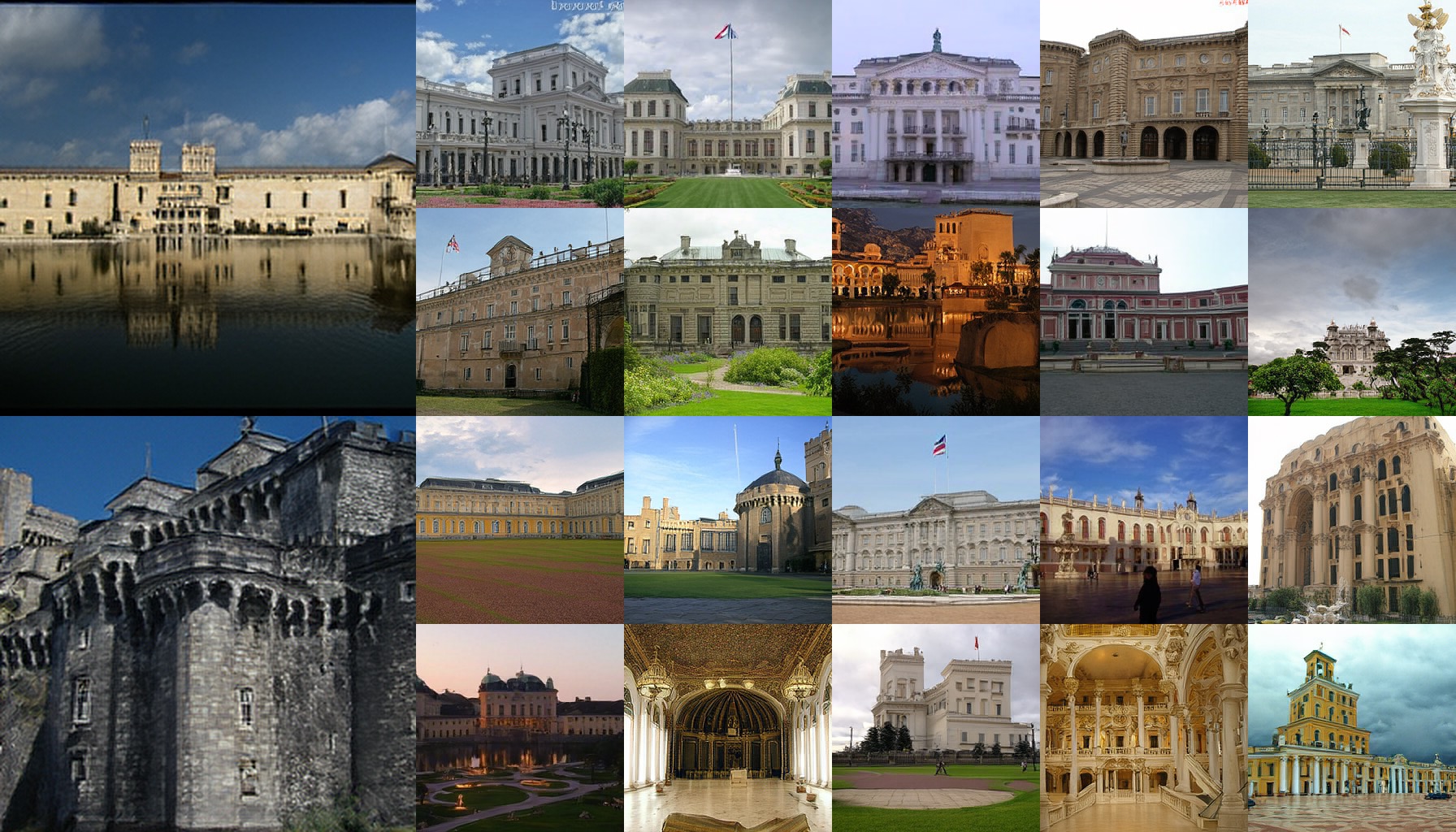}
    \caption{class 0698: palace}
\end{subfigure}
\hspace{0.1cm}
\begin{subfigure}[t]{0.47\linewidth}
    \centering
    \includegraphics[width=\linewidth]{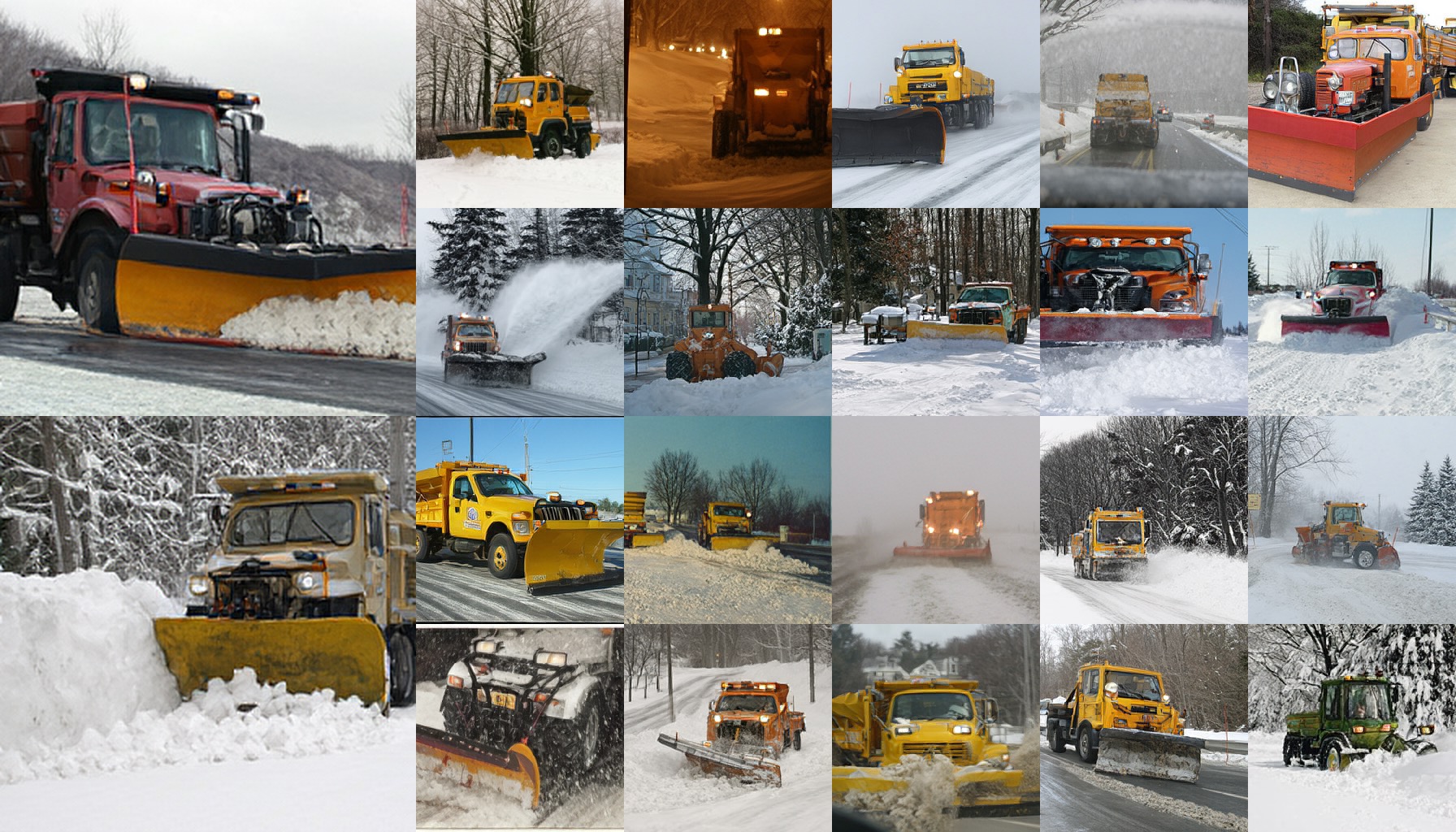}
    \caption{class 0803: snowplow, snowplough}
\end{subfigure}

\begin{subfigure}[t]{0.47\linewidth}
    \centering
    \includegraphics[width=\linewidth]{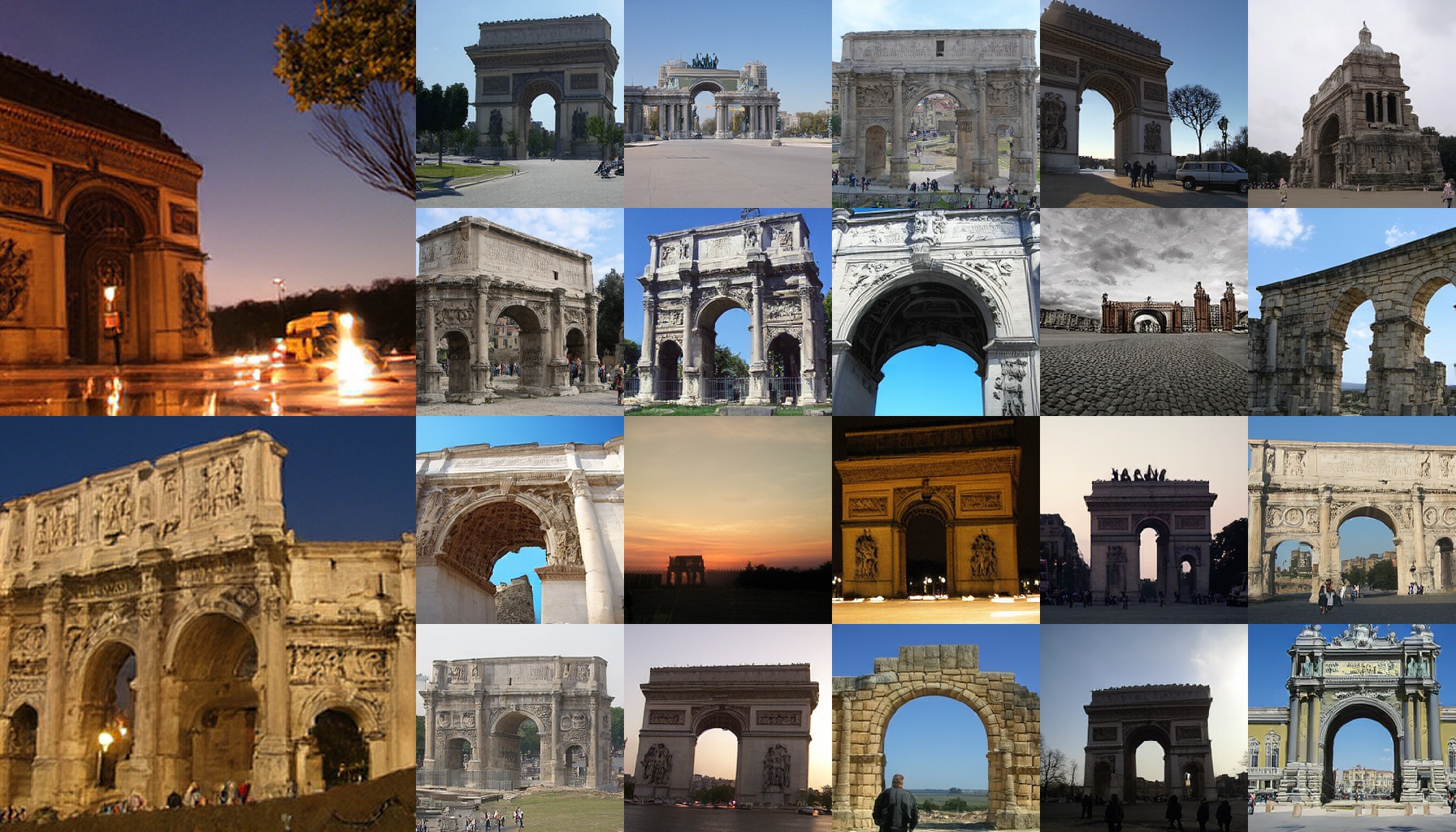}
    \caption{class 0873: triumphal arch}
\end{subfigure}
\hspace{0.1cm}
\begin{subfigure}[t]{0.47\linewidth}
    \centering
    \includegraphics[width=\linewidth]{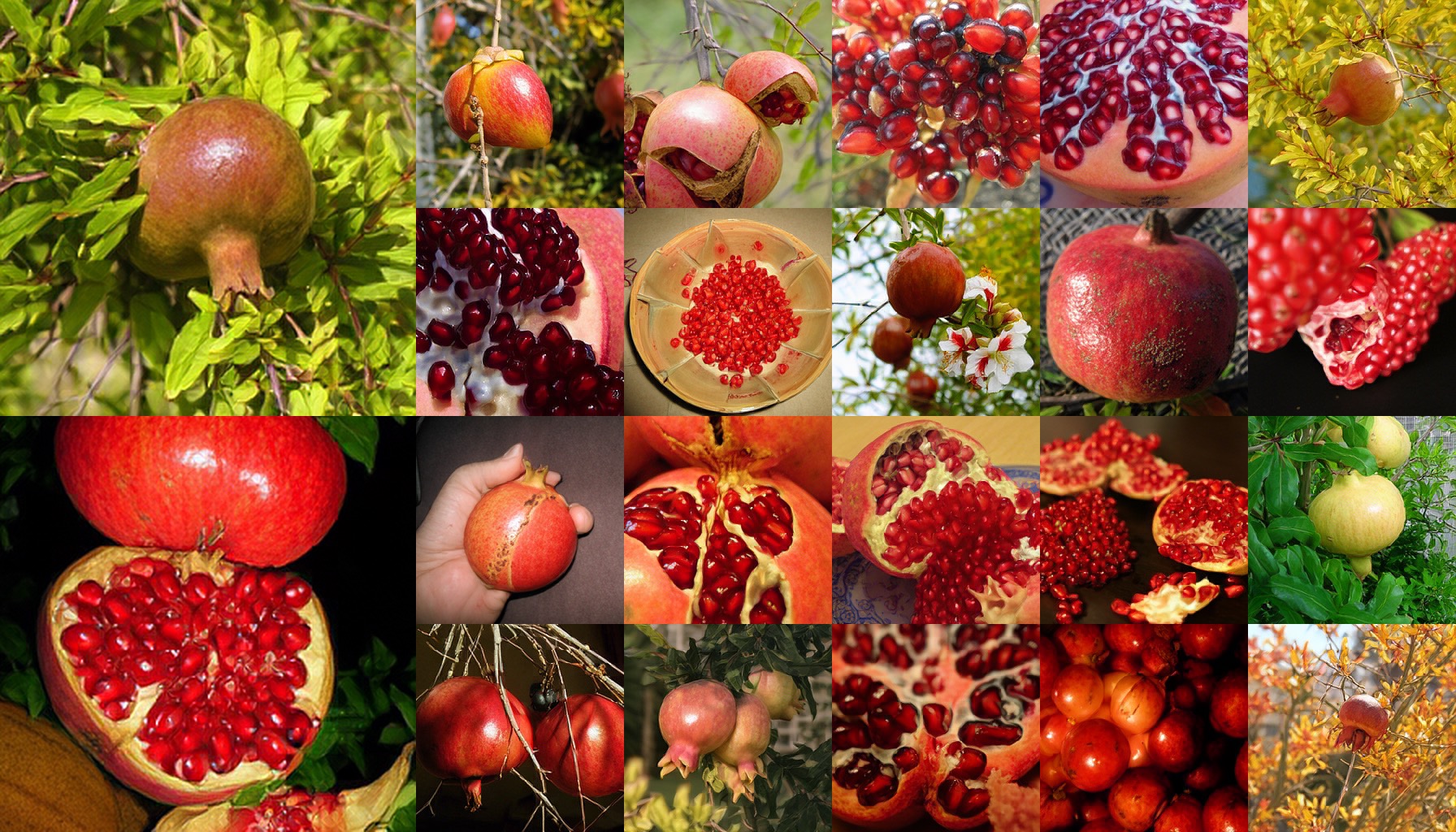}
    \caption{class 0957: pomegranate}
\end{subfigure}

\begin{subfigure}[t]{0.47\linewidth}
    \centering
    \includegraphics[width=\linewidth]{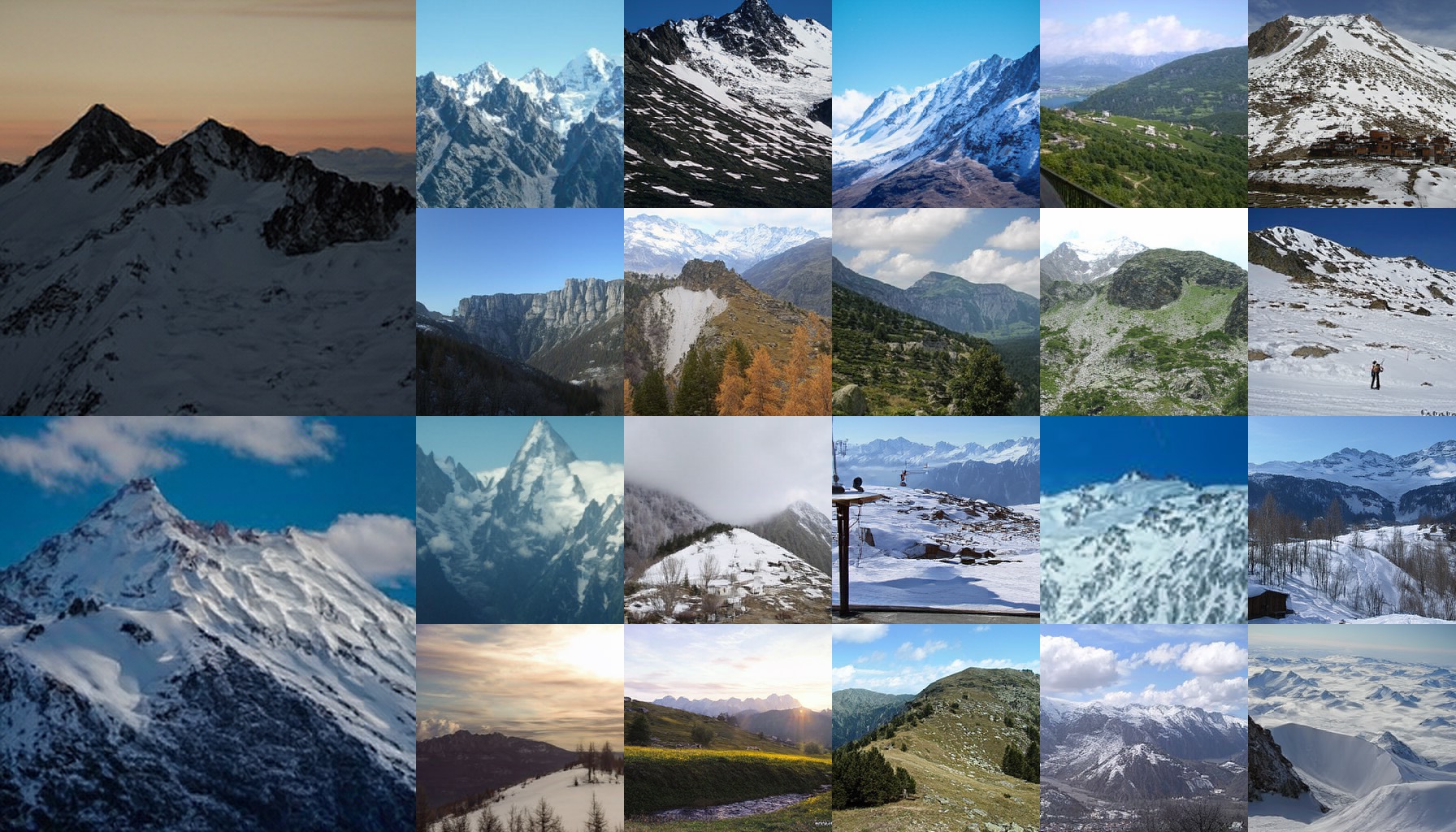}
    \caption{class 0970: alp}
\end{subfigure}
\hspace{0.1cm}
\begin{subfigure}[t]{0.47\linewidth}
    \centering
    \includegraphics[width=\linewidth]{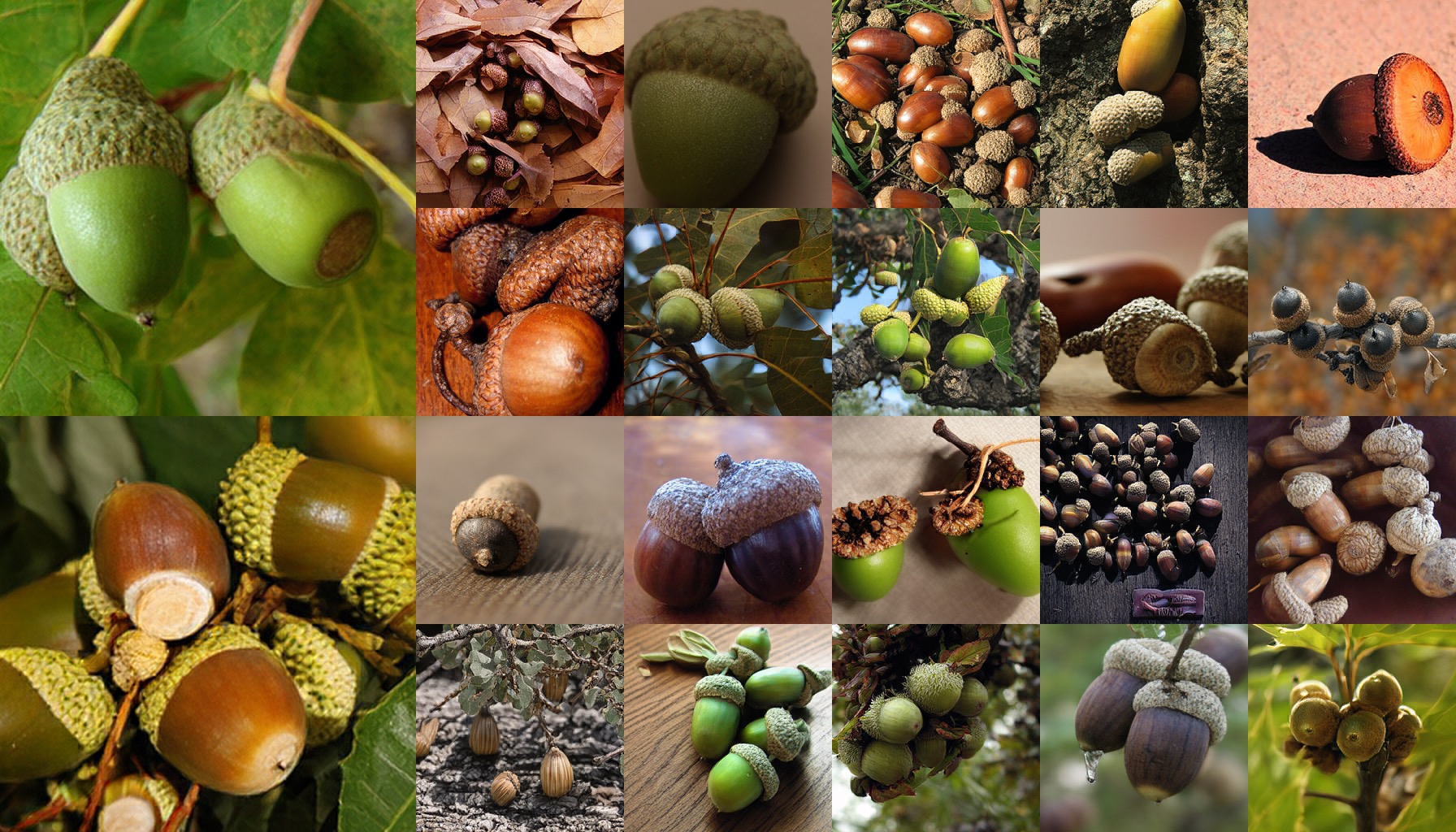}
    \caption{class 0988: acorn}
\end{subfigure}

\caption{\textbf{4-NFE Generation Results.} Examples of class-conditional generation on ImageNet $256\times256$ using our 4-NFE model.}

\end{figure*}

\begin{figure*}[t]
\centering

\begin{subfigure}[t]{0.47\linewidth}
    \includegraphics[width=\linewidth]{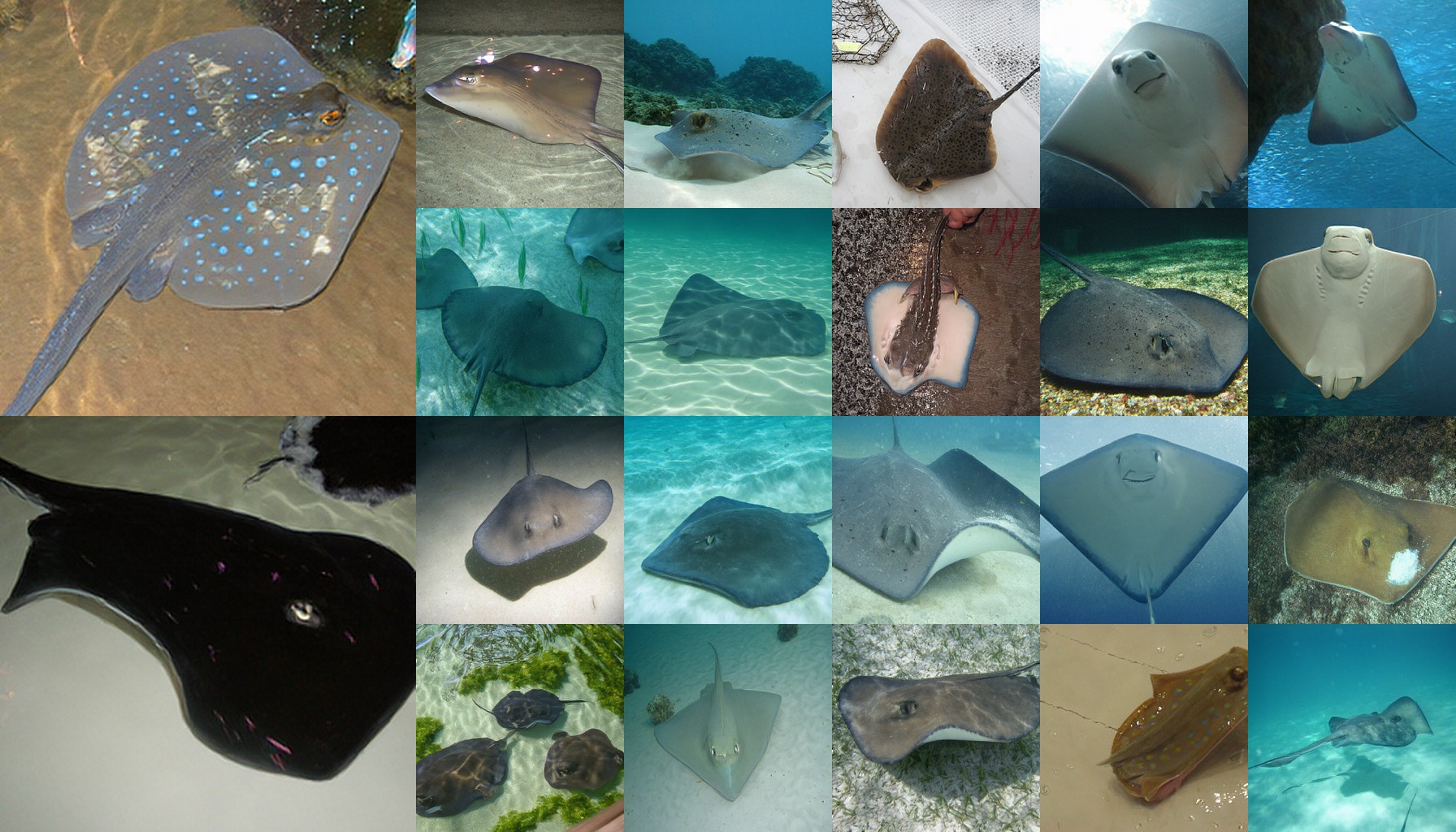}
    \caption{class 0006: stingray}
\end{subfigure}
\hspace{0.1cm}
\begin{subfigure}[t]{0.47\linewidth}
    \includegraphics[width=\linewidth]{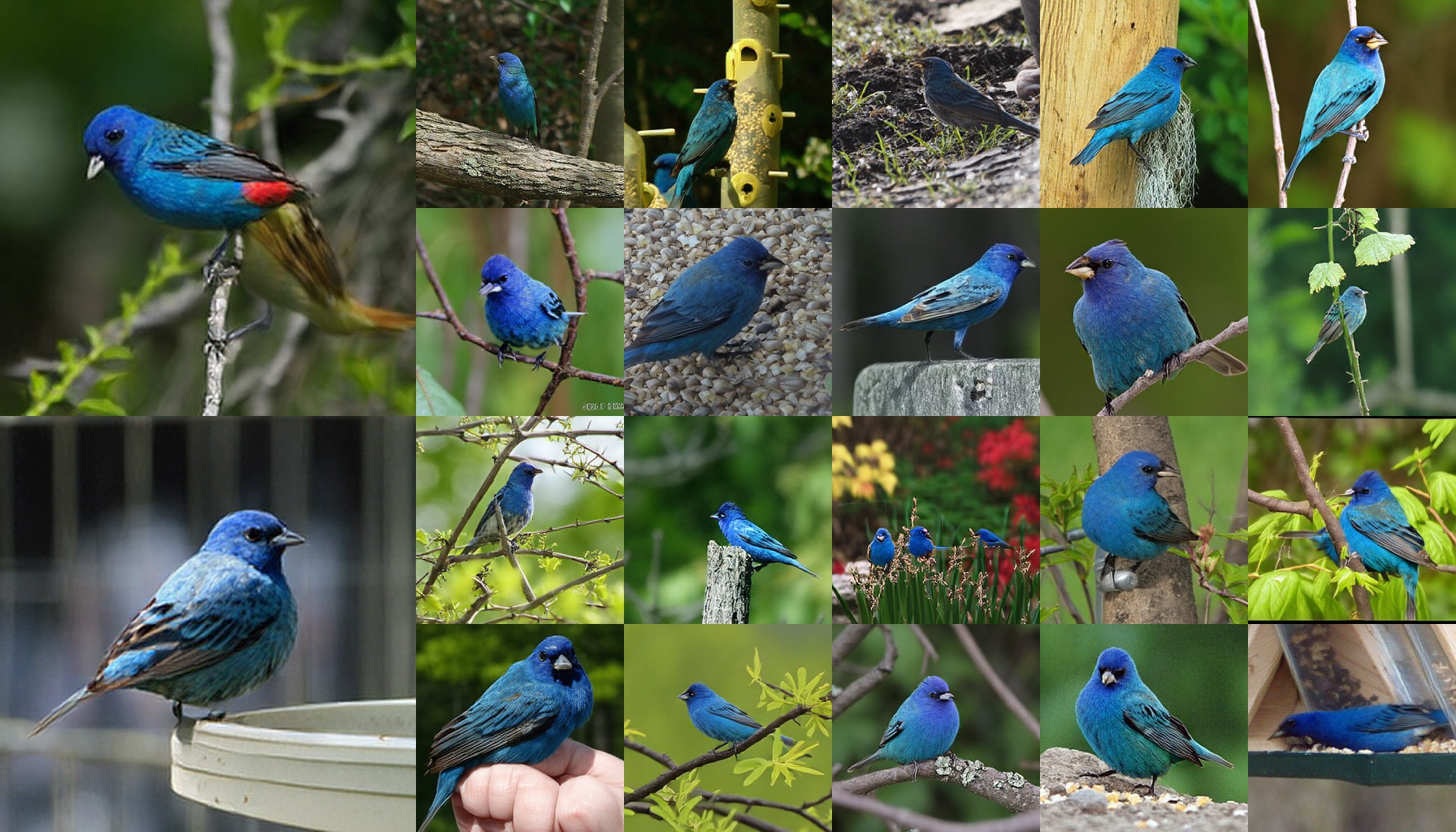}
    \caption{class 0014: indigo bunting}
\end{subfigure}

\begin{subfigure}[t]{0.47\linewidth}
    \includegraphics[width=\linewidth]{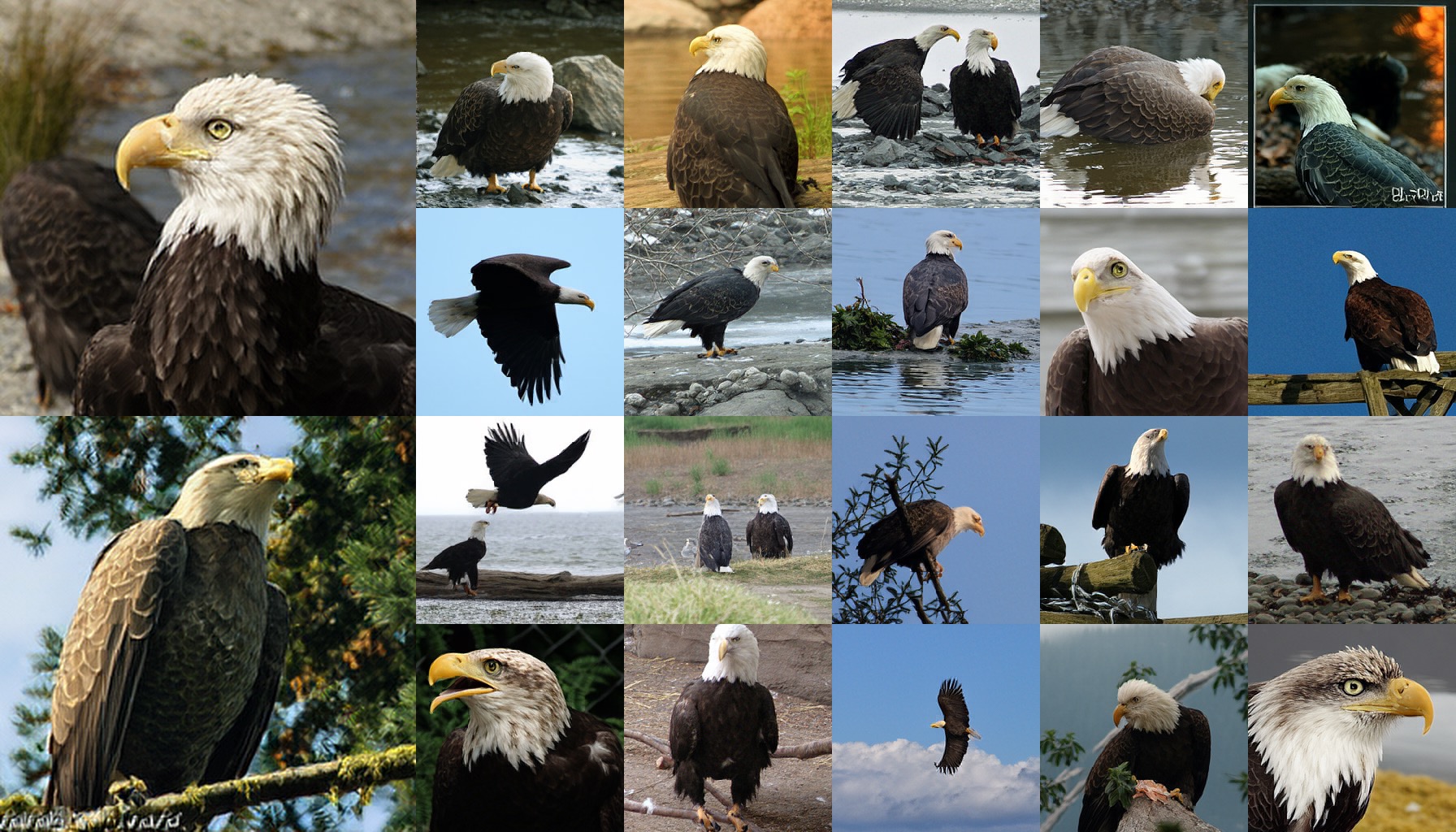}
    \caption{class 0022: bald eagle, American eagle}
\end{subfigure}
\hspace{0.1cm}
\begin{subfigure}[t]{0.47\linewidth}
    \includegraphics[width=\linewidth]{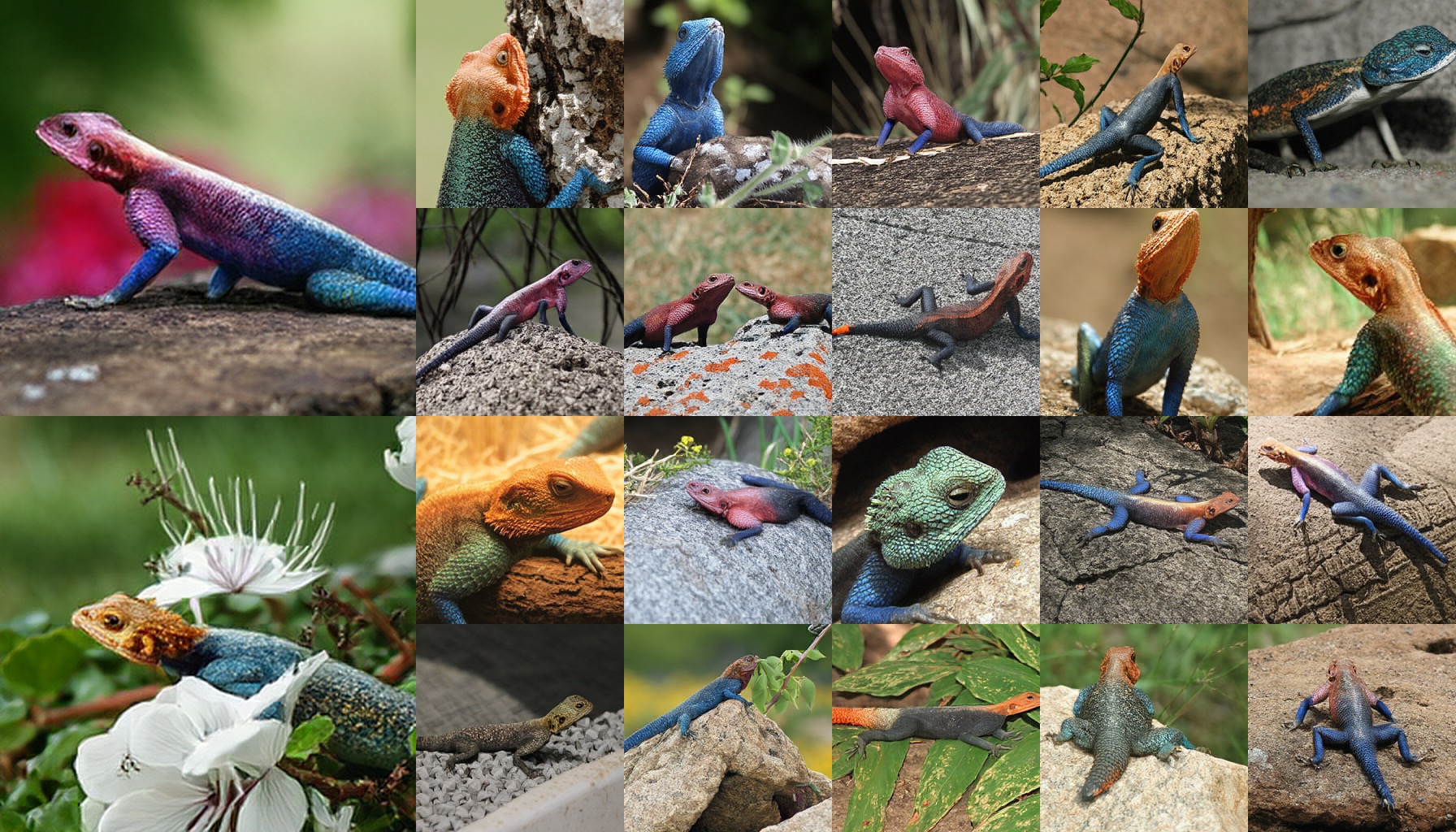}
    \caption{class 0042: agama}
\end{subfigure}

\begin{subfigure}[t]{0.47\linewidth}
    \includegraphics[width=\linewidth]{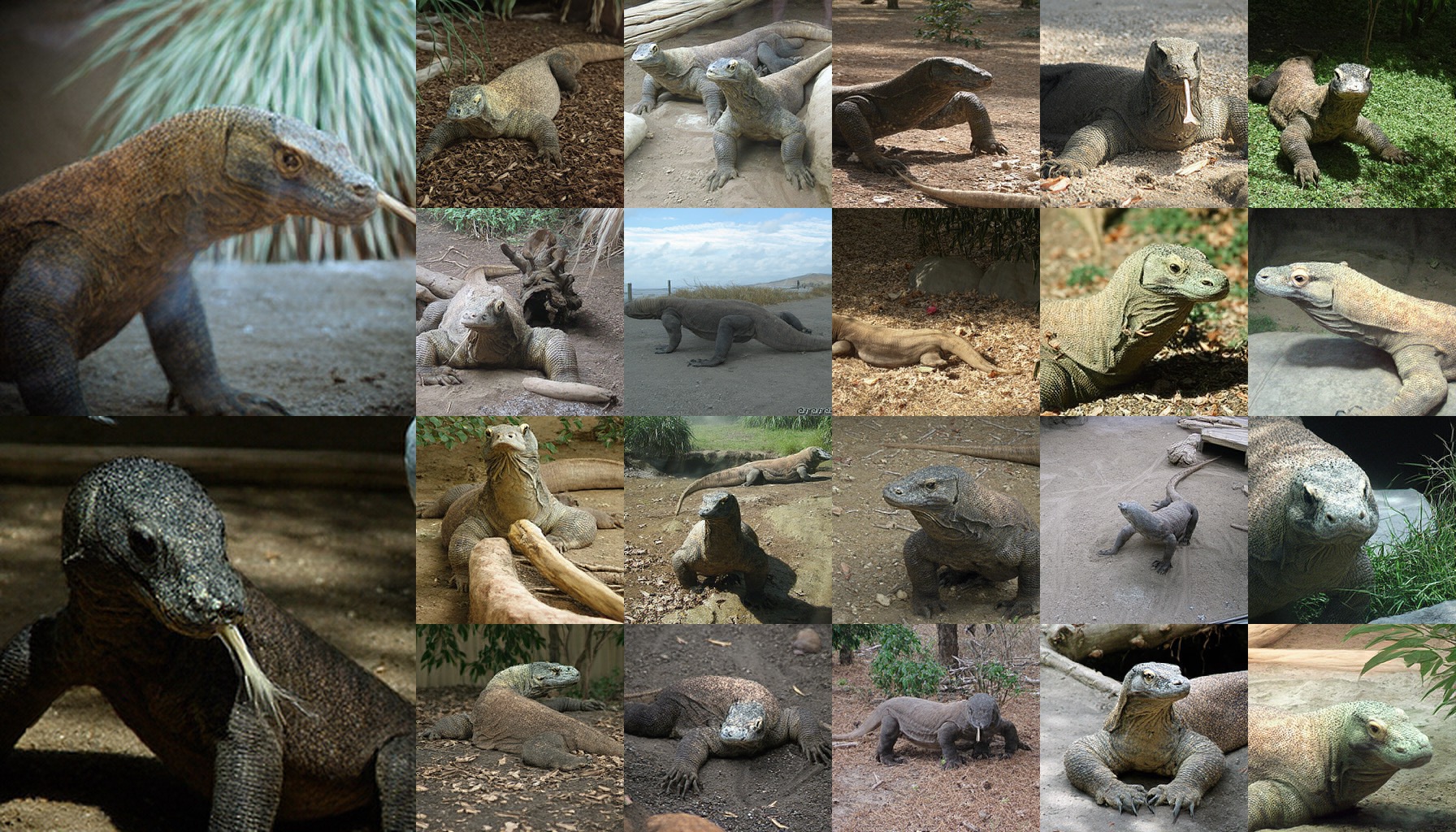}
    \caption{class 0048: Komodo dragon}
\end{subfigure}
\hspace{0.1cm}
\begin{subfigure}[t]{0.47\linewidth}
    \includegraphics[width=\linewidth]{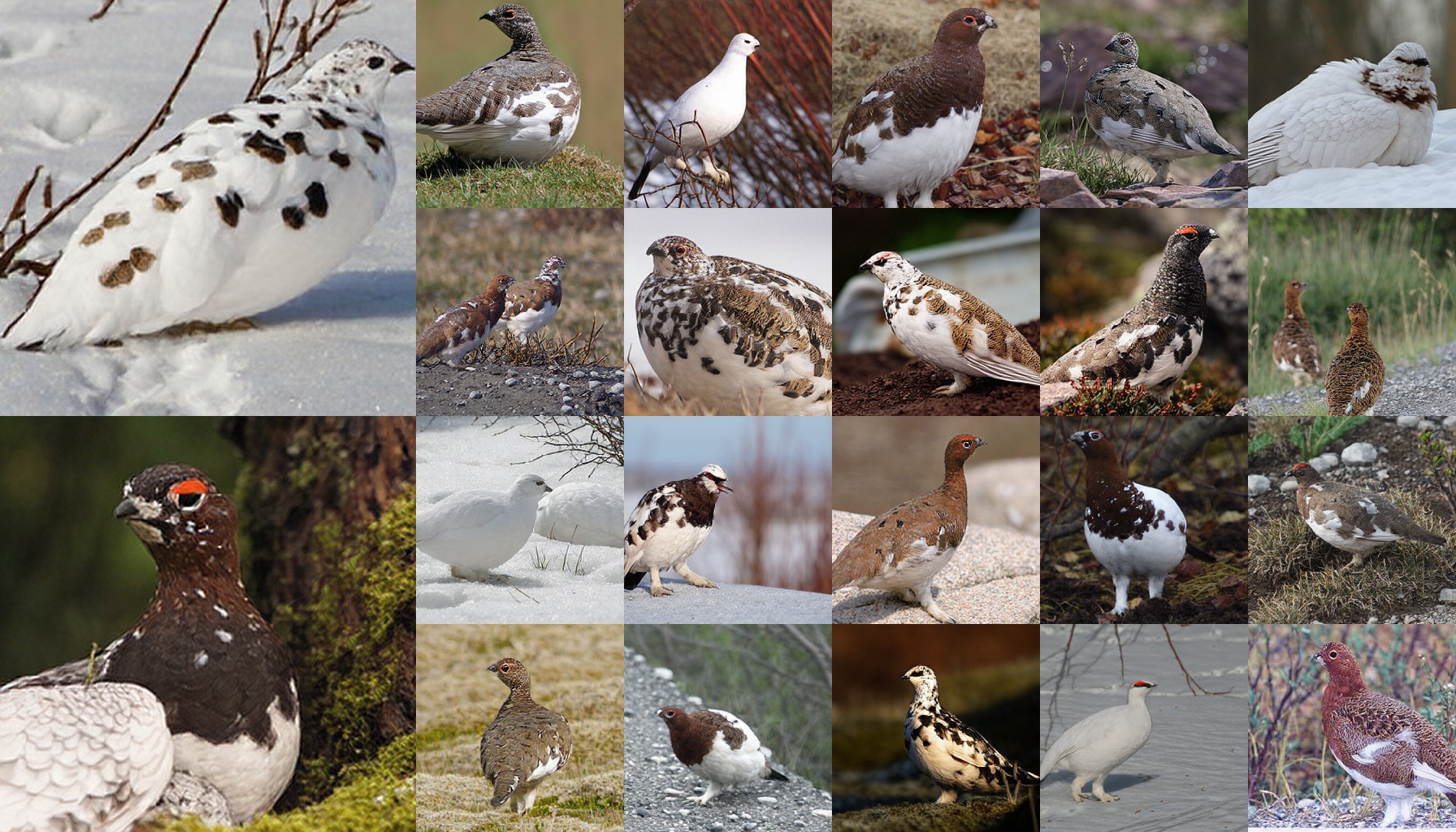}
    \caption{class 0081: ptarmigan}
\end{subfigure}

\begin{subfigure}[t]{0.47\linewidth}
    \includegraphics[width=\linewidth]{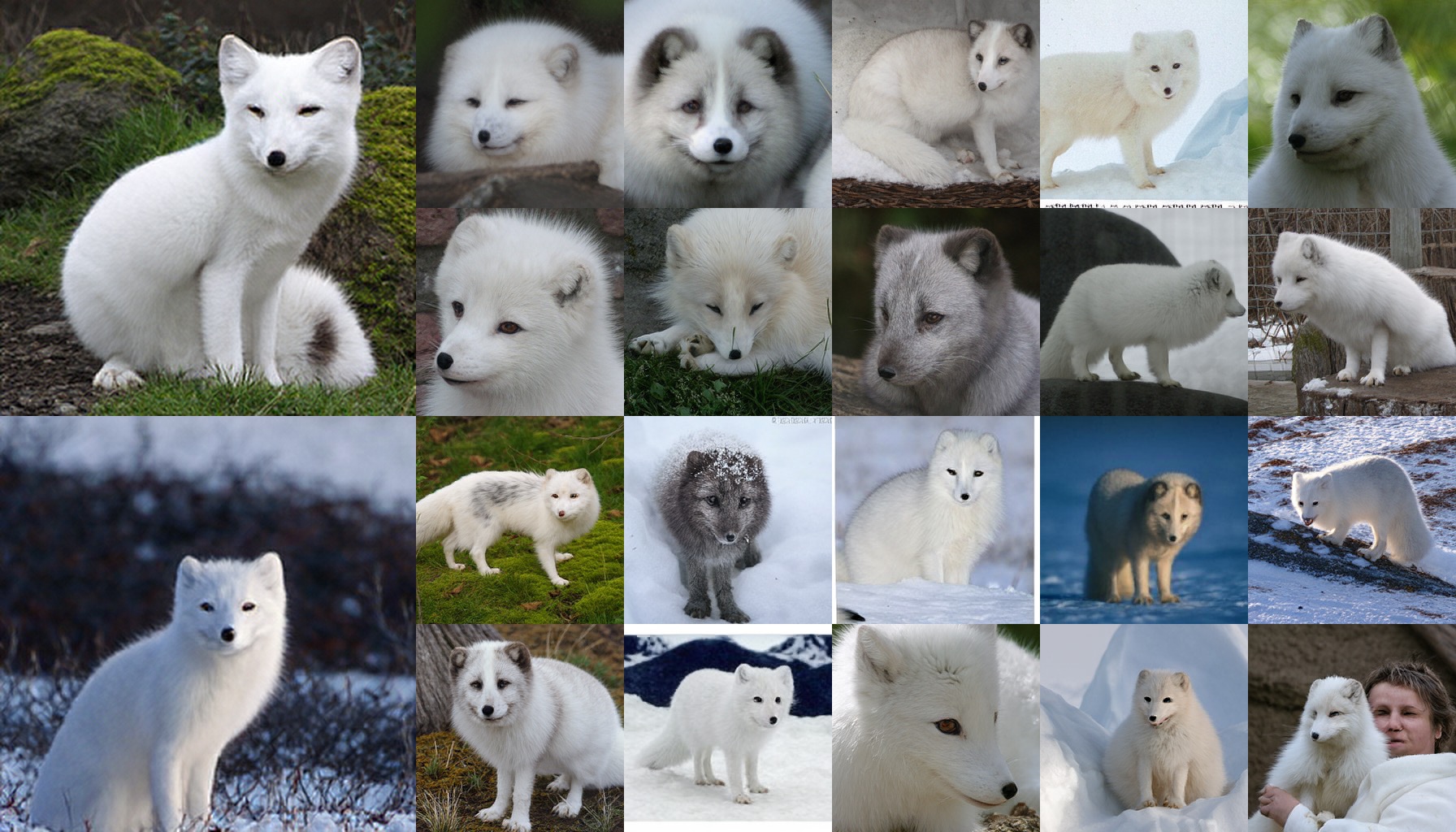}
    \caption{class 0279: Arctic fox}
\end{subfigure}
\hspace{0.1cm}
\begin{subfigure}[t]{0.47\linewidth}
    \includegraphics[width=\linewidth]{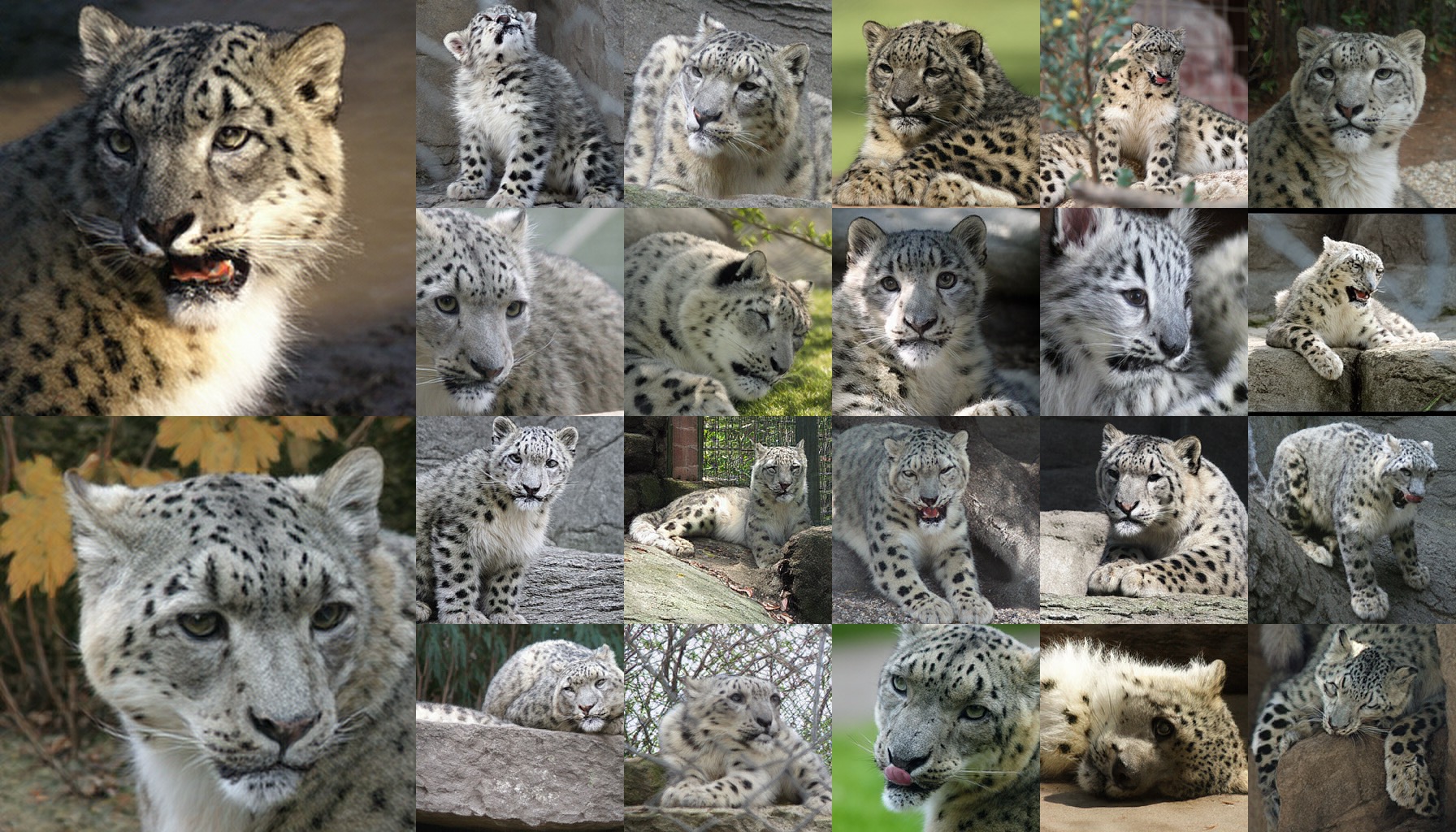}
    \caption{class 0289: snow leopard}
\end{subfigure}

\begin{subfigure}[t]{0.47\linewidth}
    \includegraphics[width=\linewidth]{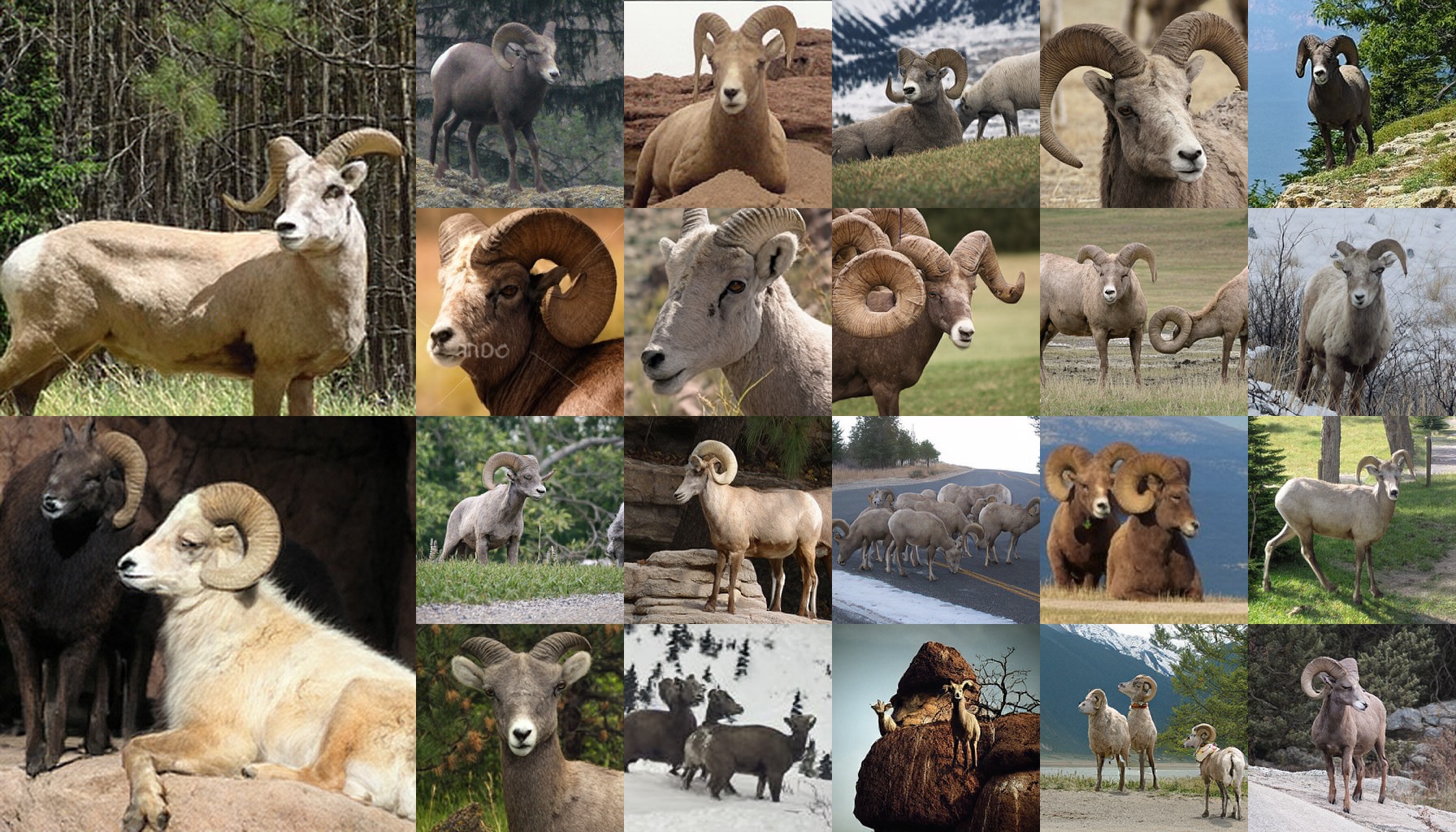}
    \caption{class 0349: bighorn sheep}
\end{subfigure}
\hspace{0.1cm}
\begin{subfigure}[t]{0.47\linewidth}
    \includegraphics[width=\linewidth]{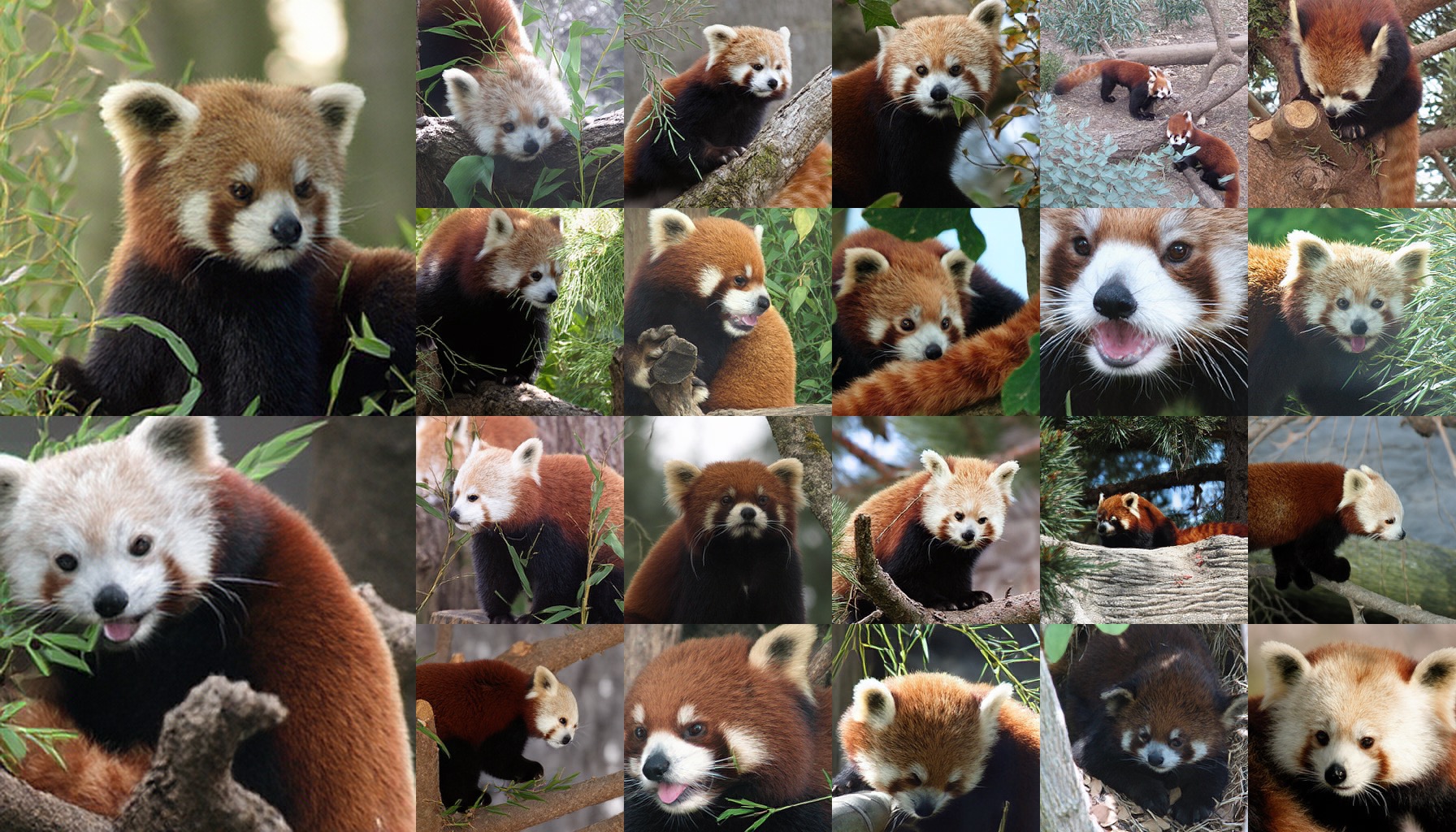}
    \caption{class 0387: red panda}
\end{subfigure}

\caption{\textbf{6-NFE Generation Results.} Examples of class-conditional generation on ImageNet $256\times256$.}
\end{figure*}

\begin{figure*}[t]
\centering

\begin{subfigure}[t]{0.47\linewidth}
    \includegraphics[width=\linewidth]{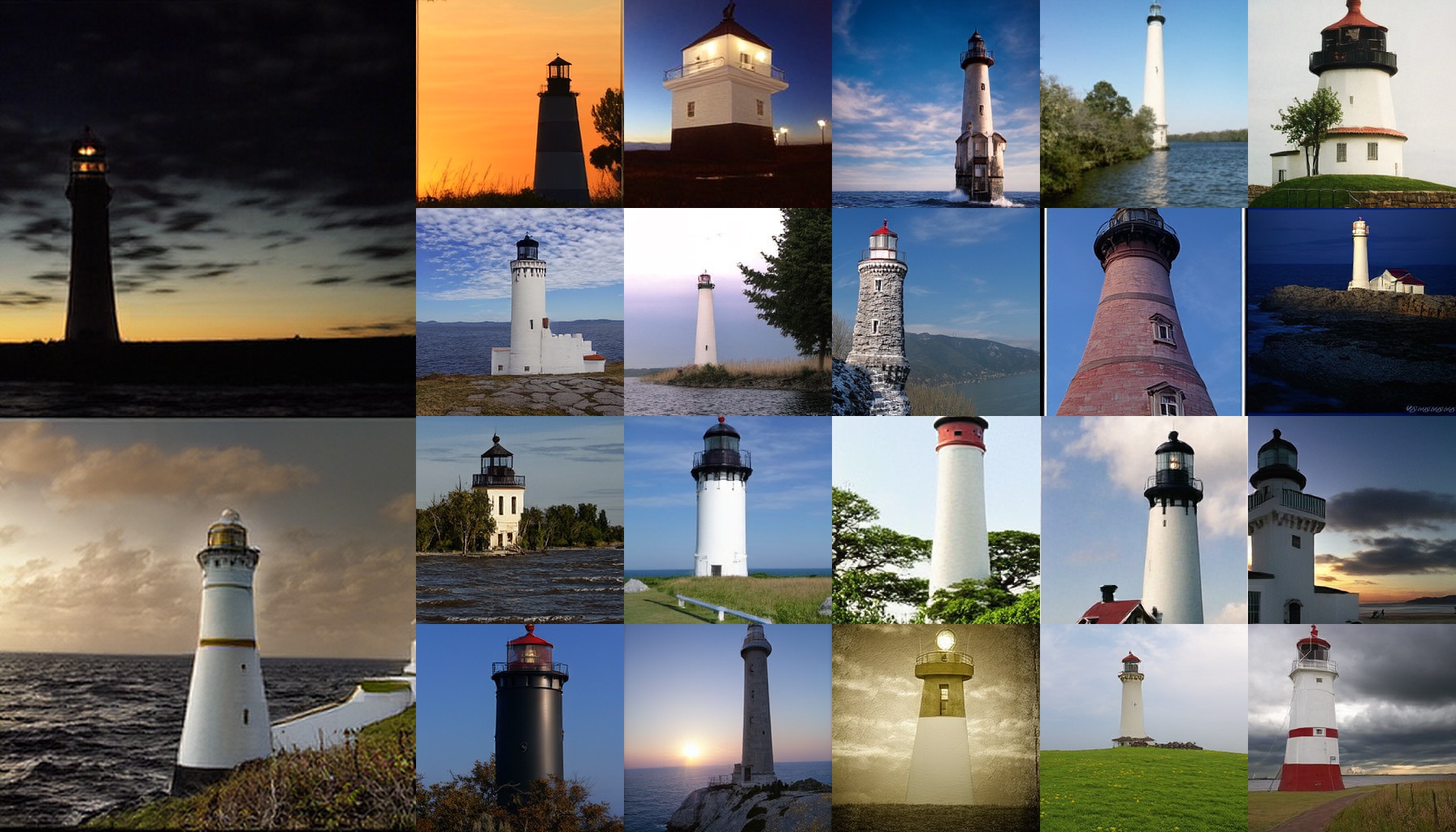}
    \caption{class 0437: beacon}
\end{subfigure}
\hspace{0.1cm}
\begin{subfigure}[t]{0.47\linewidth}
    \includegraphics[width=\linewidth]{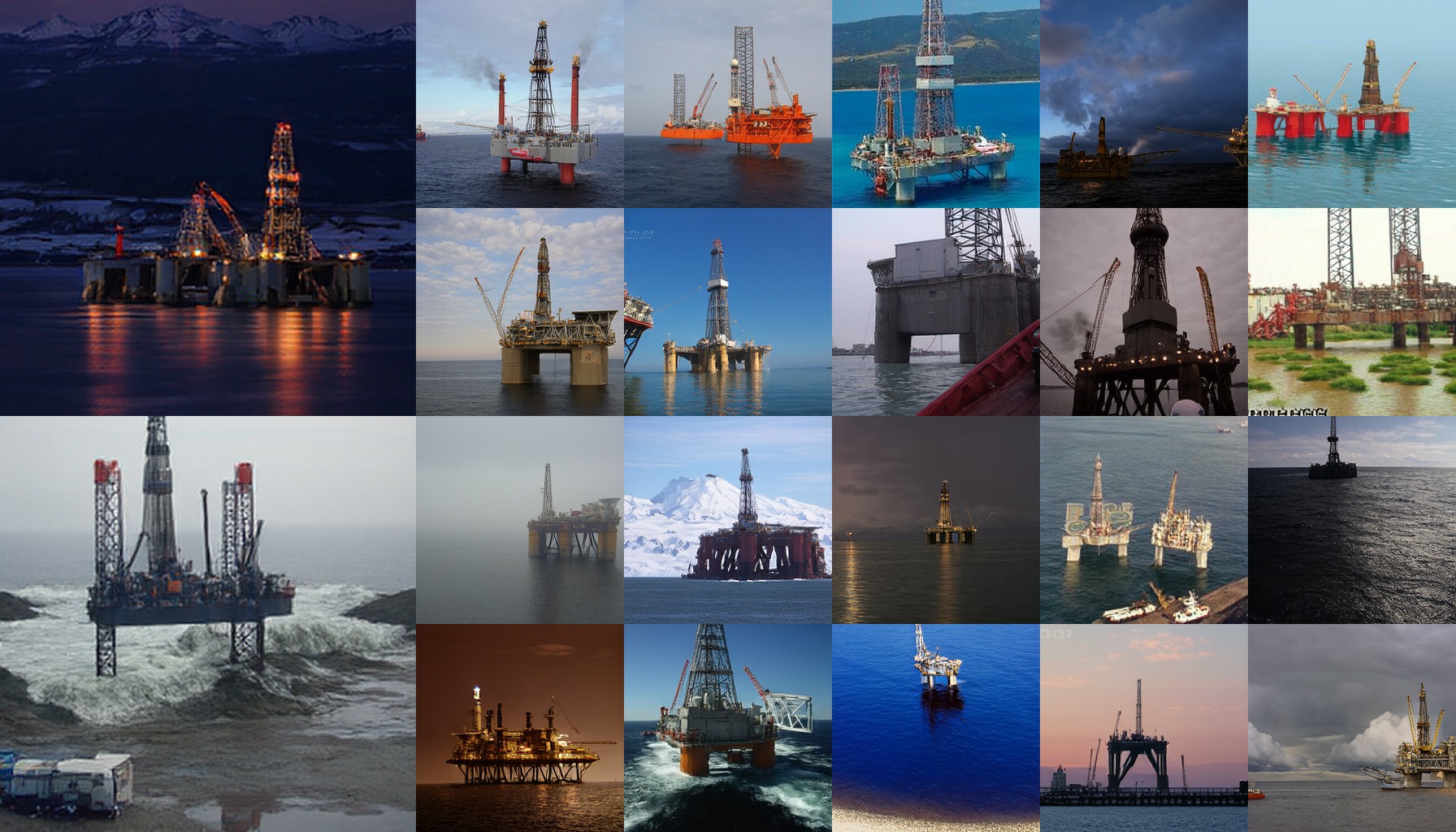}
    \caption{class 0540: drilling platform}
\end{subfigure}

\begin{subfigure}[t]{0.47\linewidth}
    \includegraphics[width=\linewidth]{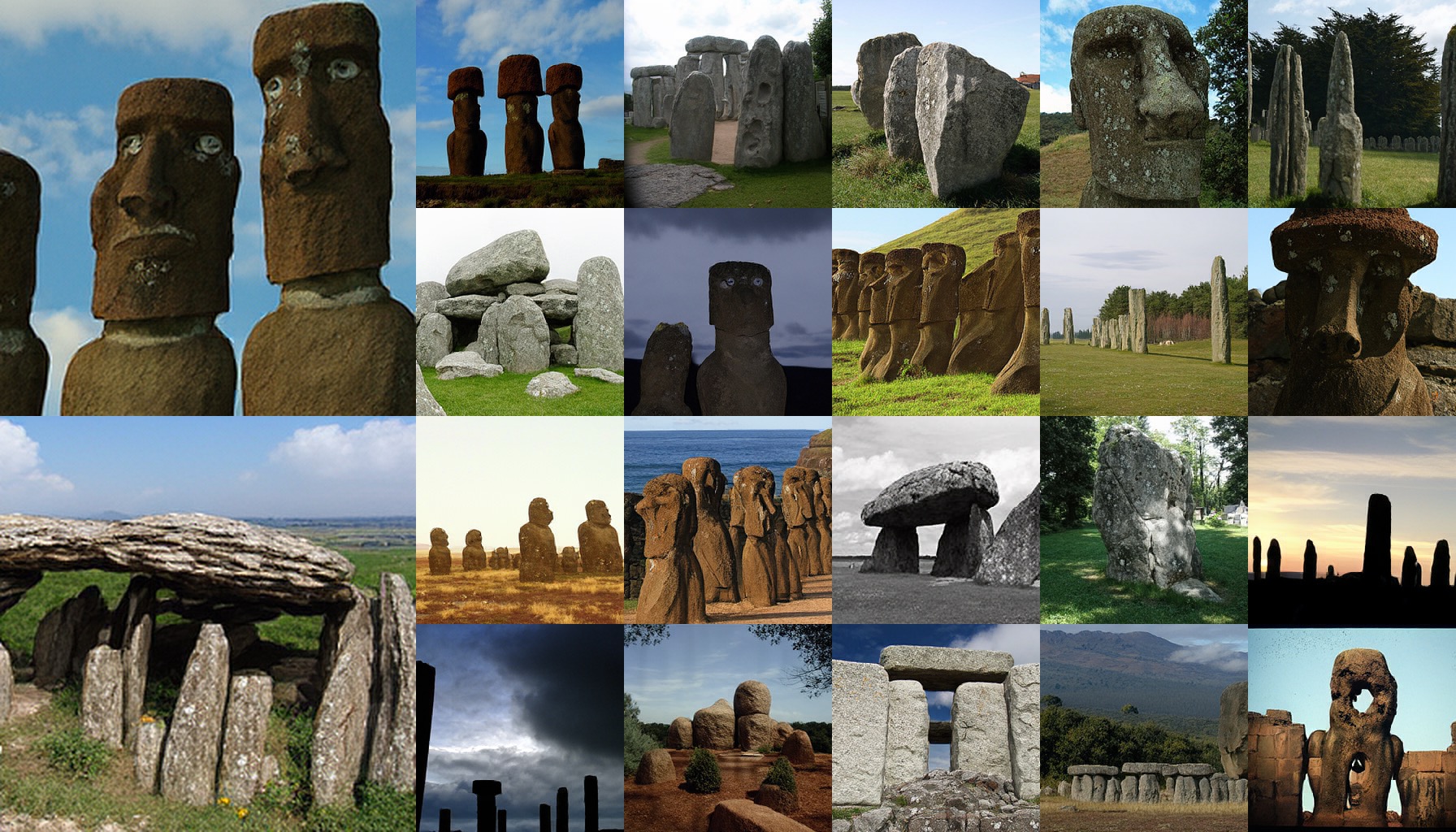}
    \caption{class 0649: megalith}
\end{subfigure}
\hspace{0.1cm}
\begin{subfigure}[t]{0.47\linewidth}
    \includegraphics[width=\linewidth]{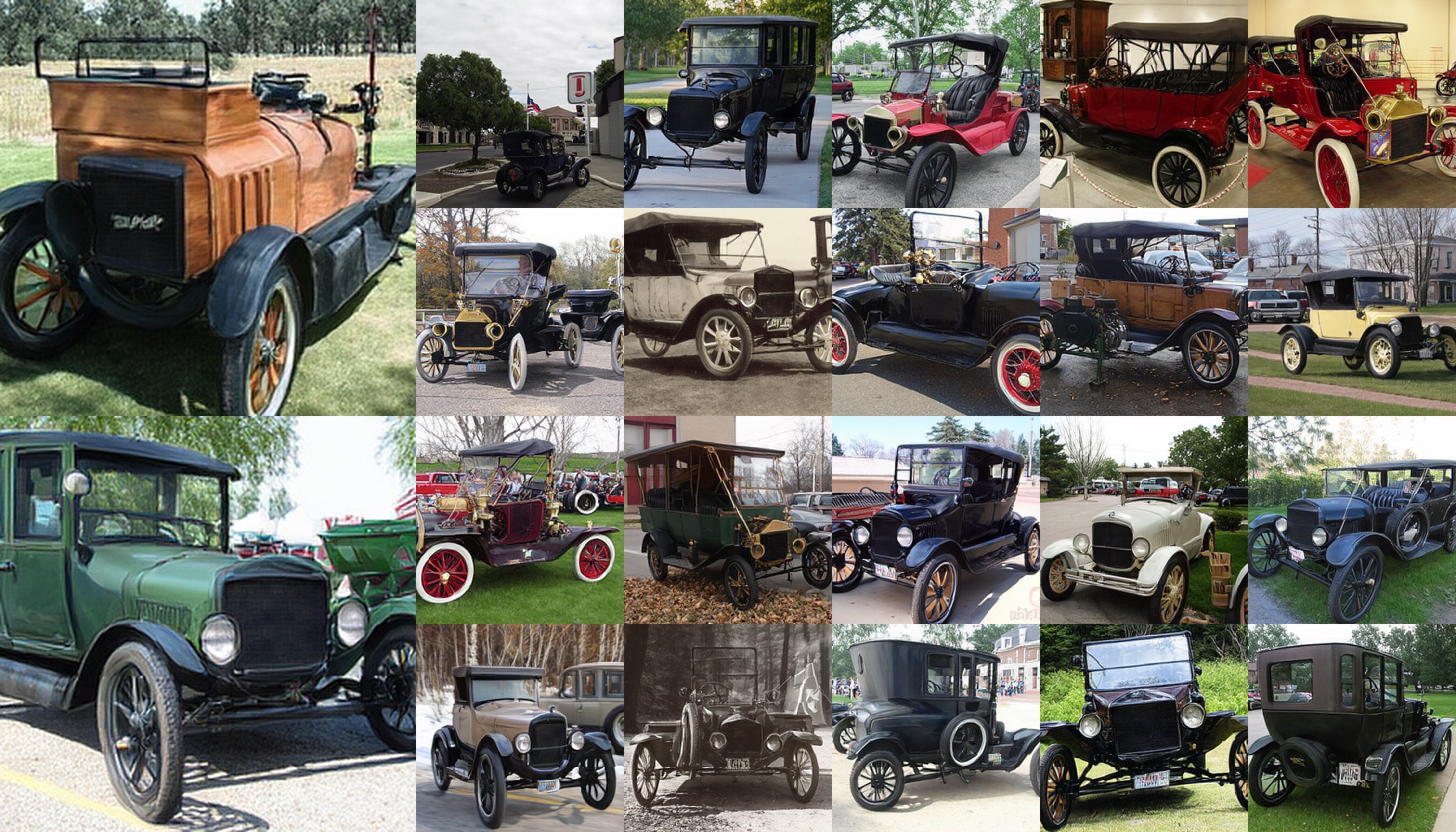}
    \caption{class 0661: Model T}
\end{subfigure}

\begin{subfigure}[t]{0.47\linewidth}
    \includegraphics[width=\linewidth]{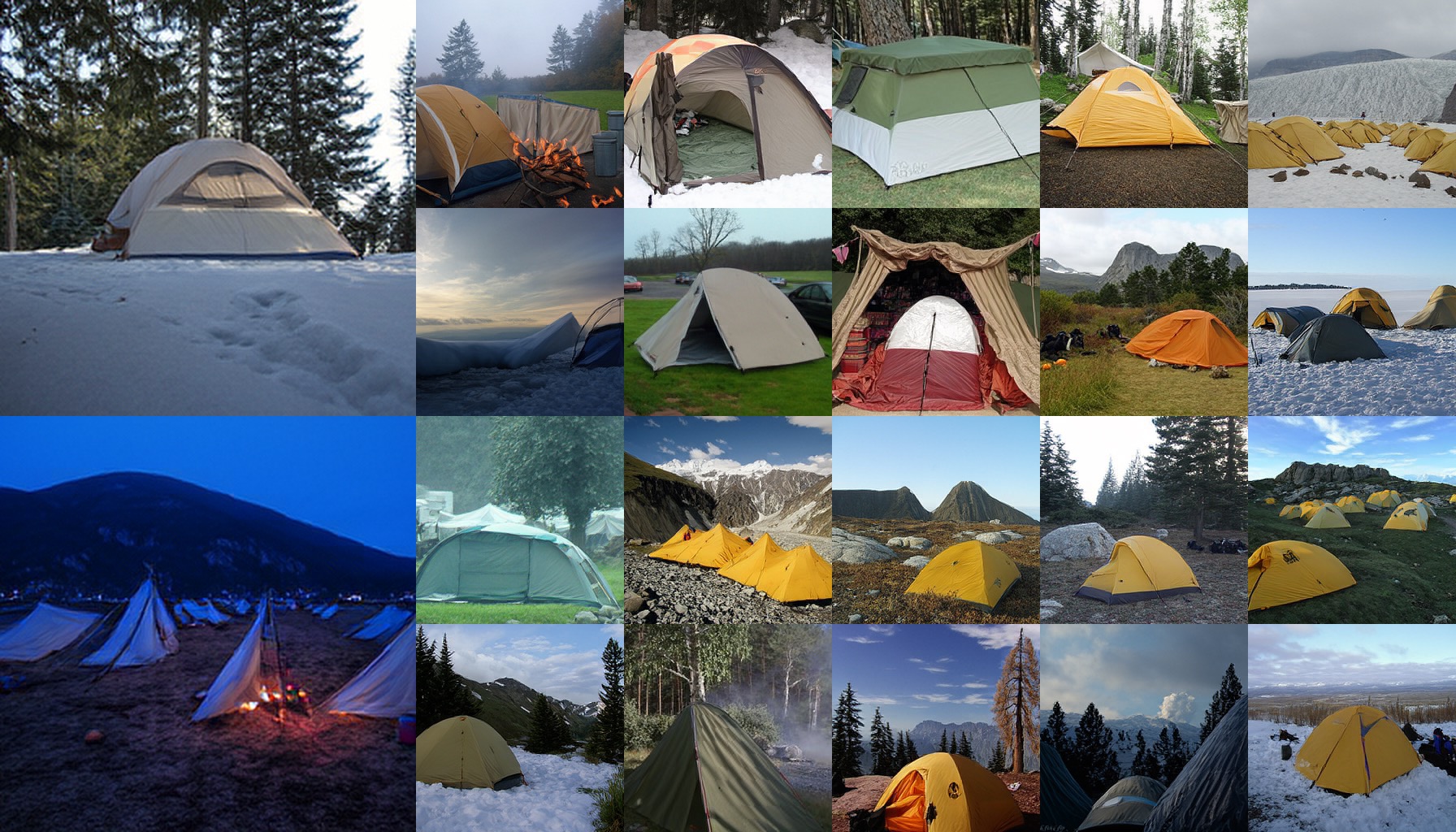}
    \caption{class 0672: mountain tent}
\end{subfigure}
\hspace{0.1cm}
\begin{subfigure}[t]{0.47\linewidth}
    \includegraphics[width=\linewidth]{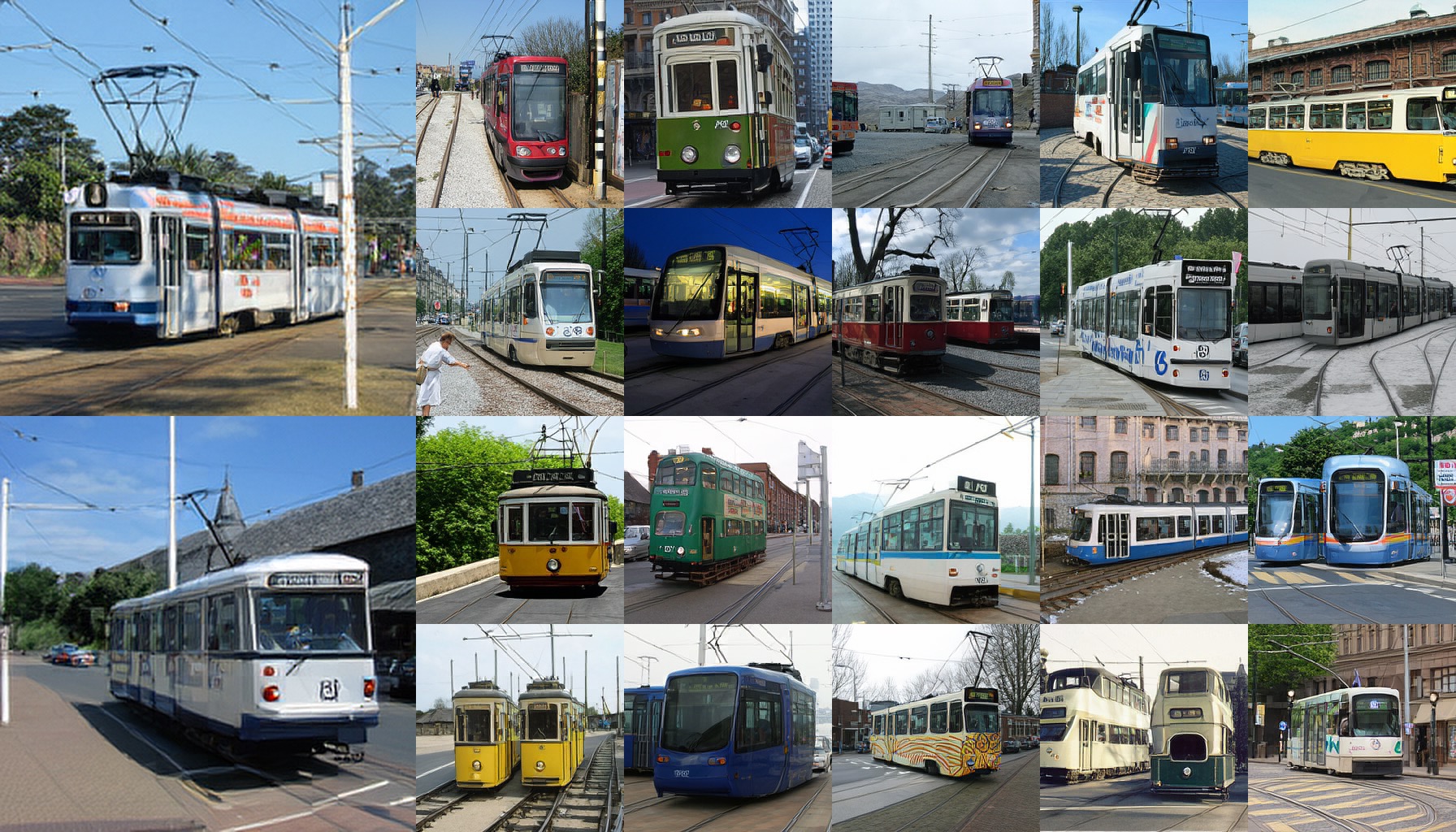}
    \caption{class 0829: streetcar}
\end{subfigure}

\begin{subfigure}[t]{0.47\linewidth}
    \includegraphics[width=\linewidth]{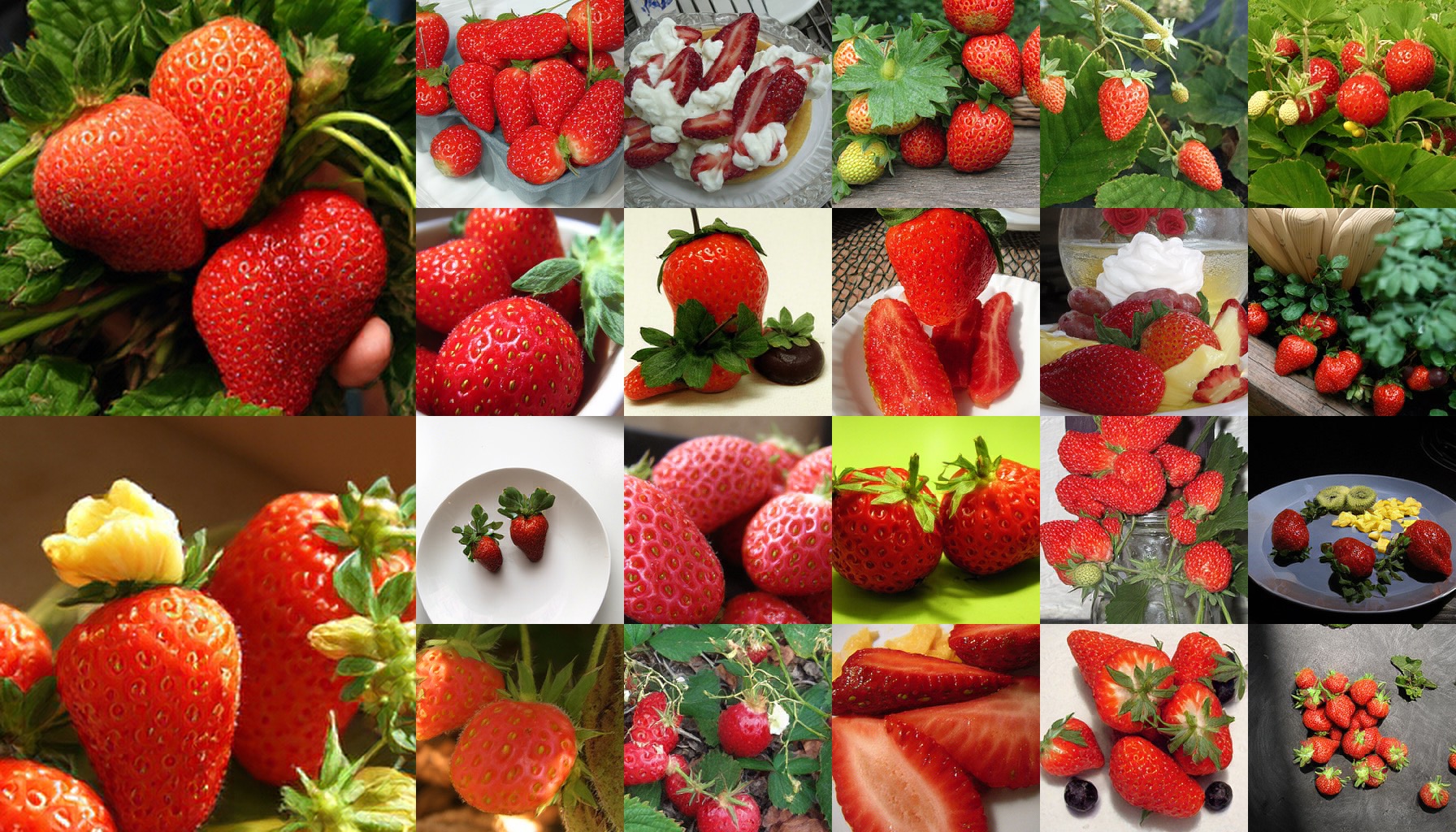}
    \caption{class 0949: strawberry}
\end{subfigure}
\hspace{0.1cm}
\begin{subfigure}[t]{0.47\linewidth}
    \includegraphics[width=\linewidth]{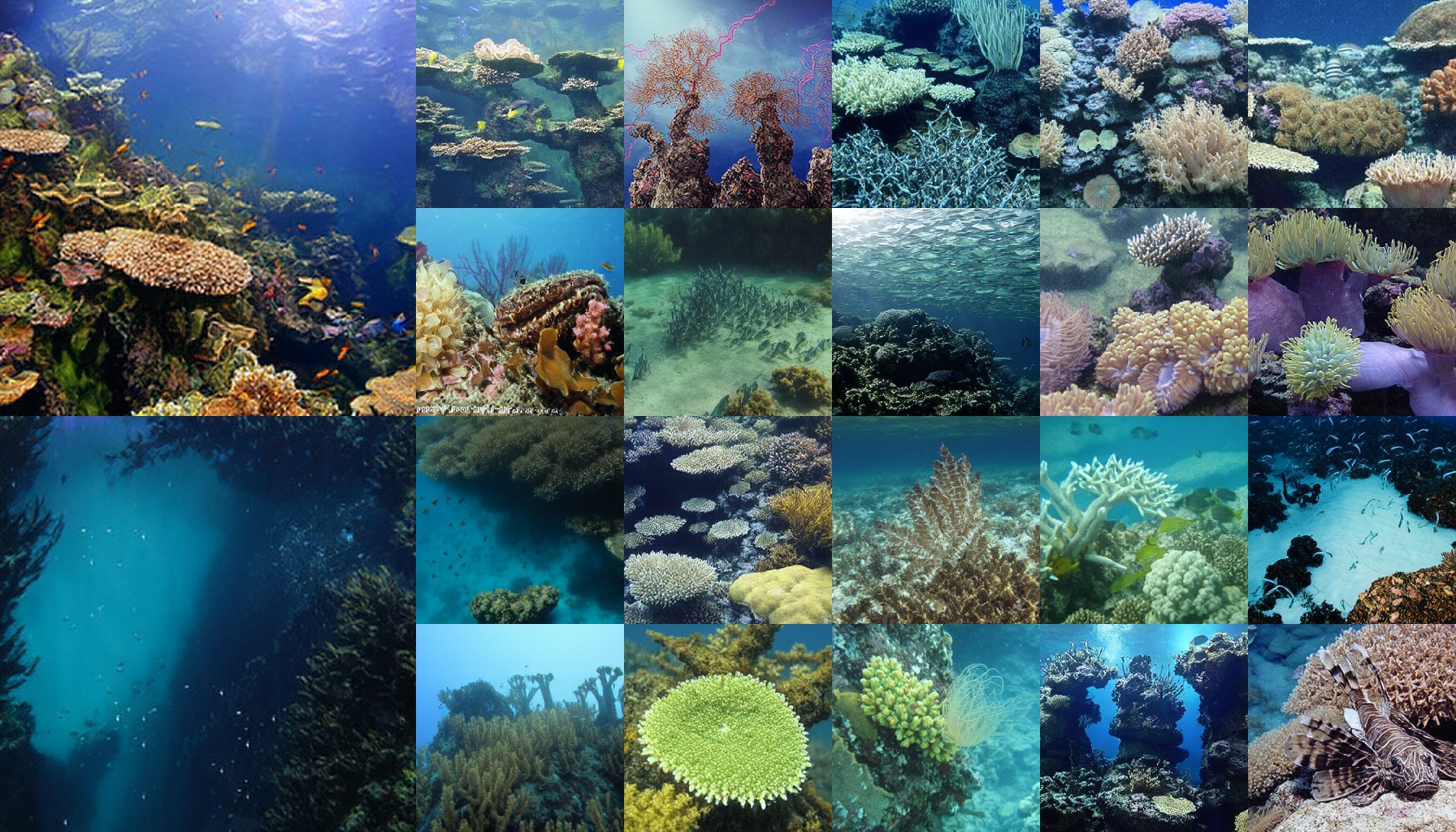}
    \caption{class 0973: coral reef}
\end{subfigure}

\begin{subfigure}[t]{0.47\linewidth}
    \includegraphics[width=\linewidth]{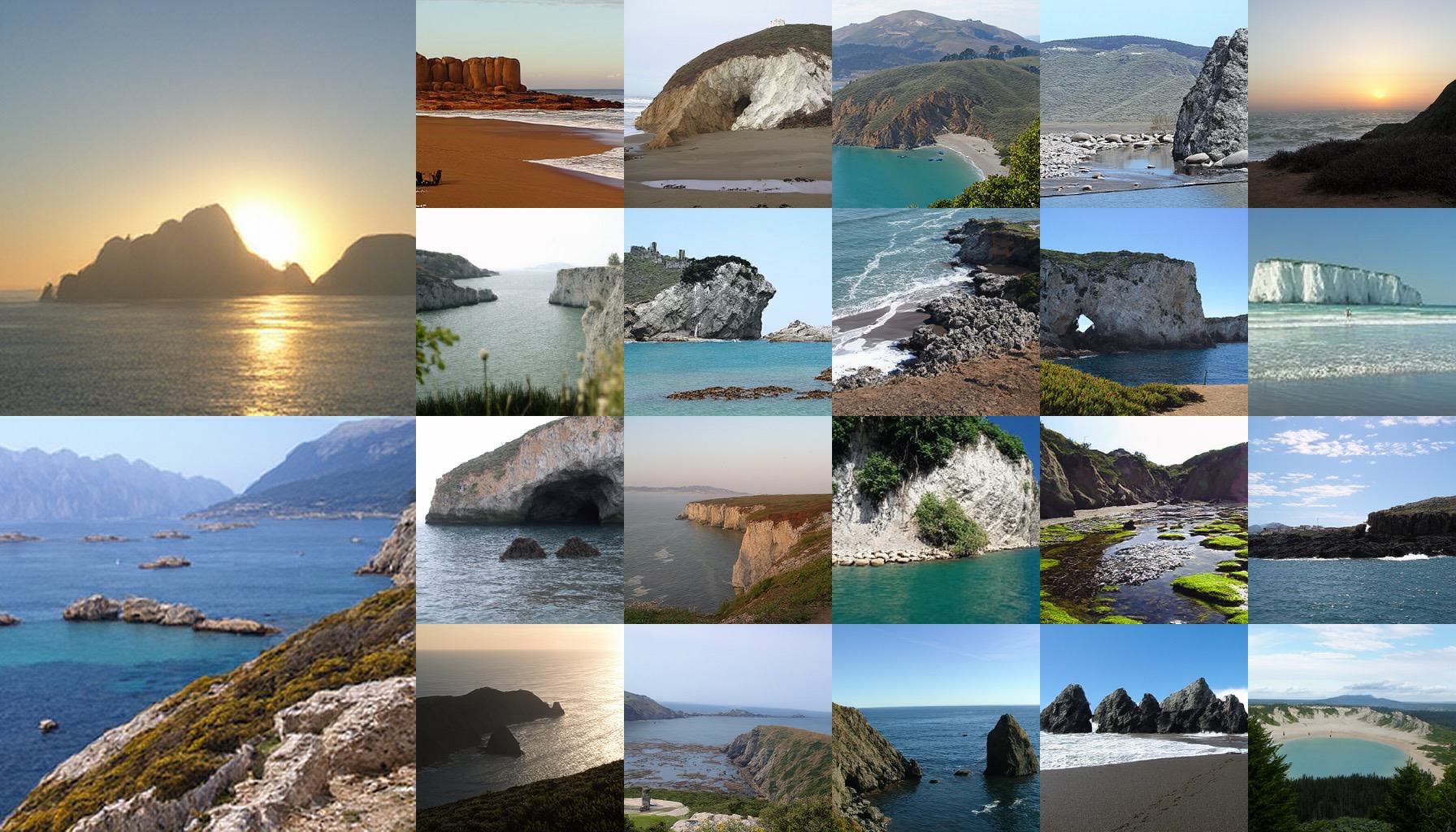}
    \caption{class 0976: promontory}
\end{subfigure}
\hspace{0.1cm}
\begin{subfigure}[t]{0.47\linewidth}
    \includegraphics[width=\linewidth]{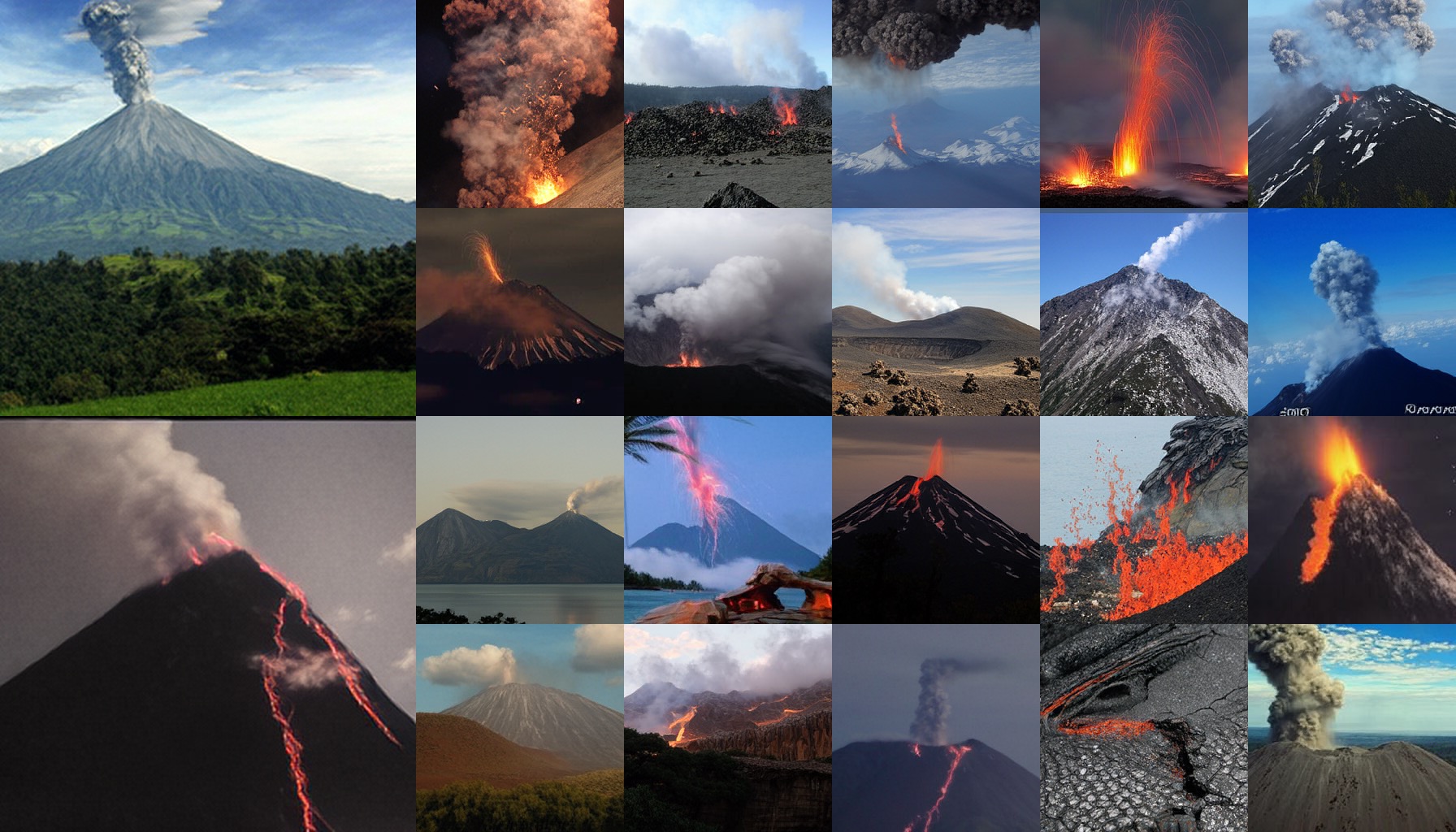}
    \caption{class 0980: volcano}
\end{subfigure}

\caption{\textbf{6-NFE Generation Results.} Examples of class-conditional generation on ImageNet $256\times256$.}
\end{figure*}

\end{document}